\documentclass[10pt,twocolumn,letterpaper]{article}

\usepackage[pagenumbers]{cvpr} % To force page numbers, e.g. for an arXiv version
\usepackage[accsupp]{axessibility}
\usepackage{graphicx}
\usepackage{amsmath}
\usepackage{amssymb}
\usepackage{booktabs}
\usepackage{paralist}
\usepackage[abs]{overpic}
\usepackage{xcolor}

\usepackage[pagebackref,breaklinks,colorlinks,citecolor=violet]{hyperref}

\usepackage[capitalize]{cleveref}
\crefname{section}{Sec.}{Secs.}
\Crefname{section}{Section}{Sections}
\Crefname{table}{Table}{Tables}
\crefname{table}{Tab.}{Tabs.}

\def\eg{\emph{e.g.}}
\def\etal{\emph{et al.}}

\def\ie{\emph{i.e.}}
\def\vs{\emph{v.s.}}

\def\Ourmodel{SLT-Net}

\newcommand\blfootnote[1]{%
  \begingroup
  \renewcommand\thefootnote{}\footnote{#1}%
  \addtocounter{footnote}{-1}%
  \endgroup
}
   
\newcommand{\supp}[1]{\textcolor{magenta}{#1}}
\def\figref#1{Figure~\ref{#1}}
\def\tabref#1{Table~\ref{#1}}

\def\cvprPaperID{6761} % *** Enter the CVPR Paper ID here
\def\confName{CVPR}
\def\confYear{2022}

\begin{document}

\title{Implicit Motion Handling for Video Camouflaged Object Detection}

\author{\textsuperscript{*}Xuelian Cheng$^{1}$,
\textsuperscript{*}Huan Xiong$^{3}$,
$^\dagger$Deng-Ping Fan$^{4}$, Yiran Zhong$^{6,7}$, \\
Mehrtash Harandi$^{1,8}$, Tom Drummond$^{1}$, Zongyuan Ge$^{1,2,5}$ \\
$^{1}$Faculty of Engineering, Monash University, 
$^{2}$eResearch Centre, Monash University \\
$^{3}$Mohamed bin Zayed University of Artificial Intelligence, 
$^{4}$CVL, ETH Zurich,\\
$^{5}$Airdoc Research Australia, 
$^{6}$SenseTime Research, $^{7}$Shanghai AI Laboratory,
$^{8}$Data61, CSIRO \\
%\tt\small{\{xuelian.cheng, mehrtash.harandi, zongyuan.ge\}@monash.edu} \\ 
%\tt\small{\{huan.xiong.math, dengpfan, zhongyiran\}@gmail.com, tom.drummond@unimelb.edu.au}
} 
%\maketitle

\twocolumn[{%
\renewcommand\twocolumn[1][]{#1}%
\maketitle
\vspace{-25pt}
\begin{center}
    \centering
    \captionsetup{type=figure}
    \begin{overpic}[width=.98\textwidth,height=.14\textwidth]{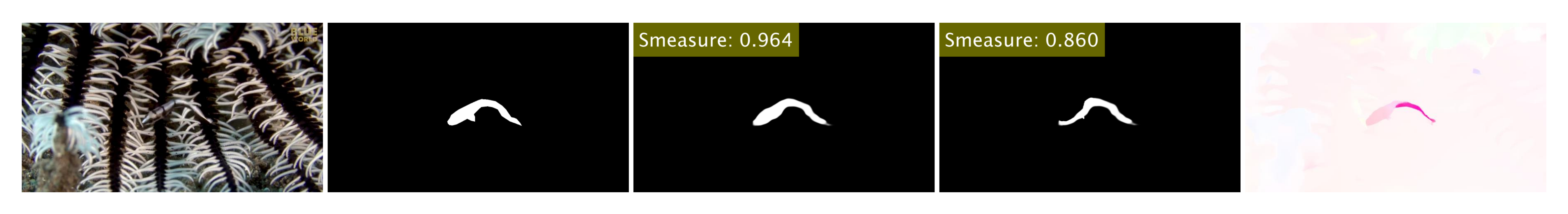}
    \put(27, -3) {\small (a) Frame $T$}
    \put(130, -3) {\small (b) GT}
    \put(230, -3) {\small (c) Ours}
    \put(305, -3) {\small (d) Flow Warping}
    \put(420, -3) {\small (e) Flow}
    \end{overpic}
    % \vspace{0.5pt}
    \captionof{figure}{\small 
    Implicit (ours) \emph{vs} Explicit (optical flow based) motion handling. Our result (c) of frame $T$ is generated by the frame $T-1$ and $T$. Figure (d) is the warped result using our result of frame $T-1$ and a pre-computed optical flow (e). As we will discuss,  the accumulated error in explicit methods (\ie, optical flow) for camouflage objects can lead to a tangible drop in the accuracy (dropped by $10.79\%$).
    }\label{fig:first_page}
\vspace{8pt}
\end{center}
}]

\blfootnote{* Indicates equal contribution; $^\dagger$ Corresponding author (dengpfan@gmail.com). Work was done while Xuelian Cheng was an MBZUAI visiting scholar mentored by Deng-Ping Fan.}

%%%%%%%%% ABSTRACT
\begin{abstract}
\vspace{-10pt}
We propose a new video camouflaged object detection (VCOD) framework that can exploit both short-term dynamics and long-term temporal consistency to detect camouflaged objects from video frames. An essential property of camouflaged objects is that they usually exhibit patterns similar to the background and thus make them hard to identify from still images. Therefore, effectively handling temporal dynamics in videos becomes the key for the VCOD task as the camouflaged objects will be noticeable when they move. However, current VCOD methods often leverage homography or optical flows to represent motions, where the detection error may accumulate from both the motion estimation error and the segmentation error. On the other hand, our method unifies motion estimation and object segmentation within a single optimization framework. Specifically, we build a dense correlation volume to implicitly capture motions between neighbouring frames and utilize the final segmentation supervision to optimize the implicit motion estimation and segmentation jointly. Furthermore, to enforce temporal consistency within a video sequence, we jointly utilize a spatio-temporal transformer to refine the short-term predictions. Extensive experiments on VCOD benchmarks demonstrate the architectural effectiveness of our approach. We also provide a large-scale VCOD dataset named \href{https://drive.google.com/file/d/1FB24BGVrPOeUpmYbKZJYL5ermqUvBo_6/view}{\textbf{MoCA-Mask}} with pixel-level handcrafted ground-truth masks and construct a comprehensive \textbf{VCOD benchmark} with previous methods to facilitate research in this direction. 
Dataset Link: \url{https://xueliancheng.github.io/SLT-Net-project}.
%Source code: \url{https://github.com/XuelianCheng/SLT-Net}.

\end{abstract}

%%%%%%%%% BODY TEXT
\vspace{-10pt}
\section{Introduction}\label{sec:intro}
\vspace{-5pt}
Video Camouflaged Object Detection (VCOD) is the task of discovering objects in a video that, appearance-wise, exhibit a great deal of similarity to the background scene. Despite enjoying wide applications (\eg, surveillance and security \cite{liu2019concealed}, autonomous driving \cite{ranjan2019competitive,cheng2019noise},  medical image segmentation~\cite{fan2020pra, wu2021jcs}, locust detection \cite{kumar2021early} and robotics \cite{michels2005high}), the problem of Camouflaged Object Detection (COD) is a daunting task as camouflaged objects are often indistinguishable to naked-eyes. This, in turn, has made VCOD a relatively under-explored problem in computer vision, as compared to several related problems such as video object detection (VOD) \cite{yang2019anchor,beery2020context}, video salient object detection (VSOD)  \cite{ji2021FSNet}, and video motion segmentation (VMS) \cite{7025056,yang2021selfsupervised}. 

In most computer vision tasks (\eg, instance segmentation~\cite{zhongicpr18}, saliency detection~\cite{Zhang_2021_ICCV}), it is assumed that objects have clear boundaries. This allows us to formulate the problem at the image level and even consider improvements if motion information is available. In contrast, object boundaries are ambiguous and indistinguishable when it comes to detecting camouflaged objects. This not only makes detection from images challenging, but also results in inaccurate estimation of optical flow and motion cues in videos~\cite{NEURIPS2020_add5aebf,Zhong_2019_CVPR,Wang_2021_CVPR}. 

The lack of clear boundaries means that the appearance of the camouflaged object resembles the background. This shows itself as two fundamental difficulties: \textbf{1)} the object boundaries are often seamlessly blended into the background and is observable only when the object moves; 
\textbf{2)} the object usually has repetitive textures similar to the environment; hence determining the movement of pixels across frames to estimate the motion (\eg, as done in optical flow) is erratic and erroneous. As the first difficulty, to successfully address VCOD, a neural network needs to effectively discover the nuances between the camouflaged object and the background with the help of motion information. %as a result of the first difficulty.
Moreover, the motion information is inherently noisy and inaccurate according to the second difficulty, as shown in \figref{fig:first_page}. As such, employing VOD, VSOD, and VMS techniques may fail miserably if naively used or combined to address the VCOD problem.

In this work, we introduce \textbf{\Ourmodel}, a new method to address VCOD that utilizes short-term dynamics and long-term temporal consistency to detect camouflaged objects in videos. 
Specifically, we employ a short-term dynamic module to \textit{implicitly} capture the motion between consecutive frames. Rather than using optical flow to \textit{explicitly} represent motions, we use a full-range correlation pyramid strategy to represent them implicitly. The primary motivation behind using a correlation pyramid is that even SOTA optical flow algorithms fail to estimate motions for camouflaged objects and their errors accumulate over the video's duration. Also, it allows us to jointly optimize the motion estimation (implicitly) and the predictions with only the detection supervision.
To provide a stable estimation, we further introduce a long-term refinement module to alleviate accumulated inaccuracies in the short-term dynamic module.

We realize the \Ourmodel~as a hybrid neural network with both transformer and CNN components. In particular, we use a transformer structure to encode features for constructing a correlation pyramid. Aside from its design flexibility, features extracted by the transformer contain global contextual information with long-range dependencies and less inductive bias~\cite{wang2021pyramid,zhen2022cosformer}, which we observe to be more distinguishable in estimating the motion.

While the correlation pyramid strategy can effectively capture motions for detecting camouflaged objects, it cannot scale gracefully to long video sequences due to its computational complexity. To solve this issue, we adopt a sequence-to-sequence model with a spatial-temporal transformer to refine the pair-wise prediction with long-term consistency across the videos as we empirically find it is more accurate than the standard ConvLSTM model~\cite{xingjian2015convolutional,Zhong_2018_ECCV}. %\MH{maybe citing the original LSTM paper}.

Moreover, being a less-explored problem, large-scale datasets are not available to evaluate and benchmark VCOD systems. To promote new developments in this domain, we have curated a large-scale VCOD dataset based on the Moving Camouflaged Animals (MoCA)~\cite{lamdouar2020betrayed}. The new dataset, or \textbf{MoCA-Mask} for short, contains 87 video sequences with 22,939 frames in total with pixel-wise ground truth masks. MoCA-Mask encapsulates a variety of challenges, such as complex backgrounds and tiny and well-camouflaged objects. We provide annotations, bounding boxes, and dense segmentation masks for every five
frames for all the videos in the dataset. 
We also provide the first comprehensive benchmark for existing VCOD methods. 
In a nutshell, our contributions are as follows:
\begin{compactitem}
    % \vspace{-3pt}
    \item We propose a new VCOD framework that can effectively model short-term dynamics and long-term temporal consistency from videos, where the motion and the camouflage object segmentation can be jointly optimized through a single optimization target. 
    % \vspace{-3pt}
    \item We collect the first large-scale VCOD dataset, the \textbf{MoCA-Mask} dataset, to promote developments in VCOD as well as a comprehensive VCOD benchmark to facilitate research in VCOD.
    % \vspace{-3pt}
    \item We set a new state-of-the-art on the VCOD task, outperforming a previous SOTA method~\cite{yan2019semi} by $9.88\%$.% on $S_\alpha$.
\end{compactitem}

\begin{figure*}[ht]
\begin{center}
\includegraphics[width=0.65\textwidth]{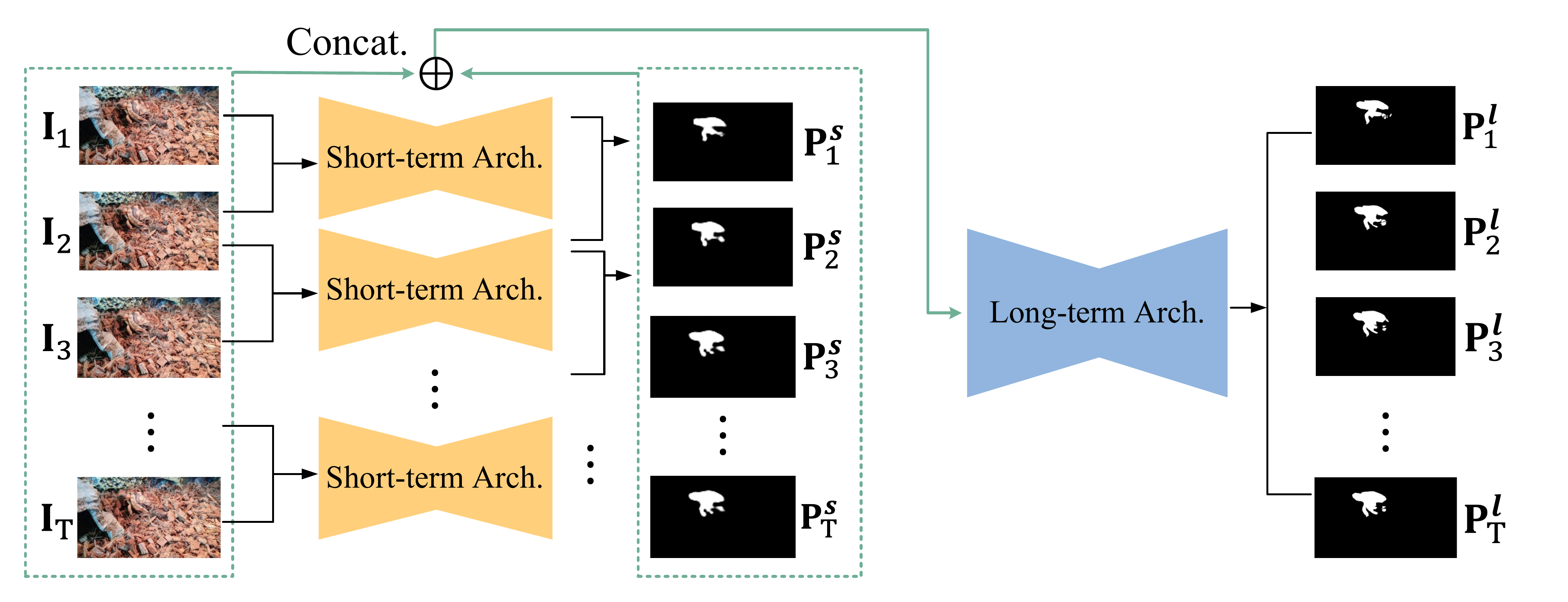}
\end{center}
\vspace{-20pt}
\caption{The overall pipeline of the  \Ourmodel. The \Ourmodel~consists of a short-term detection module and a long-term refinement module. The short-term detection module takes a pair of consecutive frames and predicts the camouflaged object mask for the reference frame. The long-term refinement module takes $\mathrm{T}$ predictions from the short-term detection module along with their corresponding referenced frames to generate the final predictions.}
\label{fig:overall}
\end{figure*}

\section{Related Work}\label{sec:related_work}
% \vspace{-5pt}
%\paragraph{COD.}
\noindent\textbf{COD.} 
Without any prior, even humans can easily miss camouflaged objects. However, once informed that a camouflaged object exists in an image, we can carefully scan the entire image to identify it. 
Inspired by this fact, ANet~\cite{ltnghia-CVIU2019} incorporated classification stream as the awareness of camouflaged objects and segmentation stream. Sharing a similar idea, SINet~\cite{fan2020Camouflage} and PFNet \cite{mei2021camouflaged} addressed the problem by first positioning coarse camouflaged objects and then refining it by segmentation. SINet-v2~\cite{fan2021concealed} extended this idea by incorporating the reverse guidance before learning complementary regions. MGL \cite{zhai2021mutual} incorporated edge details into the segmentation stream via two graph-based modules. By modeling the conspicuousness of camouflaged objects against backgrounds, Lv \etal~\cite{lv2021simultaneously} introduced two new tasks, namely camouflaged object ranking and camouflaged object localization, along with relabeled NC4K dataset.  

% \vspace{-5pt}
%\paragraph{VSOD.}
\noindent\textbf{VSOD.}
To detect salient objects in videos, DLVS \cite{wang2017video} introduced fully convolutional networks for pixel-wise saliency prediction. DSR3 \cite{le2017deeply} exploited an end-to-end 3D neural network to produce video sequences, which incorporates 3D CNN modules combined with recurrent refinement units to predict saliency maps. To better learn temporal information over frames, following works considered SpatioTemporal CRF \cite{le2018video}, pyramid dilated convLSTM \cite{song2018pyramid} in the design of their networks. FGRN \cite{li2018flow}, RCRNet \cite{yan2019semi} adopted extra flow-guided networks to improve temporal coherence. Later, SSAV \cite{Fan2019VideoSal} specifically focused on the saliency shift phenomenon and established a comprehensive benchmark for VSOD. FSNet \cite{ji2021FSNet} leveraged the mutual constraints of appearance and motion cues, demonstrating superior performances to many existing methods.%\MH{this part has a better flow compared to COD}

% \vspace{-5pt}
%\paragraph{VMS.}
\noindent\textbf{VMS.}
The task of VMS focuses on discovering moving objects in videos. Traditional methods usually address this problem by extracting motion boundaries in the flow field and then refining the initial estimate with appearance features~\cite{papazoglou2013fast}, or combining motion and appearance cues by a fusion architecture~\cite{jain2017fusionseg}. Another line of work explicitly leverages optical flow as the input to train a CNN-based network and generate pixel-level motion labels based on supervised learning~\cite{tokmakov2017learning} or in an unsupervised manner~\cite{yang2021selfsupervised}. 

% \vspace{-5pt}
%\paragraph{VCOD.} 
\noindent\textbf{VCOD.}
Different from VMS, visual cues of camouflage objects are considered less effective than motion cues. Prior works mainly relied on homography or optical flows to detect motion patterns. Bidau \etal proposed to segment moving objects from the environment by approximating different motion models computed from dense optical flow~\cite{bideau2016s, bideau2018moa}. In particular, in~\cite{bideau2016s} authors proposed a two-step segmentation algorithm, which first compensated for the camera rotation and then segmented the angle of the optical flow into objects and the background. Although each motion model is updated with optical flow orientations over time, the initial motion is heuristic. In~\cite{bideau2018moa}, authors used a  network to segment the angle field rather than raw optical flow. \cite{lamdouar2020betrayed} proposed a video registration and motion segmentation framework, along with a larger camouflaged dataset (MoCA) labeled by bounding boxes for every five frames. %$5^\text{th}$ frame. 
The explicit alignment method by optical flow builds spatial correspondence between neighboring frames. However, the optical flow estimation may not be accurate enough to support effective alignment, particularly in dynamic scenes with fast object motions.

% \vspace{-6pt}
\section{Proposed Framework}\label{sec:method}
% \vspace{-5pt}
%Here, we provide a detailed description of our framework. 
The input of our \Ourmodel~is a video clip containing camouflaged objects, and the output is a set of pixel-wise binary masks of the camouflaged objects for each frame in the video. Specifically, denote the video clip with $\mathrm{T}$ frames by $\{\mathbf{I}^\mathrm{t}\}_{t=1}^{\mathrm{T}}, \mathbf{I}^\mathrm{t}\in\mathbb{R}^{3 \times H \times W}$, where $H, W$ are the height and the width of the frame. Our network is to assign a binary mask $\mathbf{M}^\mathrm{t}\in \{0,1\}^{ H \times W}$ for the video frame $\mathbf{I}^\mathrm{t}$ at time $\mathrm{t}$. 

\subsection{Overview}
% \vspace{-5pt}
The overall framework of the \Ourmodel~ is shown in \figref{fig:overall}. The \Ourmodel~ consists of a short-term detection module and a long-term refinement module. The short-term detection module takes a pair of consecutive frames and predicts the camouflaged object mask for the reference frame.%\footnote{The result for the last frame can be achieved by swapping the reference frame for the last pair.}. 
A sequence-to-sequence translation module is adapted to jointly refine the input video clip frame results with temporal consistency priors. It takes $\mathrm{T}$ predictions from the short-term detection module as well as their corresponding referenced frames to generate the final prediction results. %\textcolor{red}{To the best of our knowledge, we are the first ones to formulate this dense prediction refining process as a sequence-to-sequence modeling problem ARE WE?}.
To the best of our knowledge, we are the first ones to formulate this dense prediction refining process as a sequence-to-sequence modeling problem.

To train the \Ourmodel, we adopt a two-stage strategy. 
We first train the short-term detection module using pixel-wise annotations only. Once the model converges, we attach the long-term refinement module to the \Ourmodel~and train the whole model while fixing the short-term detection module. 

\begin{figure*}[t!]
\begin{center}
\includegraphics[width=.98\textwidth]{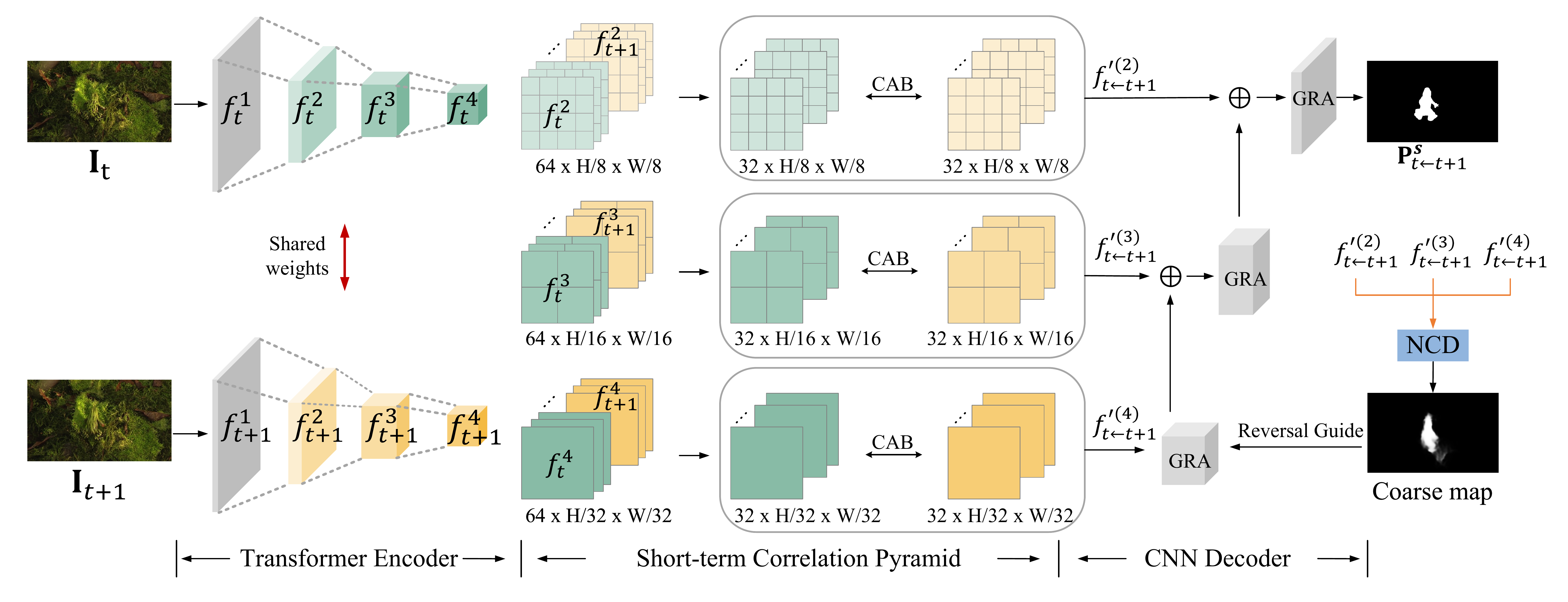}
\end{center}
\vspace{-20pt}
\caption{\small
The overview of our short-term pipeline. The network first extracts features from the input frames by a transformer encoder, then computes a full-range volumetric correspondence between the reference frame $\mathbf{I}_{t}$ and its neighboring frame $\mathbf{I}_{t+1}$ to form a correlation volume pyramid. A  CNN decoder is used to predict the final prediction from the motions captured by the short-term correlation pyramid.}
\vspace{-2mm}
\label{fig:short_term}
\end{figure*}

\subsection{Short-term Architecture}
% \vspace{-5pt}
We illustrate our short-term architecture in \figref{fig:short_term}.
It takes two consecutive frames as input from a video and predicts a binary mask of the reference frame. Our model consists of three main modules: (1) \textbf{Transformer Encoder} for feature extraction; (2) \textbf{Short-term Correlation Pyramid} for capturing short-term dynamics; and (3) \textbf{CNN Decoder} to predict the short-term segmentation. Below we describe the details of each module. 

%\paragraph{1. Transformer Encoder.}
\noindent\textbf{1. Transformer Encoder.}
We adopt a Siamese structure with the pyramid vision transformer (PVT)~\cite{wang2021pvtv2} to extract features from two consecutive frames. 
The encoder consists of four stages that generate feature maps at four different scales. All stages share a similar structure, including a patch embedding layer and transformer blocks. The sizes of the features at each stage are $ C_i \times H / 2^{i+1} \times W/2^{i+1} $, $i \in \{1, 2, 3, 4\}$, where the $H,W,C$ represent the height, the width and the channels. We set $C=32$ in our experiments. Following \cite{fan2021concealed}, we adapt three texture enhanced modules (TEM) for the features from the last three stages. To attain more discriminative feature representations, each TEM includes four parallel residual branches. 

%\paragraph{2. Short-term Correlation Pyramid.} 
\noindent\textbf{2. Short-term Correlation Pyramid.}
Prior works (\eg, \cite{tokmakov2017learning, yang2021selfsupervised}) explicitly incorporate motion by taking optical flow from consecutive frames as the inputs into a deep network. However, the inaccurate optical flow may result in error accumulation at subsequent predictions. If we would like to optimize the optical flow module with the segmentation module jointly, the ground truth of optical flow is required. To solve this issue, inspired by~\cite{li2021arvo}, we propose a correlation pyramid to capture motion information implicitly. As shown in Figure~\ref{fig:short_term}, the CNN decoder directly takes the correlation pyramid as its only input. It means the network can only estimate correct segmentation with correct motion estimation. Also, since the features used to form the correlation pyramid will be updated with the segmentation ground truth, we can use the segmentation ground truth to optimize motion estimations and detection results jointly.

\begin{figure}[b!]
\vspace{-10pt}
\begin{center}
\includegraphics[width=0.35\textheight]{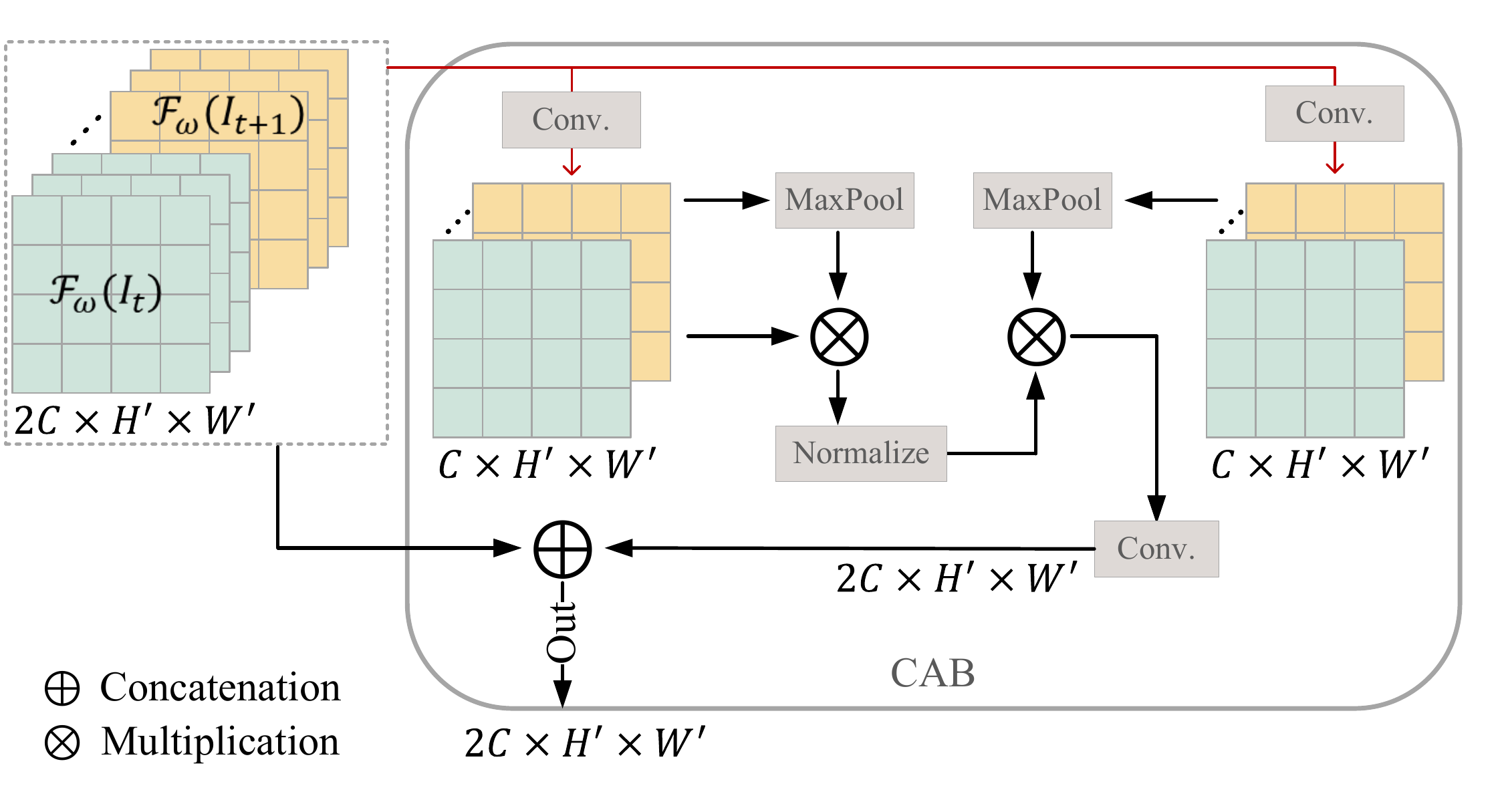}
\end{center}
\vspace{-20pt}
\caption{Correlation aggregation block (CAB) computes the normalized correlation volume of feature maps between the reference frame (green blocks) and the neighboring frame (yellow blocks).}
\label{fig:all_range}
\end{figure}

\begin{figure*}[t!]
\begin{center}
\includegraphics[width=.8\textwidth]{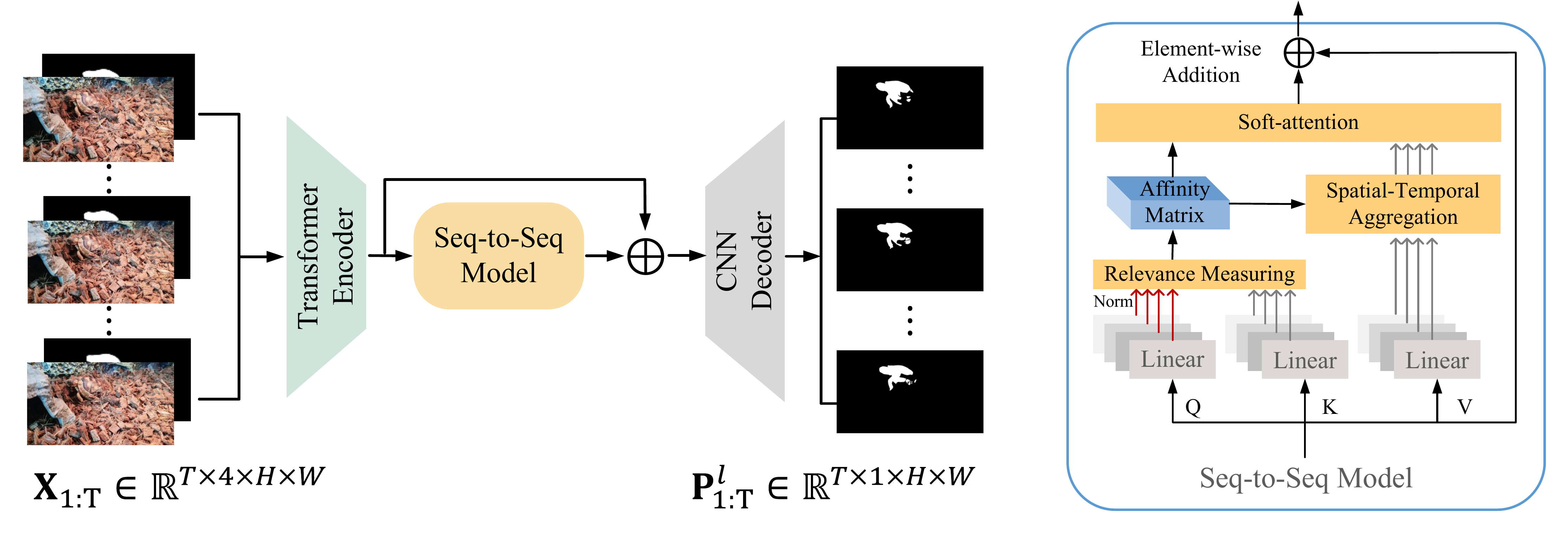}
\end{center}
\vspace{-20pt}
\caption{The overview of the proposed long-term consistency architecture. It formulates the process as a seq-to-seq modeling problem and refines the pair-wise predictions with a sequence-to-sequence transformer.}
\vspace{-2mm}
\label{fig:long_term}
\end{figure*}

We illustrate the core unit of our correlation pyramid, namely correlation aggregation block (CAB) $\mathbf{C}$, in \figref{fig:all_range}.
%As shown in \figref{fig:all_range}, a correlation aggregation block (CAB) $\mathbf{C}$ is defined as the core unit of the short-term correlation pyramid. It allows us to find correspondences at a global scale. 
Given a pair of frame features$\{ f_{t}, f_{t+1} \} \in  \mathbb{R}^{C \times H' \times W'}$, the 4D correlation volume $\mathbf{C}(\mathbf{I}_t,\,\mathbf{I}_{t+1}) \in \mathbb{R}^{ H' \times W' \times H' \times W' }$ is defined as:
\begin{equation}
\small
\textstyle{
 \mathbf{C}(\mathbf{I}_t,\mathbf{I}_{t+1})_{xyuv} = \exp\Big(\sum_{c} \mathcal{F}_\theta(\mathbf{I}_{t})_{xyc} \cdot \mathcal{F}_\theta(\mathbf{I}_{t+1})_{uvc}\Big),
 \label{ag:correlation}
}
\end{equation}
with $c$ being the index along the channel dimension of frame features. With all neighboring features are paired up with correlations, we can find correspondences at a global scale. To reduce the computational complexity, 
we downsample the adjacent frame by max-pooling over features while keeping the resolution of the reference frame. This design helps the model to learn multi-scale displacement while maintaining high-resolution image details. 

Next, we normalize the feature correlation volume $\mathbf{C}(\mathbf{I}_t,\,\mathbf{I}_{t+1})_{xyuv}$ along the last two dimensions $uv$ over their sum, as they represent the correspondence between the reference and downsampled neighboring feature frame in all the spatial position. The normalized correlation volume is computed as follows:
\begin{equation}
% \vspace{-5pt}
\textstyle{
    \Tilde{\mathbf{C}}(\mathbf{I}_t,\,\mathbf{I}_{t+1})_{xyuv} = \frac{\mathbf{C}(\mathbf{I}_t,\,\mathbf{I}_{t+1})_{xyuv}}
    {\sum\limits_u\sum\limits_v{\mathbf{C}}(\mathbf{I}_t,\mathbf{I}_{t+1})_{xyuv}}.
}
% \vspace{-5pt}
\end{equation}

We apply a convolution operation $\phi(\cdot)$ to selectively consider channel-wise information, and thus obtain a refined feature map $\phi(\mathbf{I}_{t+1}) \in\mathbb{R}^{C \times H' \times W'}$. Specifically, the aggregated features $f'_{t\leftarrow t+1}= \rho (\mathbf{I}_{t\leftarrow t+1})\in\mathbb{R}^{C \times H' \times W'}$ was computed as follows:
\begin{equation}
% \vspace{-5pt}
    \rho(\mathbf{I}_{t\leftarrow t+1})=\Tilde{\mathbf{C}}(\mathbf{I}_t,\mathbf{I}_{t+1})\phi(\mathbf{I}_{t+1}).
% \vspace{-5pt}
\end{equation}

\figref{fig:all_range} only shows a correlation on one scale. To make the network learn more detailed information, we construct a correlation pyramid $\{\mathbf{C}^{i}\}, i \in \{2, 3, 4\}$ by incorporating the extracted multi-scale features from the transformer encoder (See details in \supp{supplementary materials (Supp)} ).

%\paragraph{3. CNN Decoder.}
\noindent\textbf{3. CNN Decoder.}
As shown  by \cite{fan2021concealed}, the neighbor connection decoder is more reliable than conventional connection decoder (\ie, densely connection or short connection). In addition, the group-reversal attention (GRA) strategy used in \cite{fan2021concealed} can provide more accurate segmentation results around the object boundaries. 
Based on these, we directly feed features from the short-term correlation pyramid, \ie, $\{f'^{(i)}_{t \leftarrow t+1} \} \in \mathbb{R}^{C \times H/2^{i+1} \times W/2^{i+1}}, i \in \{2, 3, 4\}$ into the GRA blocks, and generate  refined feature maps. The neighbor connection decoder (NCD) is used to generate a coarse map, which could provide reversal guidance of rough location of the camouflaged object. In this way, we gather the low-level features from the CNN decoder and the high-level features from the correlation pyramid.

%\paragraph{Learning Strategy.}
\noindent\textbf{Learning Strategy.}
We train the short-term training stage by minimizing the loss below: 
\begin{equation}
% \vspace{-5pt}
    \mathcal{L} = \mathcal{L}^{w}_{ce} + \mathcal{L}^{w}_{iou}.
% \vspace{-5pt}
\end{equation}
The weighted cross-entropy loss $\mathcal{L}^{w}_{ce}$ increases the weights of hard pixels to emphasize their importance. The weighted intersection-over-union loss $\mathcal{L}^{w}_{iou}$ pays more attention to hard pixels rather than assigning all pixels with equal weights. Readers could refer to prior work \cite{wei2020f3net} to find more details regarding the definitions of these two loss functions. 

\subsection{Long-term Consistency Architecture}
% \vspace{-5pt}
%To encourage long-term temporal consistency, we introduce a refinement network with the spatial-temporal information to generate final predictions.
Given a sequence of $\mathbf{I}_{1:T} = \{ \mathbf{I}_1, \mathbf{I}_2, \dots, \mathbf{I}_T \}$ and the pixel-wise predictions of $\mathbf{P}^{s}_{1:T} = \{ \mathbf{P}^{s}_1, \mathbf{P}^{s}_2, \dots, \mathbf{P}^{s}_T \}$ from our short-term architecture, we formulate the long-term consistency refinement process as a seq-to-seq problem. 
\figref{fig:long_term} illustrates the long-term consistency architecture. We use the same backbone as the short-term architecture, \ie, transformer encoder and CNN decoder modules, since it has been already pre-trained on camouflaged datasets that could largely accelerate the long-term training processing. 
For each frame of the input sequence, we concatenate the color frame $\mathbf{I}_t$ with its corresponding prediction $\mathbf{P}^s_t, t \in [1:T]$ on the channel dimension, and then stack every concatenated frame within the sequence to form a 4D tensor $\mathbf{X}_{1:T} \in \mathbb{R}^{ T \times 4 \times H \times W } $. The network takes $\mathbf{X}_{1:T}$ as the input and output the final prediction sequence $\mathbf{P}^{l}_{1:T} \in \mathbb{R}^{ T \times 1 \times H \times W}$.

There are two kinds of seq-to-seq modeling architecture: one uses convLSTM to model the temporal information, and the other uses a transformer-based seq-to-seq modeling network. We implement both architectures and compare their results in Section \ref{sec:ablation}. We empirically find that using the transformer structure can lead to better results, so we select it as our seq-to-seq modeling network to enforce the long-term consistency.

We show the details of the seq-to-seq modeling network on the right side of \figref{fig:long_term}. For each target pixel, to reduce the complexity for building a dense spatial-temporal affinity matrix, we select a fixed number of relevance measuring blocks to construct the affinity matrix within a constrained neighborhood of it.
We apply the hybrid loss \cite{21Fan_HybridLoss} during the training: 
\begin{equation}
    \mathcal{L}_{hybrid} = \mathcal{L}^{w}_{ce} + \mathcal{L}^{w}_{iou} + \mathcal{L}_{e},
\end{equation}
where $\mathcal{L}_{e}$ is the Enhanced-alignment loss, the hybrid loss can guide the network to learn pixel-, object- and image-level features.

\begin{table*}[t!]
  \footnotesize
  \centering
  \caption{Quantitative results on our MoCA-Mask with (w/) and without (w/o) our pseudo labels. The best performing method of each category is highlighted in \textbf{bold}. 
  Noting that MG~\cite{yang2021selfsupervised} performs unsupervised learning that are trained without labels.} 
  \label{tab:Moca}
  \vspace{-5pt}
  \tabcolsep=0.3cm
  \renewcommand{\arraystretch}{0.6}
%   \vspace{-5pt}
  \begin{tabular}{r|cccccccccccc } 
  \toprule
  & \multicolumn{6}{c}{MoCA-Mask w/o pseudo labels} & \multicolumn{6}{c}{w/ pseudo labels} \\
  \cmidrule(lr){2-7}
  \cmidrule(lr){8-13}
  Models & $S_\alpha\uparrow$ &$F_\beta^w\uparrow$ &$E_\phi\uparrow$ &$M\downarrow$ & mDic & mIoU
  & $S_\alpha\uparrow$ &$F_\beta^w\uparrow$ &$E_\phi\uparrow$ &$M\downarrow$ & mDic & mIoU \\
  \cmidrule(lr){1-7}
  \cmidrule(lr){8-13}
  {EGNet} \cite{zhao2019EGNet} & 0.547 & 0.110 & 0.574 & 0.035 & 0.143 & 0.096 & 0.546 & 0.105 & 0.573  & 0.034 & 0.135 & 0.090 \\
  {BASNet} \cite{Qin_2019_CVPR}  & 0.561 & 0.154 & 0.598 & 0.042 & 0.190 & 0.137 & 0.537  & 0.114 & 0.579 & 0.045 & 0.135 & 0.100  \\
  {CPD} \cite{Wu_2019_CVPR} & 0.561 & 0.121 & 0.613 & 0.041 & 0.162 & 0.113 & 0.550 & 0.117 & 0.613 & 0.038 & 0.147 & 0.104 \\
  {PraNet} \cite{fan2020pra} & 0.614 & 0.266  & 0.674 & 0.030 & 0.311 & 0.234 & 0.568 & 0.171 & 0.576 & 0.045 & 0.211 & 0.152 \\
  {SINet}  \cite{fan2020Camouflage}  & 0.598 & 0.231 & 0.699 & 0.028 & 0.276 & 0.202  & 0.574 & 0.185  & 0.655 & 0.030 &  0.221 &  0.156 \\
  {SINet-v2} \cite{fan2021concealed}  & 0.588 & 0.204 & 0.642 & 0.031 & 0.245 & 0.180 & 0.571 & 0.175 & 0.608 & 0.035 & 0.211 & 0.153  \\

  \cmidrule(lr){1-7}
  \cmidrule(lr){8-13}
  {PNS-Net} \cite{ji2021progressively} & 0.544 & 0.097 &  0.510 & 0.033 & 0.121 & 0.101 & 0.576 & 0.134 & 0.562 & 0.038  & 0.189  & 0.133 \\
  {RCRNet} \cite{yan2019semi} & 0.555 &  0.138 &  0.527 &  0.033 &  0.171 & 0.116 & 0.597 & 0.174 & 0.583 & 0.025 & 0.194 & 0.137 \\
  {MG} \cite{yang2021selfsupervised}  &  0.530 & 0.168  & 0.561 &  0.067  & 0.181  & 0.127  & 0.547 & 0.165 &  0.537 &  0.095 & 0.197 & 0.141 \\

  \textbf{\Ourmodel~(Ours)} & \textbf{0.631} &  \textbf{0.311} &  \textbf{0.759} & \textbf{0.027} & \textbf{0.360} & \textbf{0.272} & \textbf{0.656} &  \textbf{0.357} & \textbf{0.785}  & \textbf{0.021}  & \textbf{0.397}  &  \textbf{0.310} \\
  \bottomrule
  \end{tabular}
\end{table*}

% \vspace{-10pt}
\section{Experiments}\label{sec:exp}

\begin{figure*}[t!]
\vspace{-10pt}
\small
    \centering
    \tabcolsep=0.02cm
    \renewcommand{\arraystretch}{1.0}
    \begin{tabular}{c c c c c c c c}
    \rotatebox{90}{~arctic fox} &
    \includegraphics[width=0.138\linewidth]{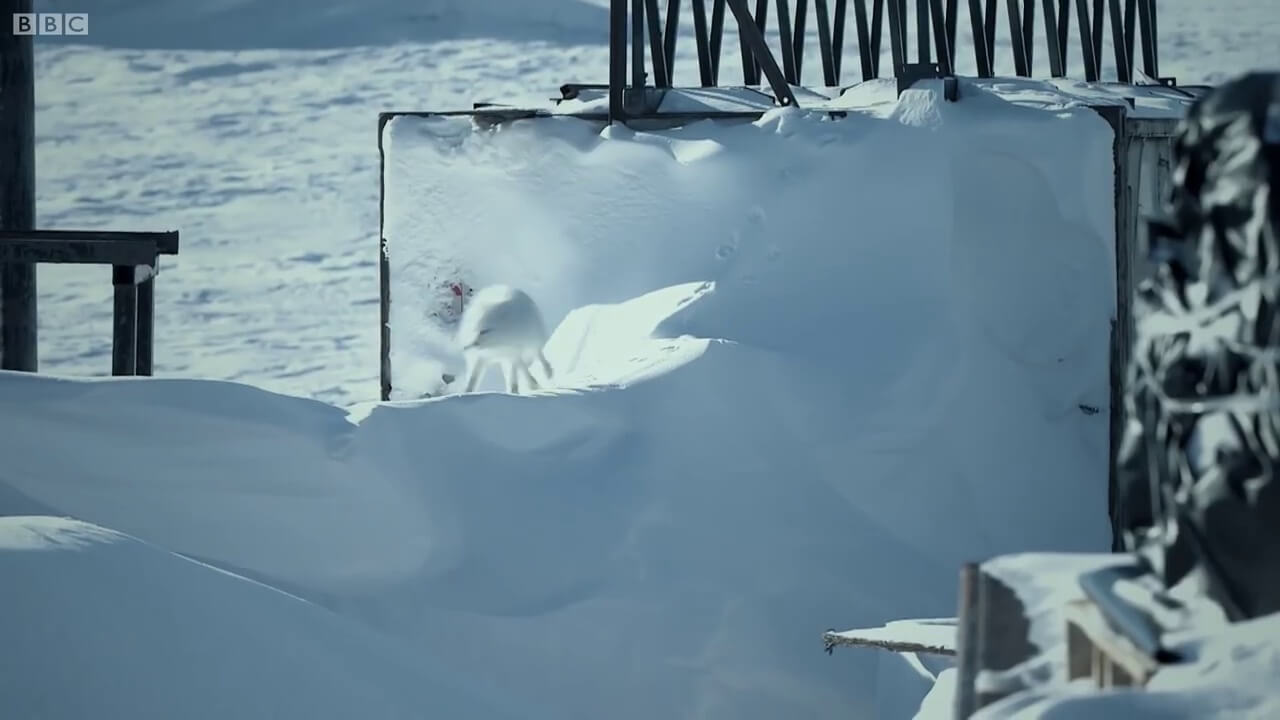} & 
    \includegraphics[width=0.138\linewidth]{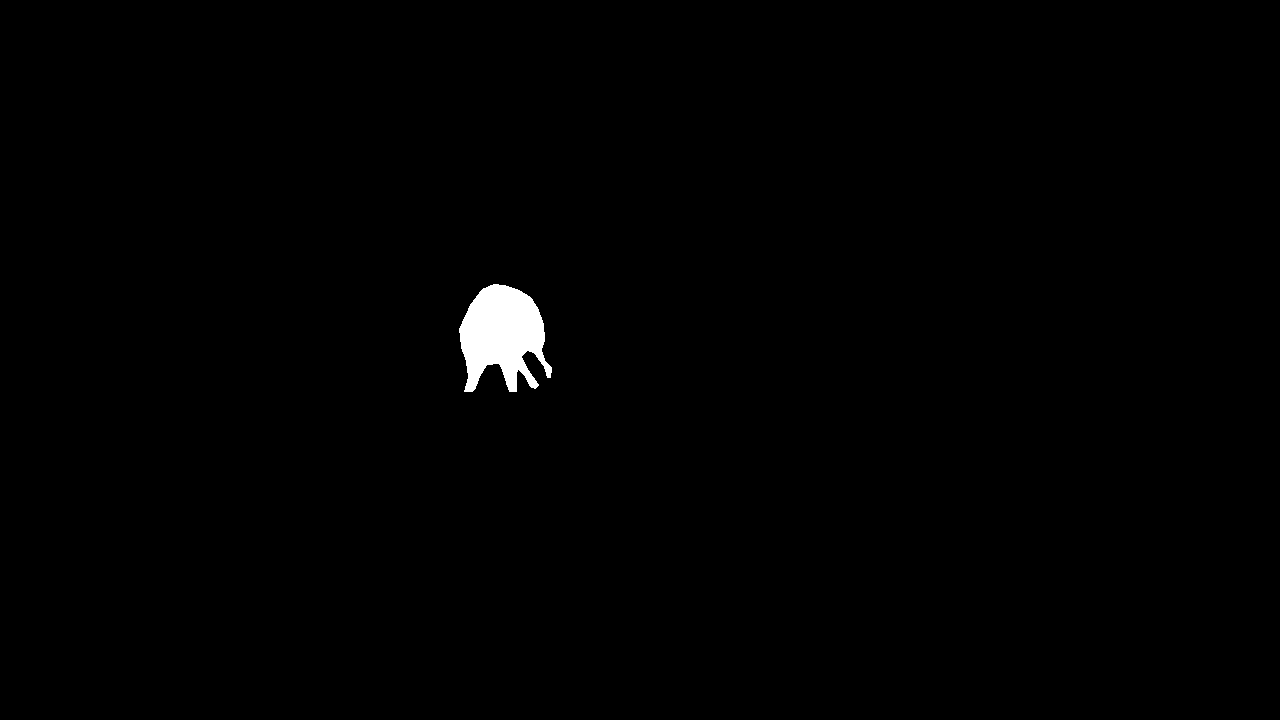} & 
    \includegraphics[width=0.138\linewidth]{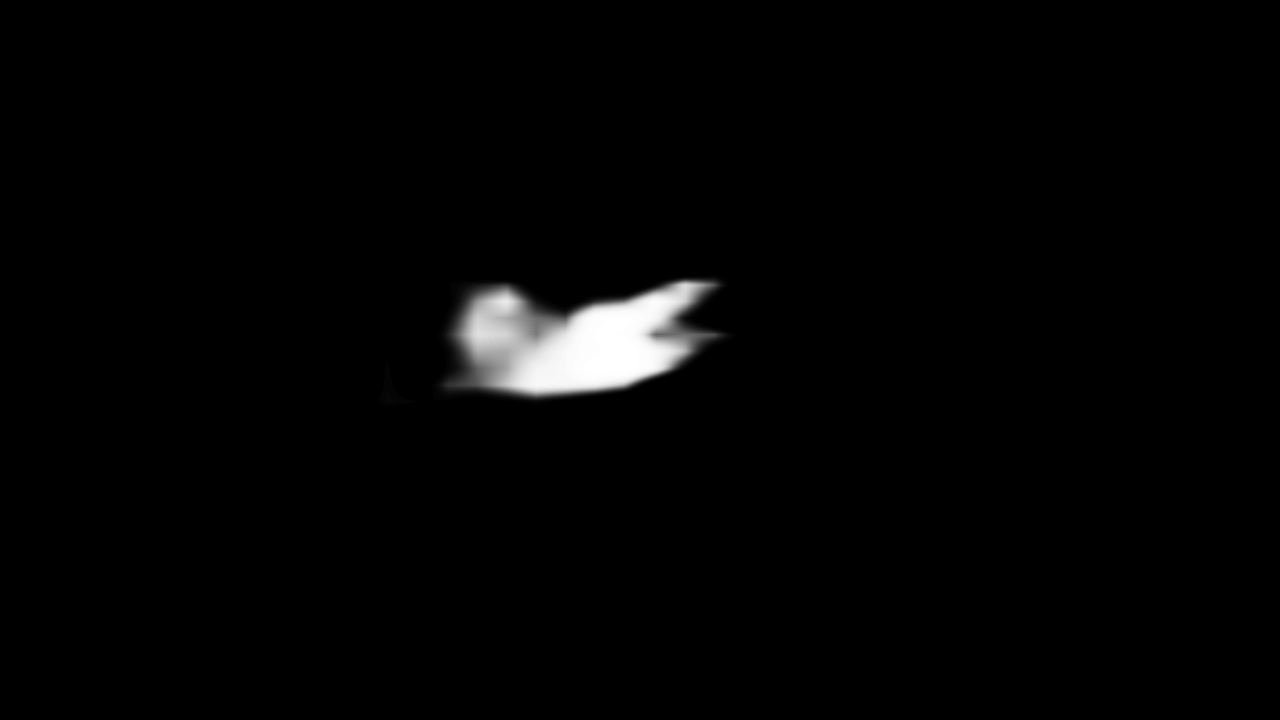} & 
    \includegraphics[width=0.138\linewidth]{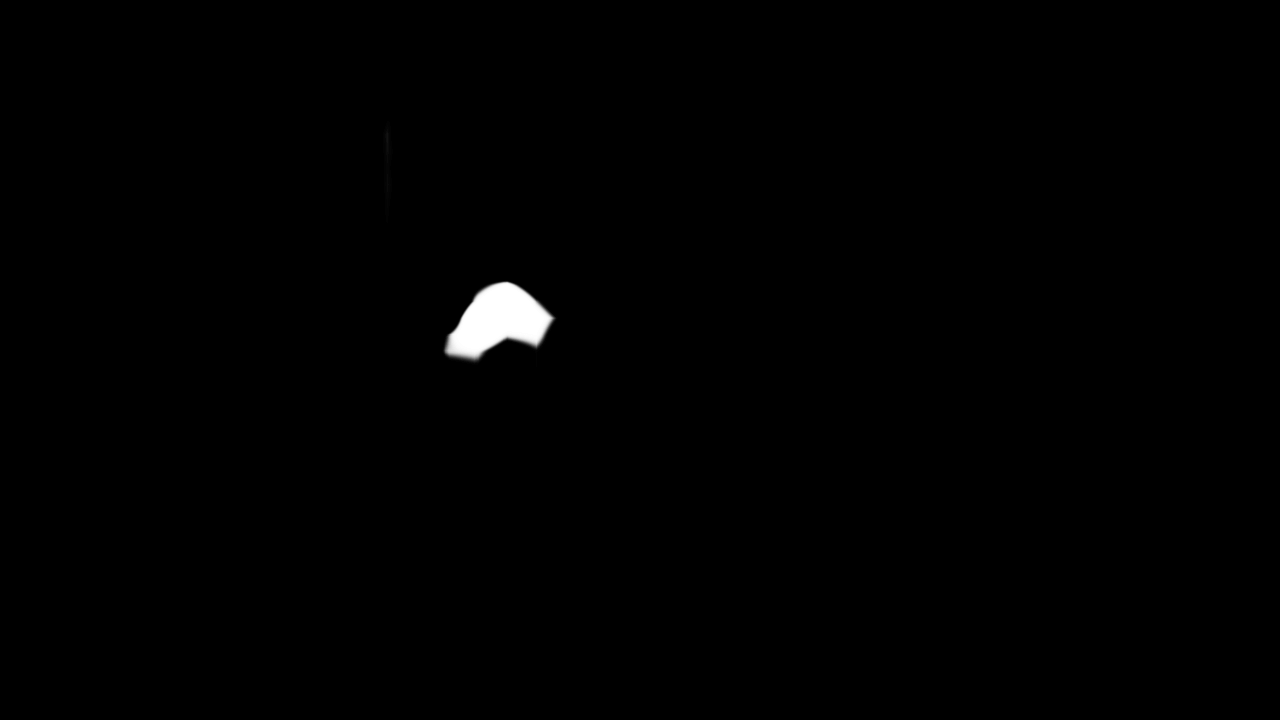} & 
    \includegraphics[width=0.138\linewidth]{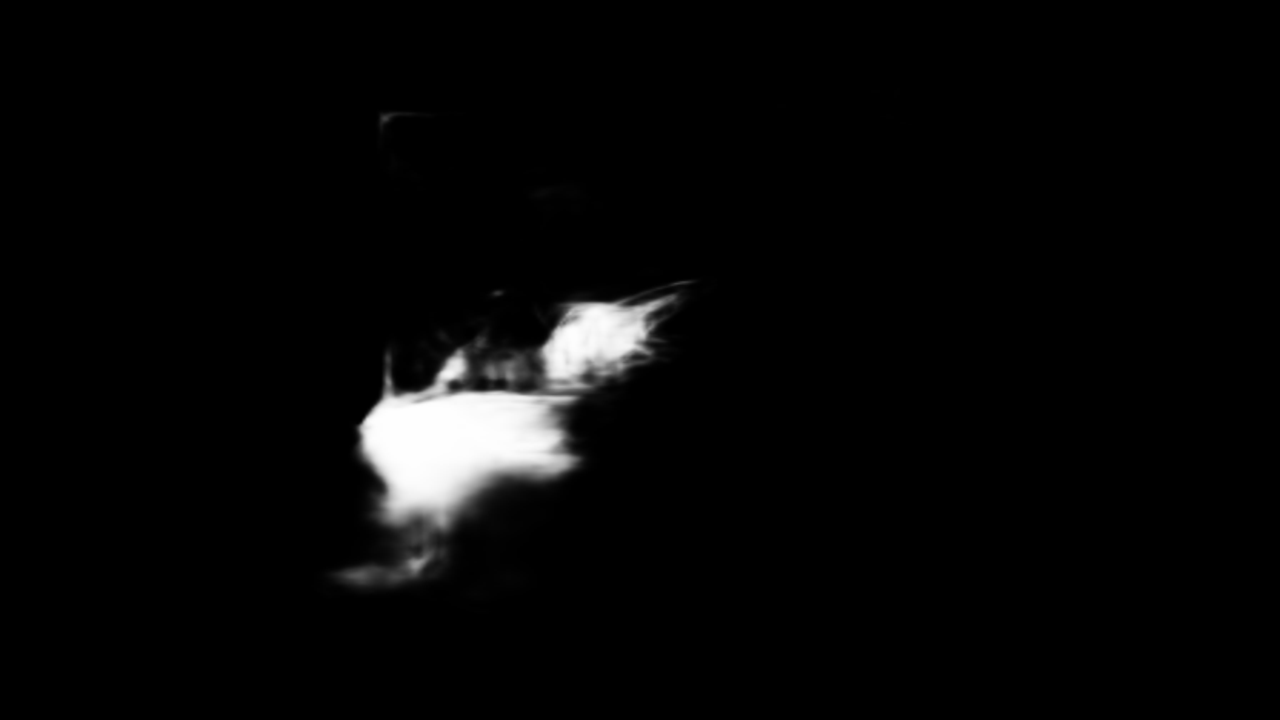} &
    \includegraphics[width=0.138\linewidth]{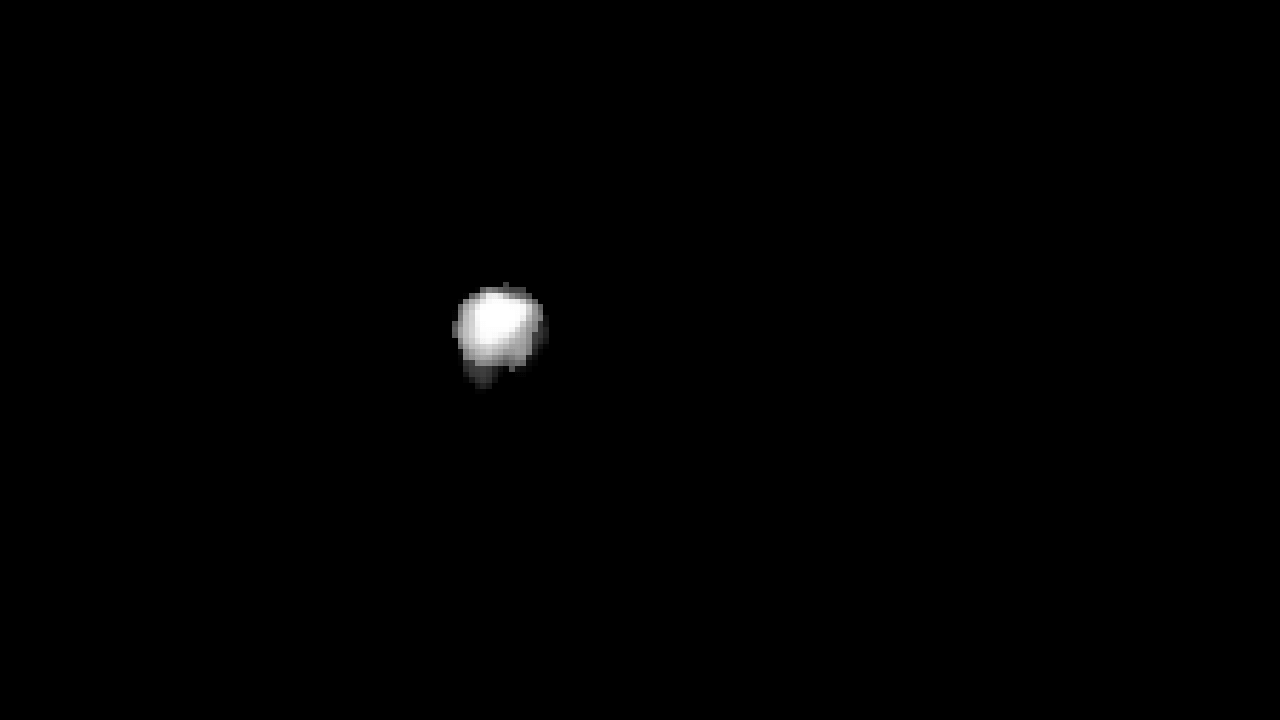} & 
    \includegraphics[width=0.138\linewidth]{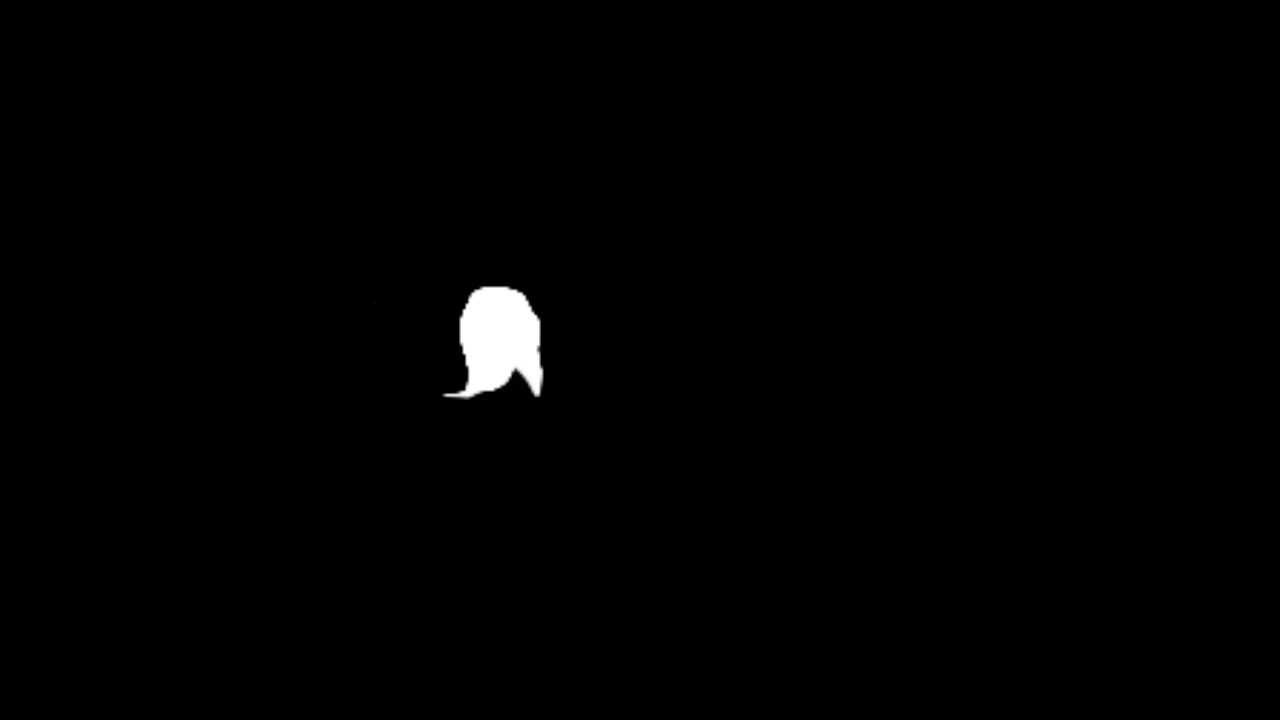}
    \\        
    \rotatebox{90}{~sand cat} &
    \includegraphics[width=0.138\linewidth]{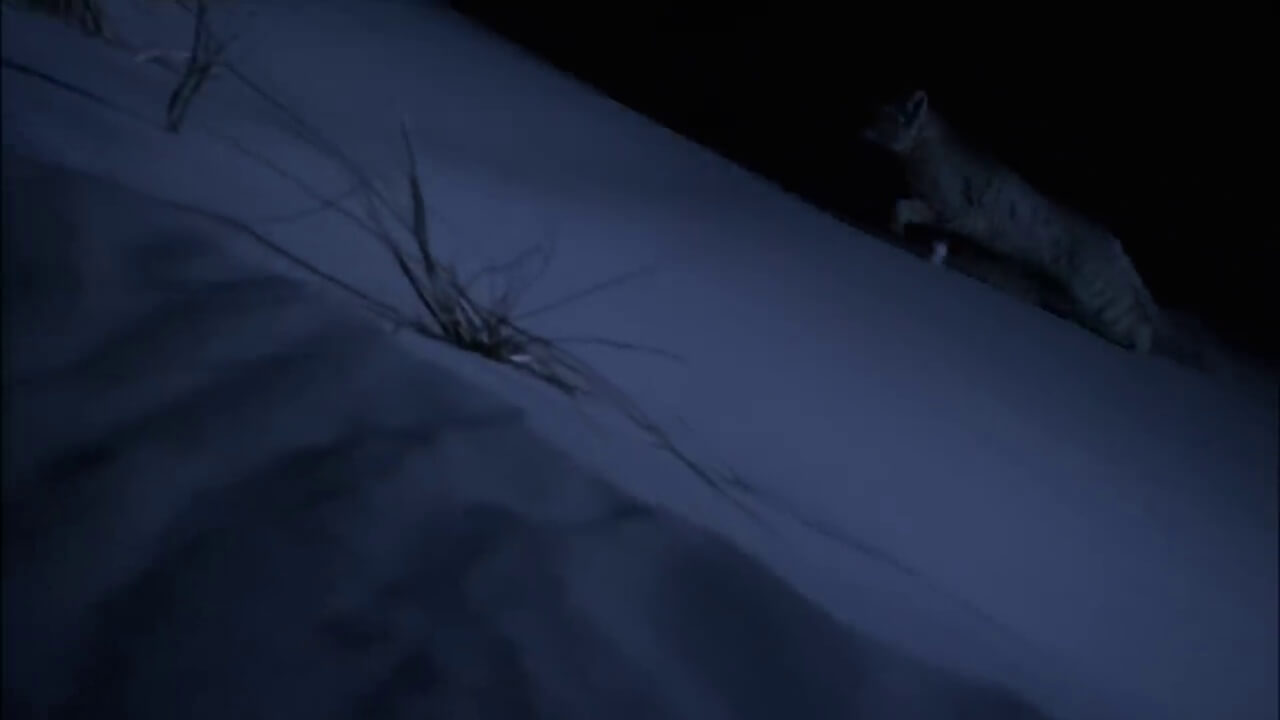} & 
    \includegraphics[width=0.138\linewidth]{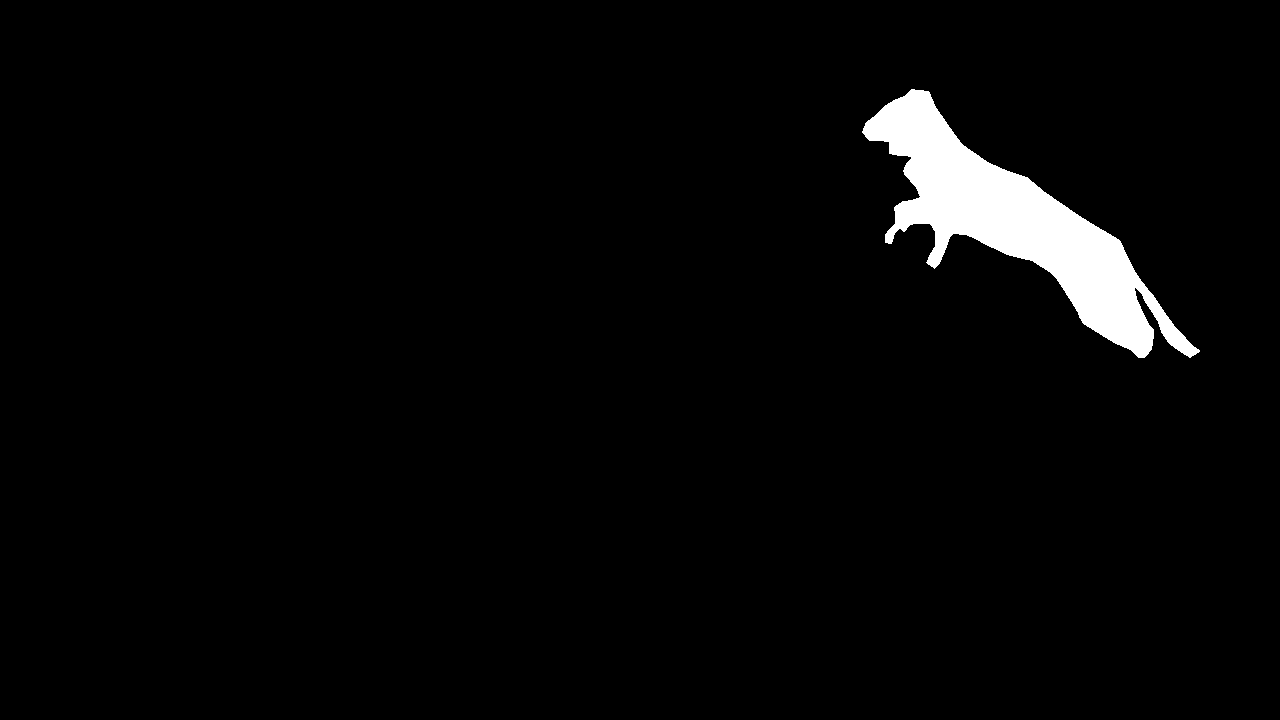} & 
    \includegraphics[width=0.138\linewidth]{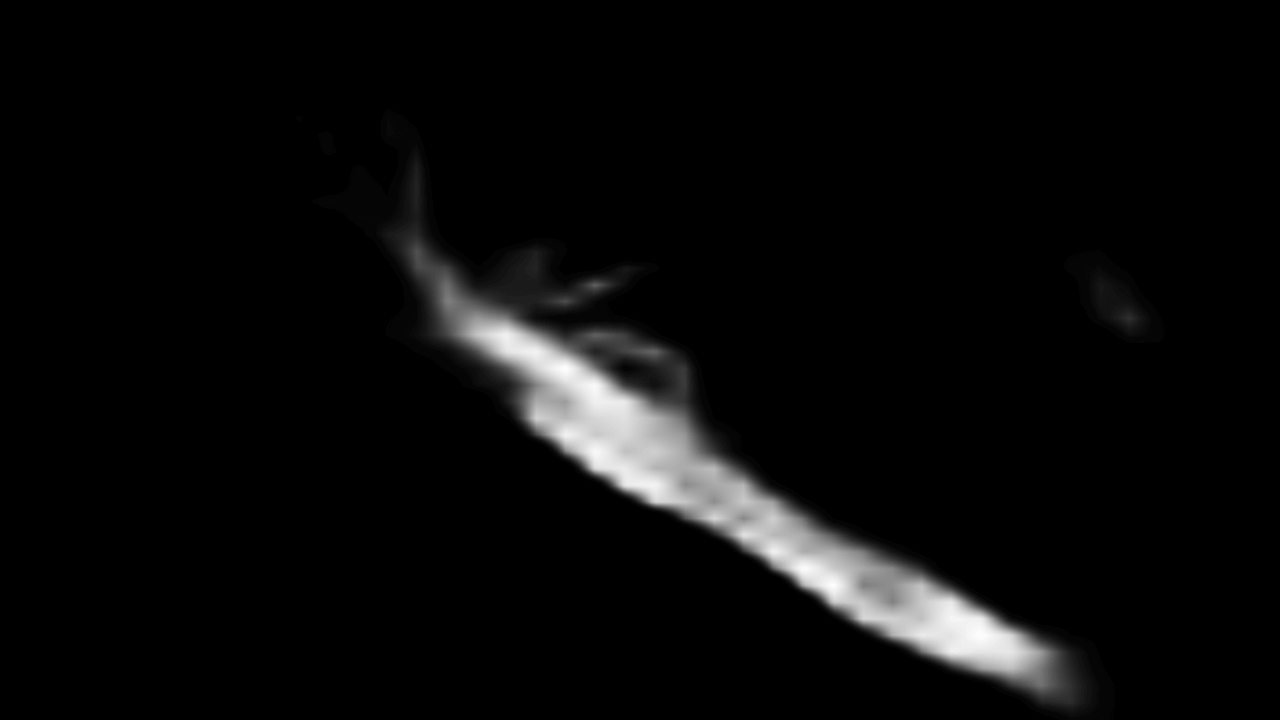} & 
    \includegraphics[width=0.138\linewidth]{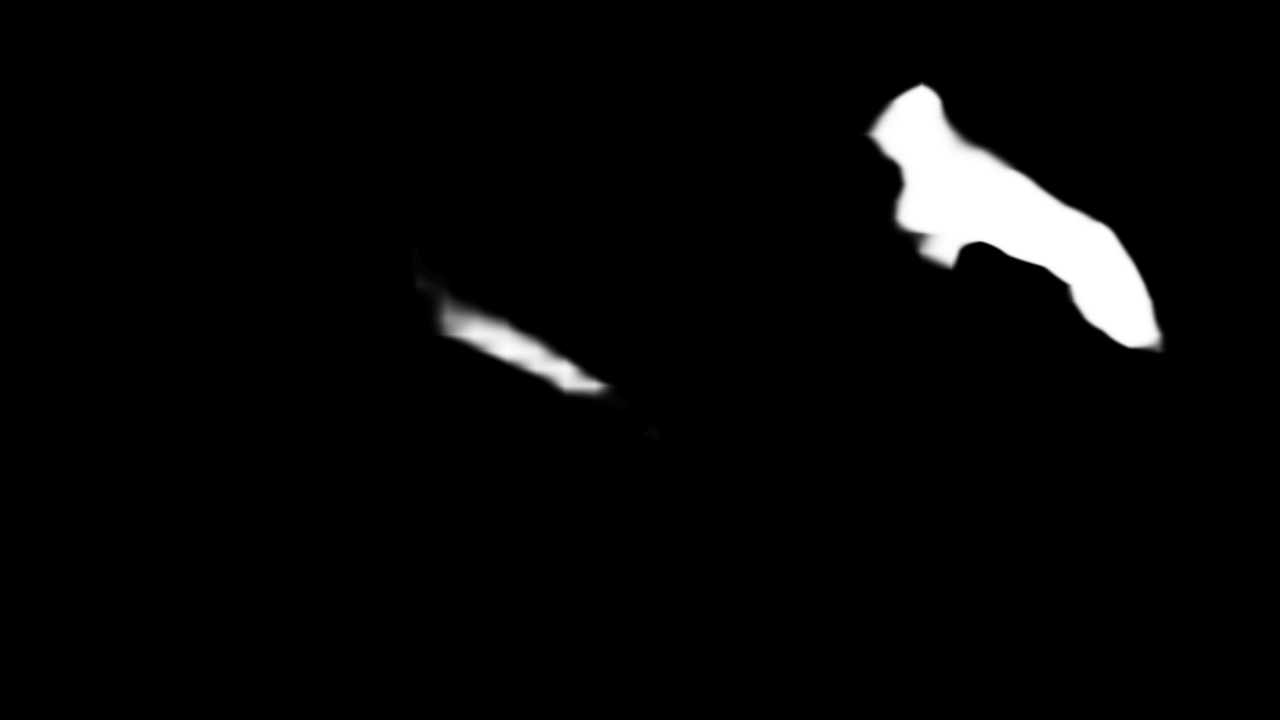} & 
    \includegraphics[width=0.138\linewidth]{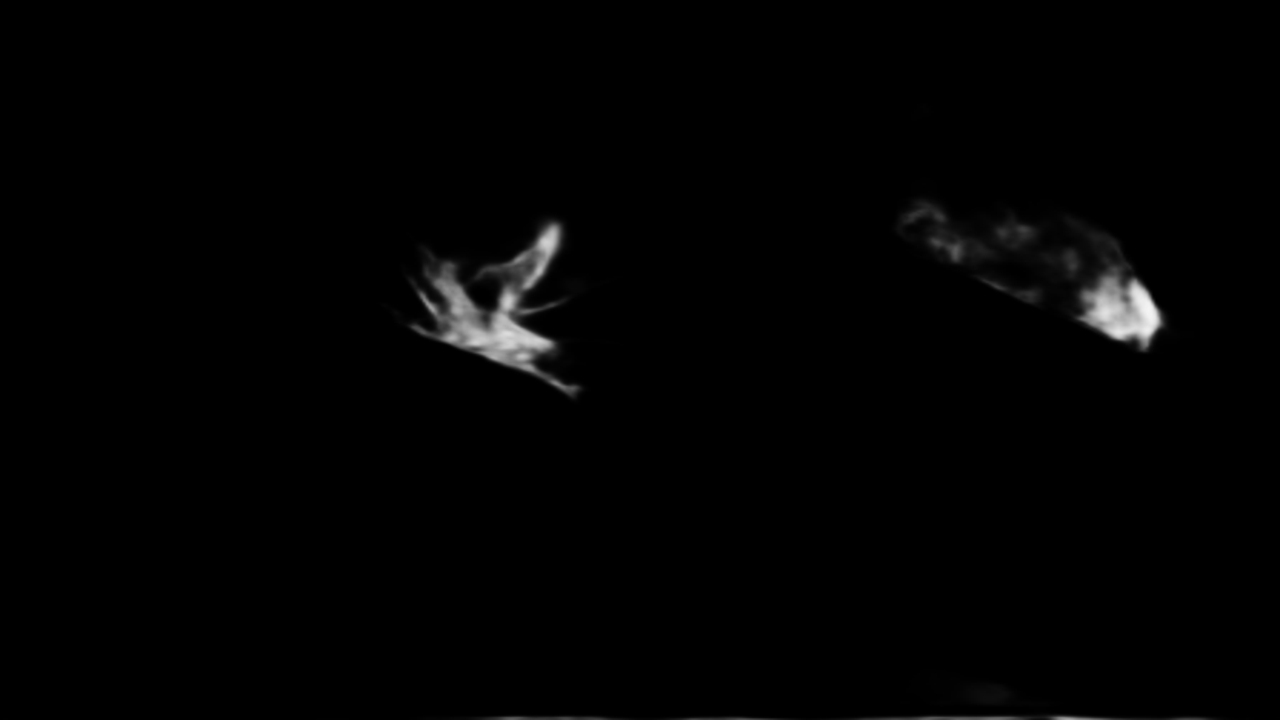} &
    \includegraphics[width=0.138\linewidth]{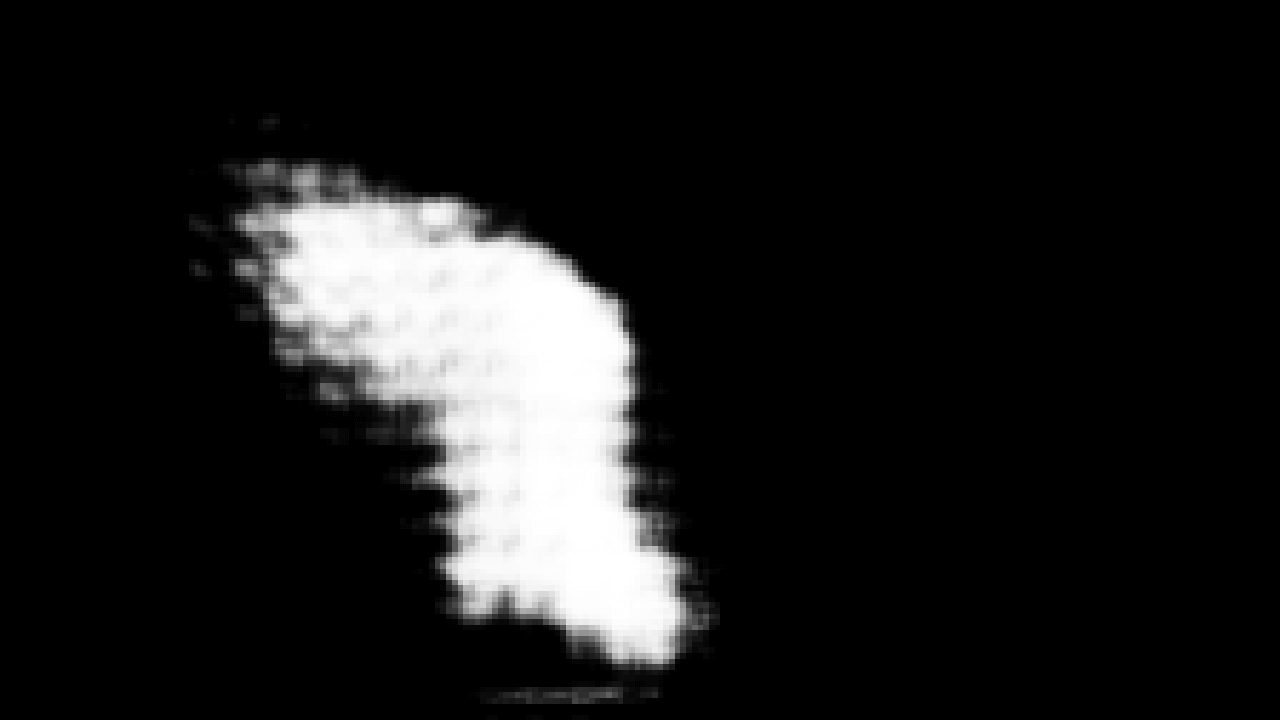} & 
    \includegraphics[width=0.138\linewidth]{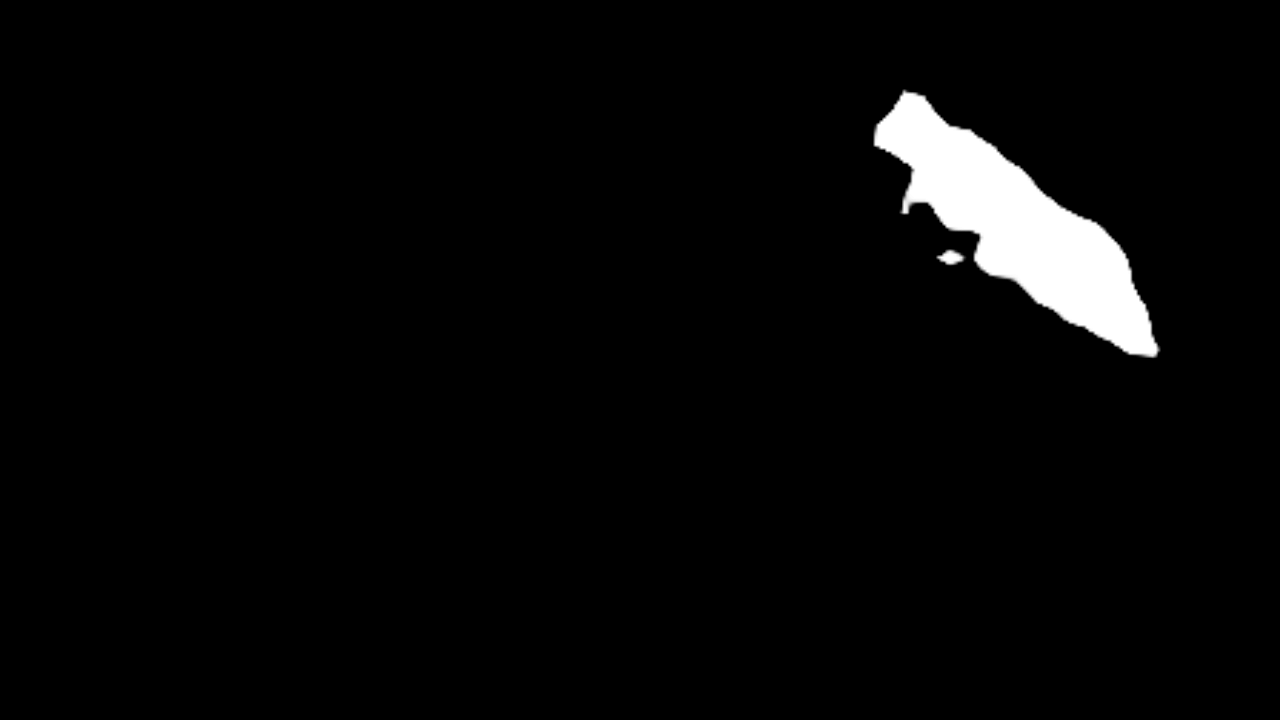}
    \\       
    \rotatebox{90}{~\quad ibex} &    
    \includegraphics[width=0.138\linewidth]{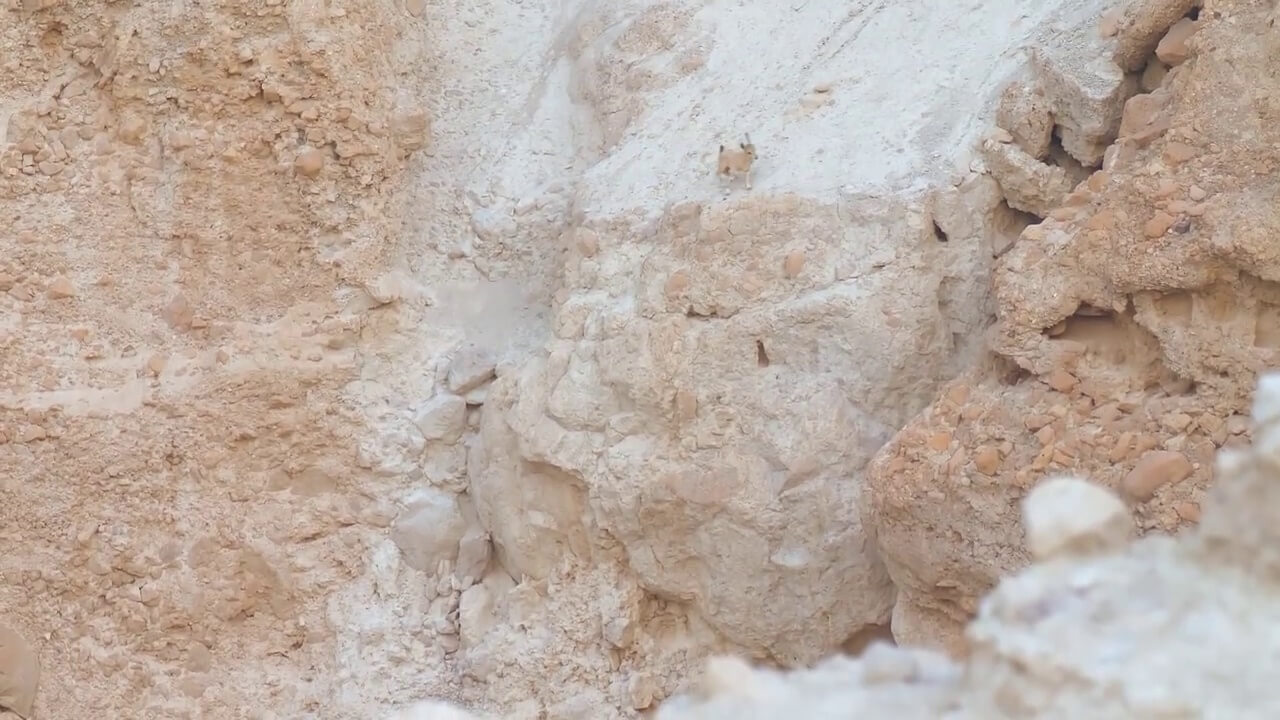} & 
    \includegraphics[width=0.138\linewidth]{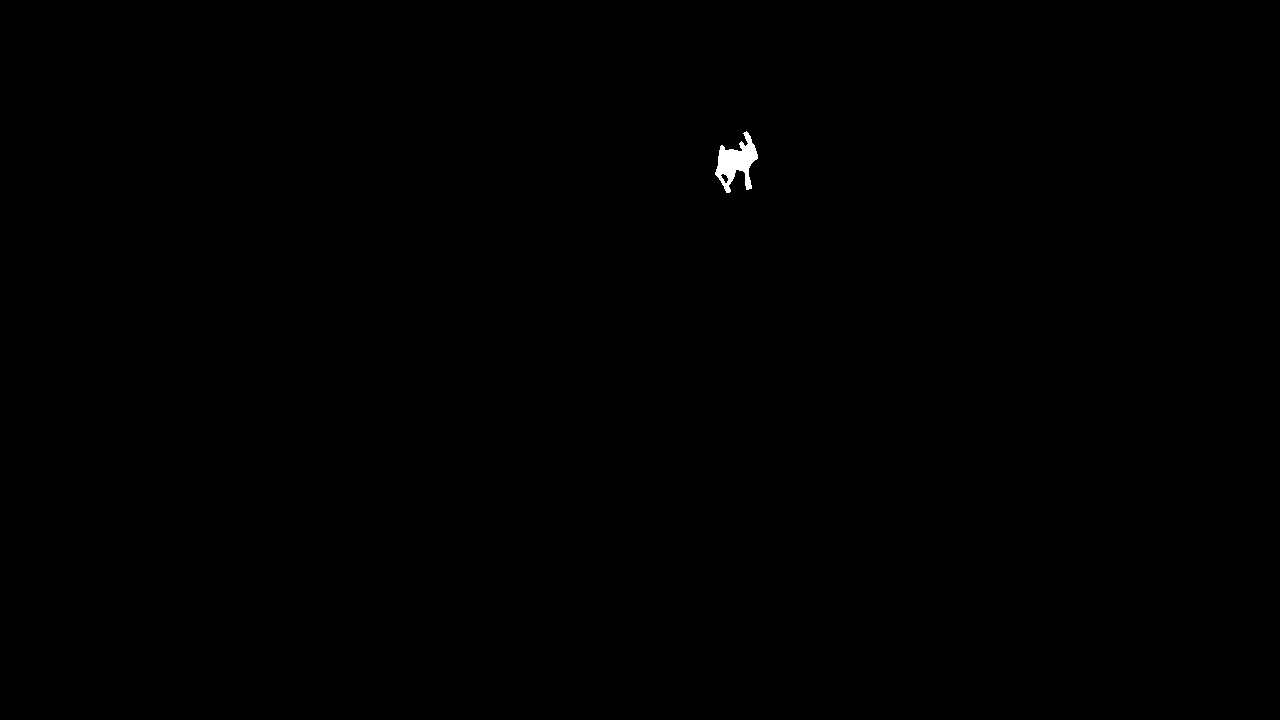} & 
    \includegraphics[width=0.138\linewidth]{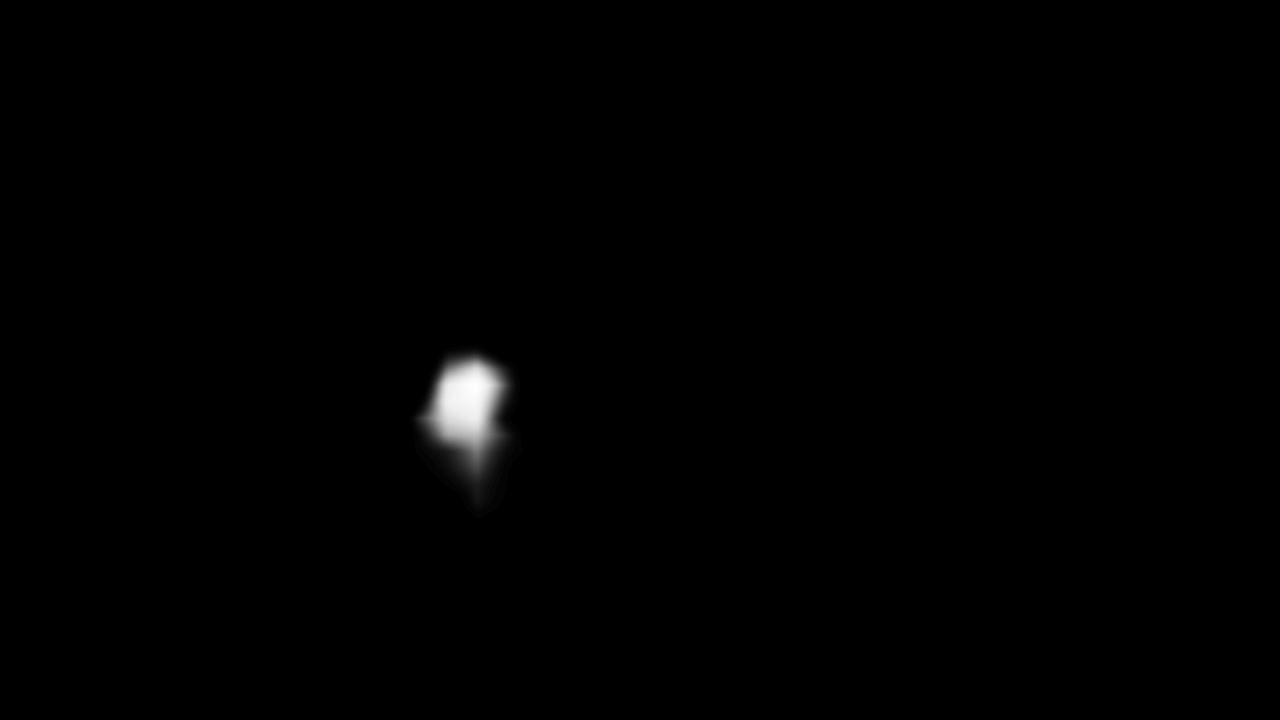} & 
    \includegraphics[width=0.138\linewidth]{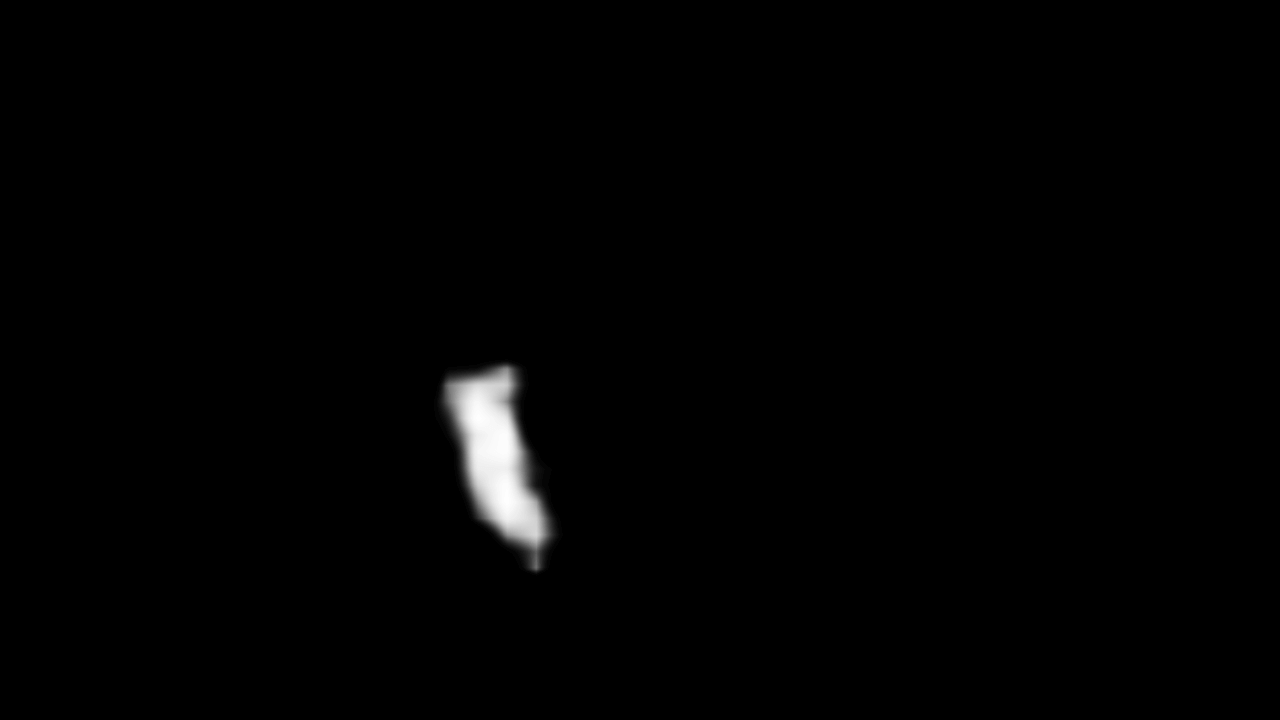} & 
    \includegraphics[width=0.138\linewidth]{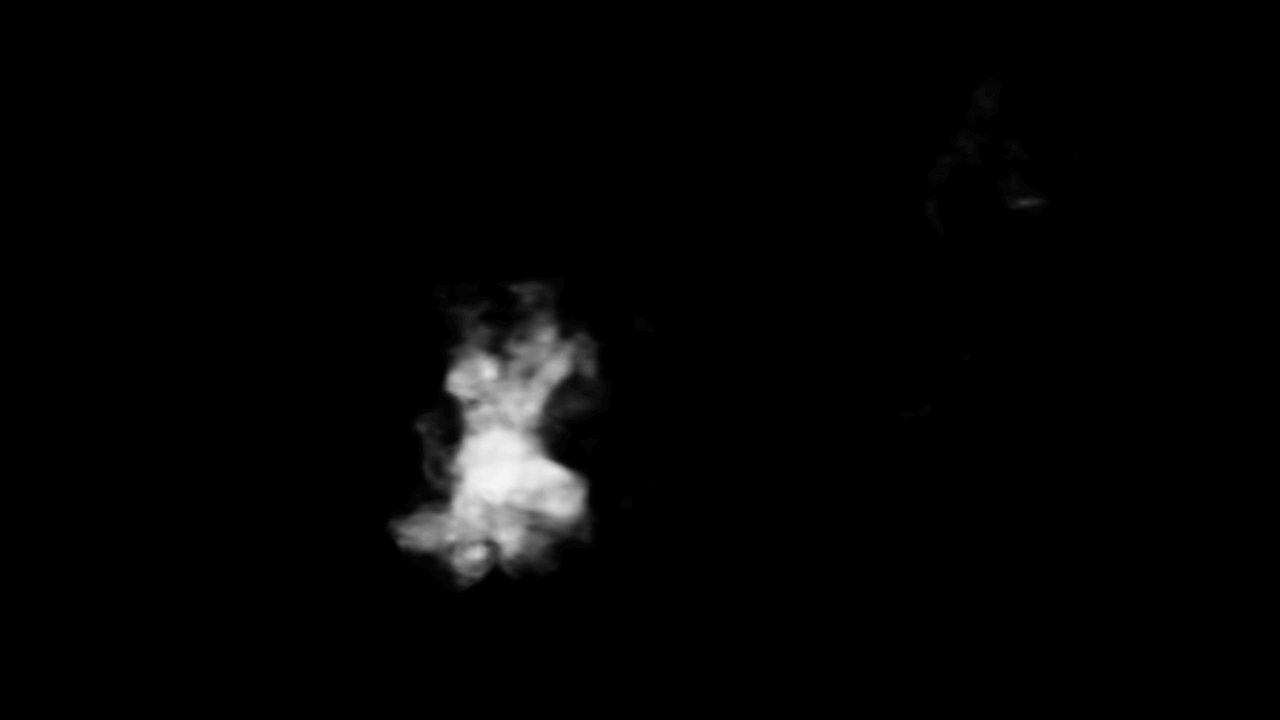} &
    \includegraphics[width=0.138\linewidth]{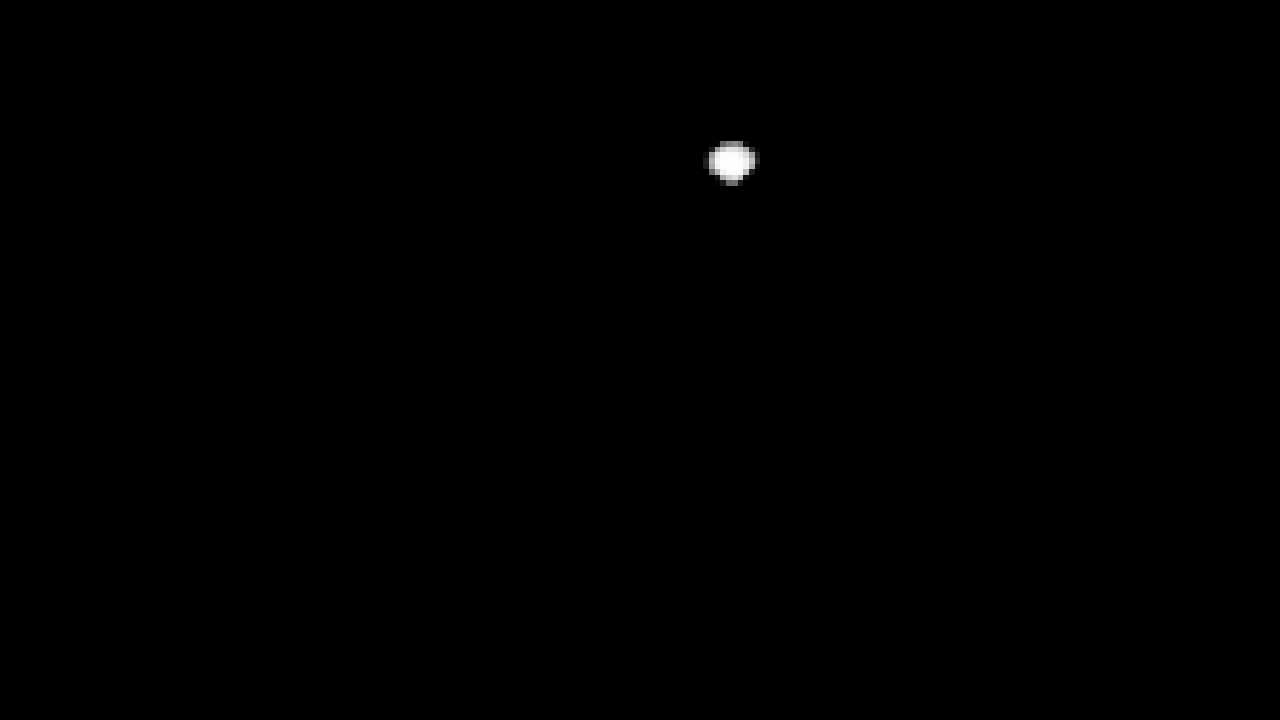} & 
    \includegraphics[width=0.138\linewidth]{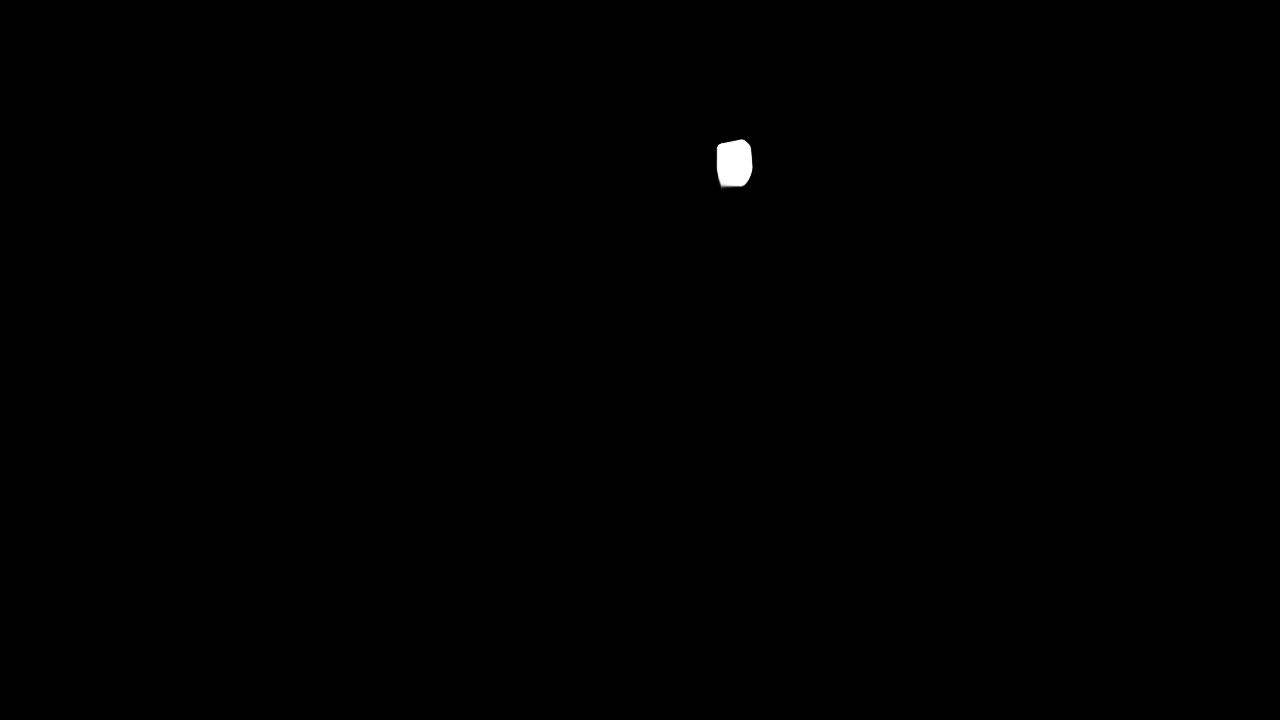}
    \\ 
    & \small{Input image} & \small{GT} &\small{CPD \cite{Wu_2019_CVPR} } & \small{SINet \cite{fan2020Camouflage}} & \small{RCRNet \cite{yan2019semi}} & \small{MG \cite{yang2021selfsupervised}} & \small{Ours}
    \\
    \end{tabular}
    \vspace{-10pt}
    \caption{Qualitative results on our MoCA-Mask benchmark. Our model provides more accurate prediction of camouflaged objects in various challenging situations, \ie, unclear appearance (arctic fox), low lighting condition (sand cat), and tiny object (ibex). }
    \label{fig:MoCA}
\end{figure*}

This section performs a thorough evaluation of our proposed framework on the CAD dataset and our proposed MoCA-Mask dataset. We also provide a comprehensive VCOD benchmark to facilitate the research of VCOD.

\subsection{Datasets}
\vspace{-5pt}
\textbf{COD10K.} We pre-train all still image-based methods as well as the encoder of the video-based methods on COD10K \cite{fan2021concealed}. It is currently the largest COD dataset which consists of 5,066 camouflaged images (3,040 for training, 2,026 for testing), and is divided into five super-classes and 69 sub-classes. This dataset also provides high-quality annotation, reaching the level of matting. 

\textbf{CAD.} Camouflaged Animal Dataset (CAD) is a small set of camouflaged animals, first introduced by \cite{bideau2016s}. It includes nine short video sequences in total that were extracted from YouTube videos and accompanying hand-labeled ground-truth masks on every $5^{th}$ frame. We also provide pseudo GT masks by a bidirectional consistency check strategy~\cite{teed2020raft} 
to enable future studies on this dataset.

%\textbf{COD10K.} We train and evaluate all still image based methods on COD10K \cite{fan2021concealed}. It is currently the largest COD dataset. It consists of 5,066 camouflaged images (3,040 for training, 2,026 for testing), which is divided into 5 super-classes and 69 sub-classes. This dataset also provides high-quality fine annotation, reaching the level of matting. 

\textbf{MoCA-Mask.}
The original Moving Camouflaged Animals (MoCA) Dataset \cite{lamdouar2020betrayed} includes 37K frames from 141 YouTube Video sequences with resolution and sampling rate of $720 \times 1280$ and 24fps in the majority of cases. The dataset covers 67 types of animals moving in natural scenes, but some are not camouflaged animals. Also, the ground truth of the original dataset is bounding boxes rather than dense segmentation masks, which makes it hard to evaluate the VCOD segmentation performance. To this end, we reorganize the dataset as \textit{MoCA-Mask} and build a comprehensive benchmark with more comprehensive evaluation criteria. 
The modifications could be found in \supp{Supp}.

\subsection{Benchmarks}
% \vspace{-5pt}
\textbf{Metrics.}  
We adopt the following evaluation metrics to measure the pixel-wise masks: (1) MAE ($M$), which assesses the pixel-level accuracy between prediction and labeled masks. (2) Enhanced-alignment measure ($E_\phi$) \cite{Fan2018Enhanced}, which simultaneously evaluates the pixel-level matching and image-level statistics. This metric is naturally suited for assessing the overall and localized accuracy of the camouflaged object detection results. Note that we report mean $E_\phi$ in the experiments. (3) S-measure ($S_\alpha$) \cite{fan2017structure}, which evaluates region-aware and object-aware structural similarity. (4) Weighted F-measure $F_\beta^w$ \cite{margolin2014evaluate} can provide more reliable evaluation results than the traditional $F_\beta$. (5) mean Dice, which measures the similarity between two sets of data. (6) meanIoU, which measures the overlap between two masks.

\textbf{Baseline.}
We select nine cutting-edge baselines, including \textbf{I.} six image based methods \ie, EGNet \cite{zhao2019EGNet}, BASNet \cite{Qin_2019_CVPR} ,   {CPD} \cite{Wu_2019_CVPR}, {PraNet} \cite{fan2020pra}, {SINet} \cite{fan2020Camouflage}, {SINet-v2} \cite{fan2021concealed}, %Polyp-PVT, 
and \textbf{II.}  three video based methods, \ie, PNS-Net~\cite{ji2021progressively}, {RCRNet} \cite{yan2019semi}, and {MotionGroup} \cite{yang2021selfsupervised}. Please refer to the \supp{Supp} for the implementation details.

\textbf{Settings.}
We compare our method primarily with the top-performing single image and video baselines. As network architectures, input resolution, modality, pre-processing, and post-processing are all different, we try our best to conduct the comparison as fairly as possible. For single image baselines, we adopt the same data pre-processing as~\cite{fan2020Camouflage, fan2021concealed} for all the compared methods. Specifically, the input images are resized to $352 \times 352$, after random flip, random rotation, and color enhance augmentation. In the training phase, we apply random pepper noise on the GT images.
As EGNet \cite{zhao2019EGNet} requires extra edge/boundary information for training, we adopt the same pre-processing techniques in their paper to obtain the edge maps. This extra information could also be found in our reorganized version of the MoCA-Mask dataset. %For video approaches, we discard the gains from post-processing, which typically include an ensemble of multiple crops and flow steps.

Most of the video approaches, \eg, PNS-Net \cite{ji2021progressively}, RCRNet \cite{yan2019semi}, employ a multistage training pipeline. The model is pre-trained using still image datasets and then equipped with temporal modules to process video datasets. We follow this training strategy and pre-train all methods on the COD10K~\cite{fan2021concealed} training set, except MotionGroup~\cite{yang2021selfsupervised} which does not have a static model. Also, per our practical experience, loading pre-trained weights on the COD10K dataset could further improve the model performance on MoCA-Mask. Compared with the COD10K image dataset, the video dataset MoCA-Mask is more challenging due to the camera motions, blurring images, small ratio of animals, and their tiny body structures, such as slim torso/limbs. In some video sequences, the animals make up a tiny proportion of the entire frame, which makes them extremely hard to be identified (see, for example, ibex in \figref{fig:MoCA}). Based upon the considerations above, we provide the results based on the following setting: (a) Training the models on COD10K; (b) Fine-tuning the models on MoCA-Mask, with pre-trained weights on COD10K; (c) Evaluate the models on the whole CAD, the test set of MoCA-Mask.

\subsection{Results}
\textbf{Performance on MoCA-Mask.} 
In \tabref{tab:Moca}, our approach outperforms all the studied methods by a significant margin, notably by $ 9.88\% $ on $S_\alpha$ over the best one in this evaluation, RCRNet \cite{yan2019semi}, and $ 92.97\% $ on $F_\beta^w$ metric over SINet \cite{fan2020Camouflage}. 
We also provide the qualitative comparisons of our method and other baselines in \figref{fig:MoCA}. Our model can accurately locate and segment camouflaged objects in many challenging situations, such as objects with the tinny torso or complex appearance textures, blur, or abrupt motions. We provide more details, \ie, per-sequence quantitative and qualitative results in the \supp{Supp}, to illustrate the consistent success over the consecutive frames.

\textbf{Performance on CAD.} In \tabref{tab:results_CAD}, we assess different approaches by studying their cross-dataset generalization on the CAD dataset. Again, the proposed network obtains the best performance in terms of all six evaluation metrics, further demonstrating its robustness. As shown in \figref{fig:CAD}, our model achieves sharper boundaries with more fine-grained visual details. This benefits from constructing pixel-level correlation pairs in the feature space.

\begin{table}[t!]
  \footnotesize
  \centering
  \caption{Quantitative results on CAD dataset. Bold indicates the best. 
  Our model consistently achieves better performance than other competitors on all metrics.
  }
  \label{tab:results_CAD}
  \vspace{-5pt}
  \tabcolsep=0.19cm
  \renewcommand{\arraystretch}{0.5}
%   \vspace{-5pt}
  \begin{tabular}{r|cccccc} 
  \toprule
  Models & $S_\alpha\uparrow$ &$F_\beta^w\uparrow$  &$E_\phi\uparrow$ &$M\downarrow$ & mDic & mIoU\\
  \cmidrule(lr){1-7}
  {EGNet} \cite{zhao2019EGNet}   & 0.619 & 0.298 & 0.666 & 0.044 & 0.324 & 0.243   \\
  {BASNet} \cite{Qin_2019_CVPR}  & 0.639 & 0.349 & 0.773 & 0.054 & 0.393 & 0.293    \\
  {CPD} \cite{Wu_2019_CVPR}      & 0.622 & 0.289 & 0.667 & 0.049 & 0.330 & 0.239 \\
  {PraNet} \cite{fan2020pra}     & 0.629 & 0.352 & 0.763 & 0.042 & 0.378 & 0.290 \\
  {SINet} \cite{fan2020Camouflage}  &  0.636 & 0.346 & 0.775 & 0.041 & 0.381 & 0.283   \\
  {SINet-v2} \cite{fan2021concealed}  &  0.653 & 0.382 & 0.762 & 0.039 & 0.413 & 0.318 \\ 
  %\textbf{Polyp-PVT}  & 0.664 & 0.426 & 0.791 &  0.039 & 0.461 & 0.357 \\
  \cmidrule(lr){1-7}
  {PNS-Net} \cite{ji2021progressively} & 0.655 & 0.325 & 0.673 & 0.048 &  0.384 &  0.290  \\
  {RCRNet} \cite{yan2019semi} & 0.627 & 0.287 & 0.666 & 0.048 & 0.309 & 0.229  \\
  {MG} \cite{yang2021selfsupervised} & 0.594 & 0.336 & 0.691 & 0.059 & 0.368 & 0.268 \\
  
 % \textbf{\Ourmodel-short}  & 0.696 & 0.471 & 0.827 & 0.031 & 0.484 & 0.392 \\
  \textbf{\Ourmodel~(Ours)} & \textbf{0.696} & \textbf{0.481} & \textbf{0.845} & \textbf{0.030} & \textbf{0.493} & \textbf{0.401} \\
  \bottomrule
  \end{tabular}
\end{table}

\begin{figure*}[t!]
\small
    \centering
    \tabcolsep=0.02cm
    \renewcommand{\arraystretch}{1.0}
    \begin{tabular}{c c c c c c c c}
    \rotatebox{90}{~\quad frog} &
    \includegraphics[width=0.138\linewidth]{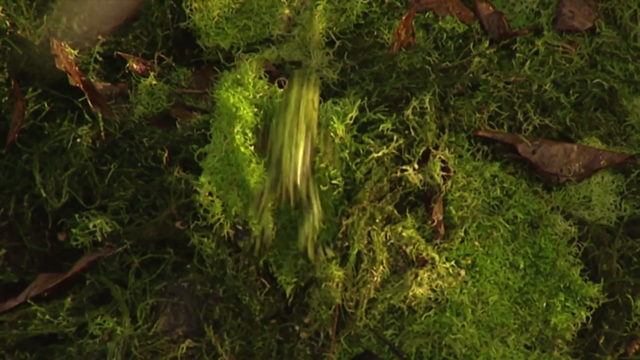} &
    \includegraphics[width=0.138\linewidth]{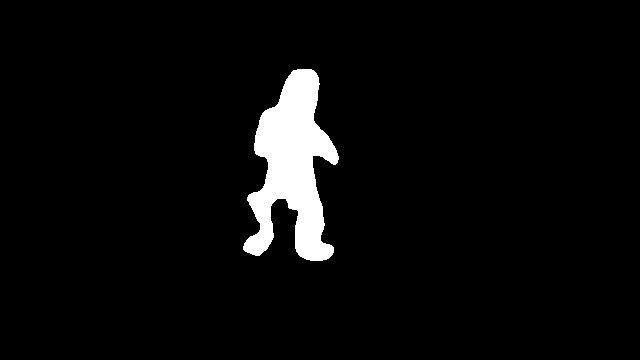} & 
    \includegraphics[width=0.138\linewidth]{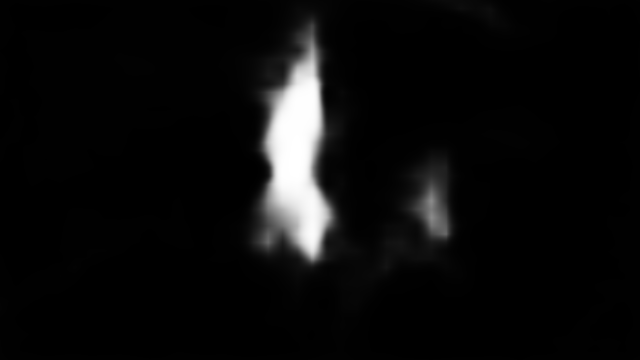} & 
    \includegraphics[width=0.138\linewidth]{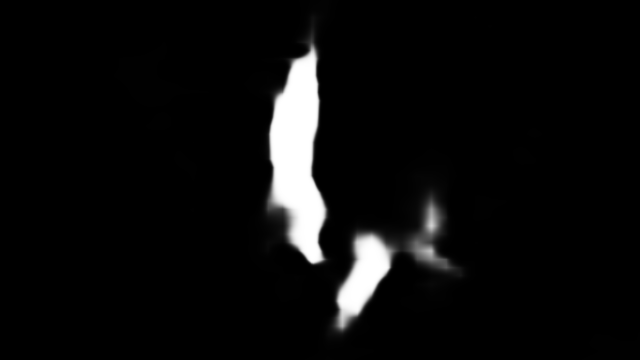} &
    \includegraphics[width=0.138\linewidth]{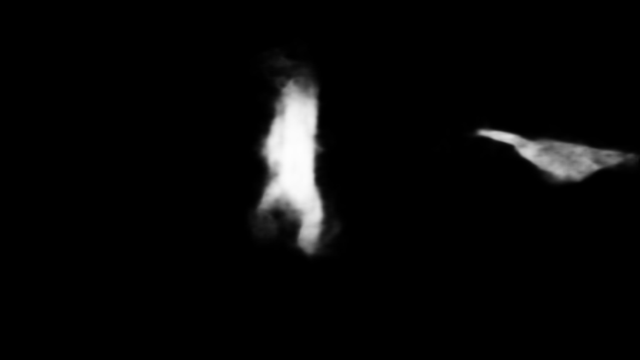} & 
    \includegraphics[width=0.138\linewidth]{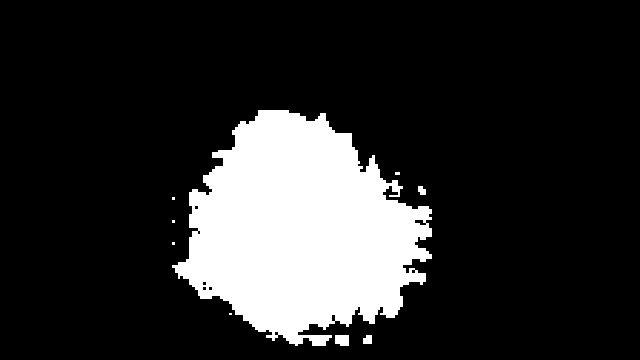} & 
    \includegraphics[width=0.138\linewidth]{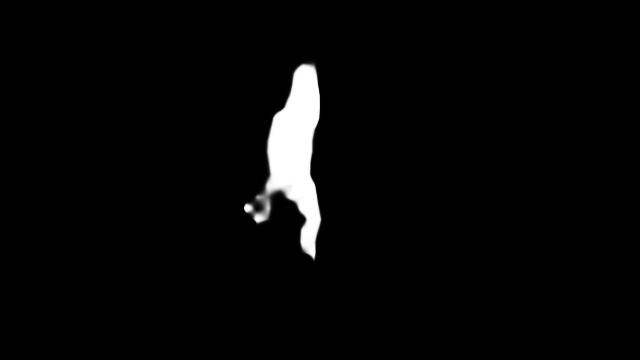}
    \\    
    \rotatebox{90}{~scorpion} &
    \includegraphics[width=0.138\linewidth]{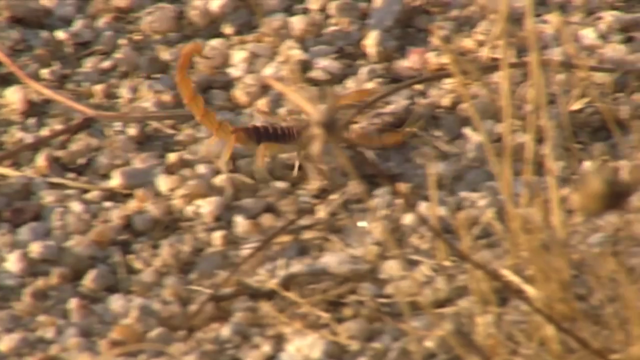} &
    \includegraphics[width=0.138\linewidth]{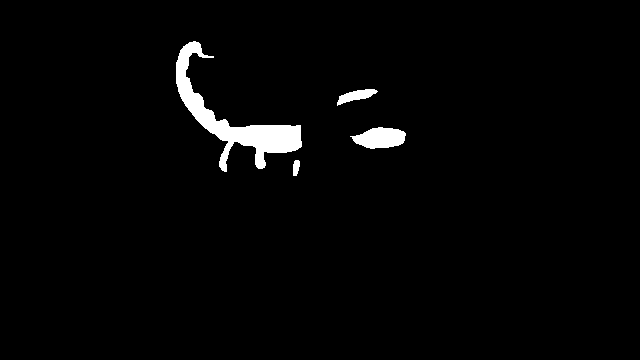} & 
    \includegraphics[width=0.138\linewidth]{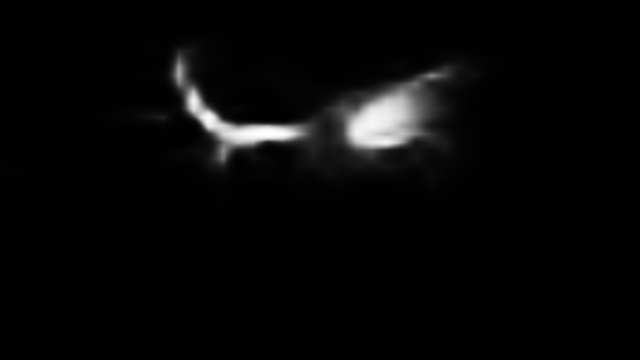} & 
    \includegraphics[width=0.138\linewidth]{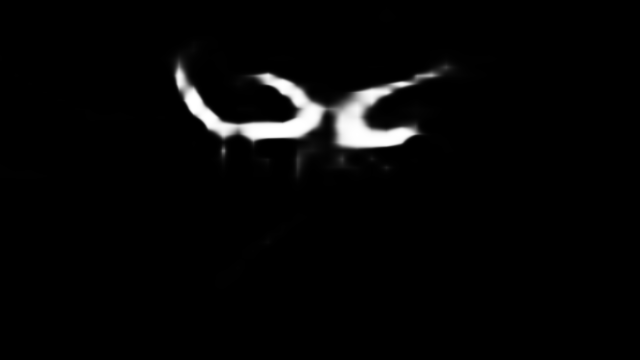} &
    \includegraphics[width=0.138\linewidth]{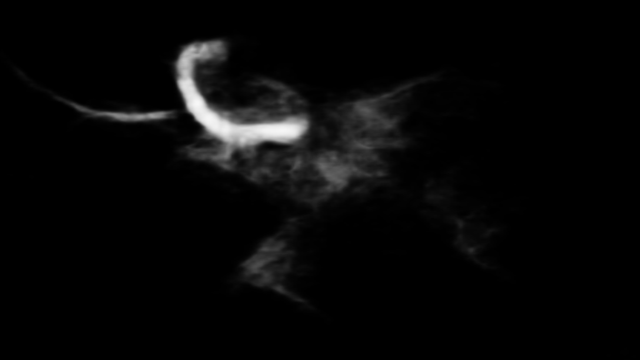} & 
    \includegraphics[width=0.138\linewidth]{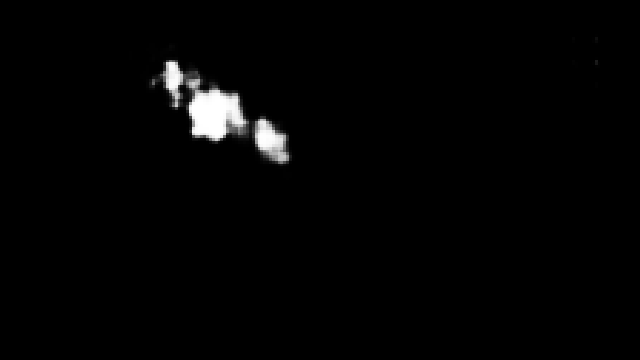} & 
    \includegraphics[width=0.138\linewidth]{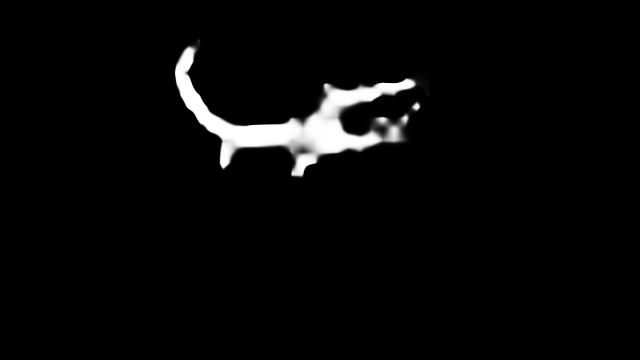}
    \\        
     
    & \small{Input image} & \small{GT}  &\small{CPD \cite{Wu_2019_CVPR} } & \small{SINet \cite{fan2020Camouflage}} & \small{RCRNet \cite{yan2019semi}} & \small{MG \cite{yang2021selfsupervised}} & \small{Ours}
    \\
    \end{tabular}
    \vspace{-10pt}
    \caption{Qualitative results on CAD dataset. As shown, our model can predict more fine-grained detail (scorpion) and work for abrupt motion (frog), which benefits from dense correspondence pair of the feature volume. }
    \label{fig:CAD}
\end{figure*}
\vspace{-10pt}

\subsection{Ablation Studies}\label{sec:ablation}
% \vspace{-5pt}
We perform ablation studies on the MoCA-Mask dataset. In particular, we look into functionality analysis for our short-term and long-term modules, the choice of sequence-to-sequence model, and our pseudo masks.

\begin{table}[t!]
    \footnotesize
    \centering
    \caption{Ablation on the short-term and long-term modules of \Ourmodel~on the MoCA-Mask dataset. Bold indicates the best.}
    \label{tab:ab_short_long}
    \vspace{-5pt}
    \tabcolsep=0.14cm
    \renewcommand{\arraystretch}{0.5}
    \begin{tabular}{ccc|cccc}
    \toprule
     Backbone & Short-term & Long-term & $S_\alpha\uparrow$ &$F_\beta^w\uparrow$  &$E_\phi\uparrow$ &$M\downarrow$ %& mDic & mIoU
     \\ 
     \midrule 
     $\surd$ & & & {0.648} &  {0.330}  & {0.748}  &  {0.025}  %& {0.375}  &  {0.289}  
     \\
     $\surd$ & $\surd$ & &  \textbf{0.662} & 0.350  & 0.766  & 0.021  %& 0.392  & 0.303   
     \\
     $\surd$ & $\surd$ & $\surd$ & 0.656 & \textbf{0.357} & \textbf{0.785} & \textbf{0.021} %& \textbf{0.399} & \textbf{0.311} 
     \\
     
    \bottomrule
    \end{tabular}
\end{table}

\begin{table}[t!]
    \footnotesize
    \centering
    \caption{Comparing different temporal information handling strategies. We swap the encoder of the RCRNet \cite{yan2019semi} with our transformer-based encoder to evaluate the performance gain caused by different handling strategies. We use ``T'' to represent the transformer encoder, ``S'' for single frame, ``V'' for video input and $\Delta$ for the improvement. Bold indicates the best.}%We compare the original CNN encoder and the same transformer~\cite{wang2021pvtv2} encoder with us.}
    \label{tab:RCRnet}
    \vspace{-5pt}
    \tabcolsep=0.34cm
    \renewcommand{\arraystretch}{0.5}
    \begin{tabular}{l|ccccccc}
    \toprule
     %\multicolumn{4}{c}{\textbf{Architecture Variant}} & \multicolumn{6}{c}{\textbf{Achieved Network}} \\
      Model & $S_\alpha\uparrow$ &$F_\beta^w\uparrow$  &$E_\phi\uparrow$ &$M\downarrow$ %& mDic & mIoU 
      \\ 
     \midrule 
     RCRNet-TS & 0.597 & \textbf{0.206} & \textbf{0.618} & 0.043 %& 0.238 & 0.168 
     \\
     RCRNet-TV & \textbf{0.606} & 0.204 & 0.617 & \textbf{0.040} %& \textbf{0.255} & \textbf{0.186} 
     \\
     \hline
      RCRNet-$\Delta$  & 1.51\% &-0.97\% &-0.16\% &6.98\% %&7.14\% &10.71\%
      \\
     \midrule
     \Ourmodel-TS & {0.648} &  {0.330}  & {0.748}  &  {0.025}  %& {0.375}  &  {0.289}  
     \\
     \Ourmodel-TV & \textbf{0.656}& \textbf{0.357} & \textbf{0.785} & \textbf{0.021} %& \textbf{0.399} & \textbf{0.311} 
     \\
     \hline
     \Ourmodel-$\Delta$ & 1.23\% & 8.18\% & 4.94\% & 16.00\% %& 6.40\% & 7.61\% 
     \\
    \bottomrule
    \end{tabular}
    \vspace{-5pt}
\end{table}

\textbf{Short-term and Long-term Modules.}
We evaluate the effectiveness of our short-term and long-term modules in two aspects. We first perform an ablation study on the short-term and the long-term modules on the MoCA-Mask dataset and show the results in \tabref{tab:ab_short_long}. By adding the short-term module, our performance is improved by 2.16\% on $S_\alpha$, 6.06\% on $F_\beta^w$, 2.41\% on $E_\phi$, 16.00\% on $M$, 4.53\% on mDic, and 4.84\% on mIoU. By adding the long-term module, we further improve our performance by 2\% on $F_\beta^w$, %2.74\% on $E_\phi$, 1.79\% on mDic and 2.64\% on mIoU,
while a slight drop 0.91\% on $S_\alpha$. 
%On CAD dataset, the long-term module consistently improves the short-term results by 0.14\% on $S_\alpha$, 2.1\% on $F_\beta^w$, 2.18\% on $E_\phi$, 3.23\% on $M$,	1.86\% on mDic, and 2.55\% on mIoU.

We then swap the encoder of a SOTA VSOD method {RCRNet} \cite{yan2019semi} with our transformer based encoder to compare the effectiveness of the temporal information handling strategies between ours and the RCRNet in \tabref{tab:RCRnet}. In terms of its spatiotemporal coherence model, it shows both positive and negative gains on the evaluated metric, \ie, 1.51\% on $S_\alpha$, -0.97\% on $F_\beta^w$, -0.16\% on $E_\phi$,  6.98\% on $M$.%, 7.14\% on mDic, and 10.71\% on mIoU.

\begin{table}[t!]
    \footnotesize
    \centering
    \tabcolsep=0.22cm
    \caption{Ablation studies of different long-term architectures on MoCA-Mask test set. The input resolution is $ 256 \times 448$.}\label{tab:long_term_archs}
    \vspace{-5pt}
    \renewcommand{\arraystretch}{0.5}
    \begin{tabular}{lr|cccc}
    \toprule
     \textbf{Arch. Variant} & Params & $S_\alpha\uparrow$ &$F_\beta^w\uparrow$  &$E_\phi\uparrow$ &$M\downarrow$ \\ 
     \midrule
      ConvLSTM & 179.03 MB & 0.651 & 0.348  & 0.767 & \textbf{0.021} \\
      Transformer   & 82.30 MB & \textbf{0.656} & \textbf{0.357} & \textbf{0.785} & \textbf{0.021} \\
    \bottomrule
    \vspace{-4mm}
    \end{tabular}
\end{table}

\textbf{Transformer \vs~ConvLSTM.}
We evaluate two different approaches for constructing long-term architecture, namely  transformer based model, and ConvLSTM based model. For the latter ConvLSTM network variant, we adopt a sequence model proposed in \cite{denton2018stochastic} but modify the original VGG-style network for the CNN encoder and decoder with our transformer-style backbone network. From the \tabref{tab:long_term_archs}, we can observe that the transformer variant is more accurate than the ConvLSTM model in all four metrics, with a much smaller number of parameters.

\textbf{Pseudo Masks.} As shown in \tabref{tab:Moca}, although the generated pseudo labels contain some noises, they can improve the performance of video approaches as they can leverage temporal information to suppress the label noises. For still image baselines, almost all of them are seriously effected by the label noises, leading to worse performance than the one without pseudo labels. It also proves that the motion estimation error can not be overlooked in the VCOD problem and we should jointly optimize it with the segmentation error for a better performance. 

\textbf{Trained from scratch on MoCA-Mask.} 
For the sake of completeness, we provide the accuracy of our network with/without pre-trained weights in Table~\ref{tab:train_from_scratch}. It shows that the gap between the train-from-scratch and the pre-trained model is minor, \ie, only a slight drop 0.15\% on $S_\alpha$.

\begin{table}[h!]
    \footnotesize
    \centering
    \caption{Comparison of trained from scratch and using pre-trained models on MoCA-Mask.  }
    \label{tab:train_from_scratch}
    \vspace{-10pt}
    \tabcolsep=0.2cm
    \renewcommand{\arraystretch}{0.5}
    \begin{tabular}{l|cccc}
    \toprule
      Model  & $S_\alpha\uparrow$ &$F_\beta^w\uparrow$  &$E_\phi\uparrow$ &$M\downarrow$ \\
     \midrule
     Trained from scratch & 0.655 & 0.351 & 0.764 & 0.024 \\
     Pre-trained & \textbf{0.656} &  \textbf{0.357} & \textbf{0.785}  & \textbf{0.021} \\
    \bottomrule
    \end{tabular}
\vspace{-5pt}
\end{table}

\vspace{-5pt}
\begin{table}[h!]
    \footnotesize
    \centering
    \caption{Comparison with \cite{lamdouar2020betrayed} on DAVIS16.  }
    \label{tab:davis}
    \vspace{-10pt}
    \tabcolsep=0.2cm
    \renewcommand{\arraystretch}{0.5}
    \begin{tabular}{l|cccc}
    \toprule
      Model & $\mathcal{J}_{Mean}\uparrow$ &$\mathcal{J}_{Recall}\uparrow$  &$\mathcal{F}_{Mean}\uparrow$ &$\mathcal{F}_{Recall}\uparrow$ \\
     \midrule
     \cite{lamdouar2020betrayed} & 65.3 & 77.3 & 65.1 & 74.1 \\
     \Ourmodel & 77.96 & 95.49 & 78.65 & 92.08 \\
    \bottomrule
    \end{tabular}
\vspace{-5pt}
\end{table}
\begin{figure}[t!]
\small
    \centering
    \tabcolsep=0.02cm
    \renewcommand{\arraystretch}{1.0}
    \begin{tabular}{c c c c}
    %\rotatebox{90}{~kite-surf} &
    \includegraphics[width=0.245\linewidth]{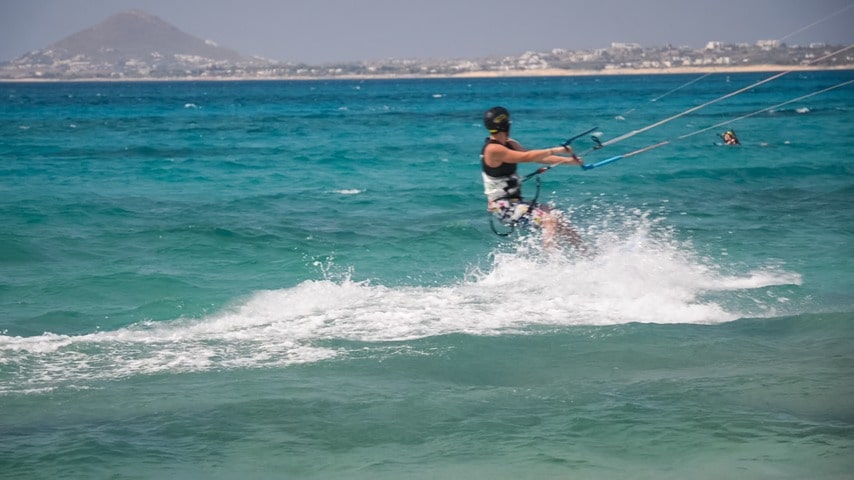} & 
    \includegraphics[width=0.245\linewidth]{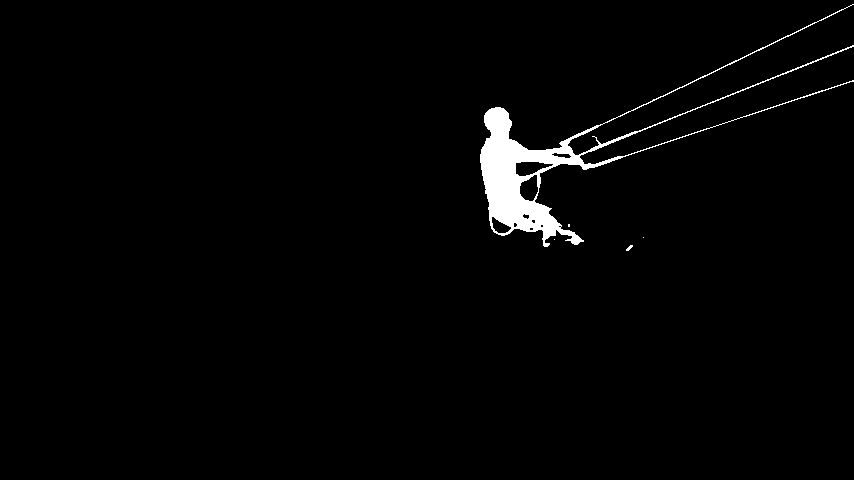} & 
    \includegraphics[width=0.245\linewidth]{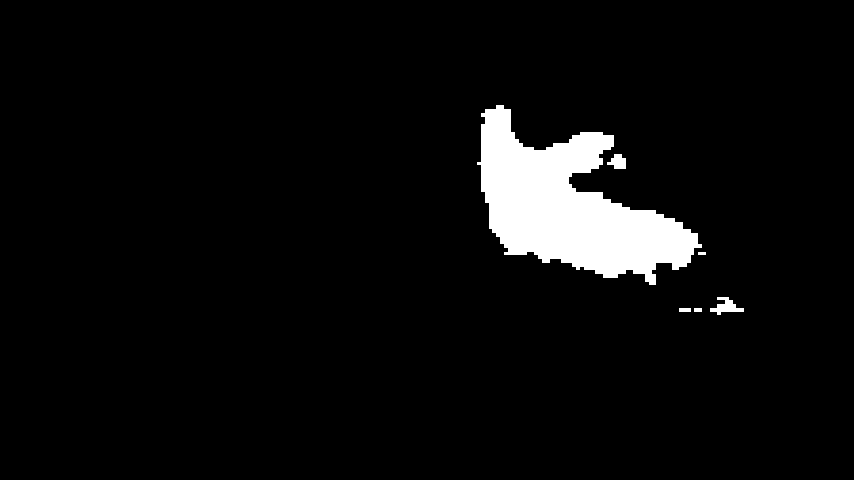} & 
    \includegraphics[width=0.245\linewidth]{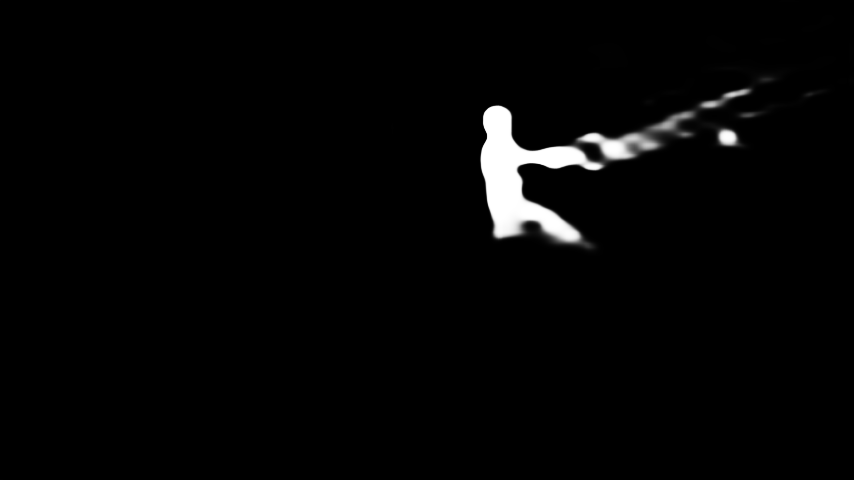}
    \\        
    \small{Input image} & \small{GT} & \small{MG \cite{yang2021selfsupervised}} & \small Ours 
    \end{tabular}
    \vspace{-10pt}
    \captionof{figure}{\small{Qualitative Results on DAVIS16. MG \cite{yang2021selfsupervised} fails where the splash created by the person is incorrectly included in the predicted segment. This mainly due to the inaccurate optical flow estimation which cannot be optimized during model training.}}
    \label{fig:failure_case}
    \vspace{-4mm}
\end{figure}

\textbf{Generalization.} Our model can be applied to the more general video object detection problem, such as video instance segmentation. Except a detailed comparison with MG~\cite{yang2021selfsupervised} in Table \ref{tab:Moca}, we compare with~\cite{lamdouar2020betrayed} on DAVIS16 (Table \ref{tab:davis}) and demonstrate superiority of our method.

\section{Conclusion}
We presented a new \Ourmodel~framework for learning to segment camouflaged objects in a video. Specifically, we proposed a short-term module to implicitly capture motions between consecutive frames which allows us to learn motion estimation and segmentation in a single optimization target. We also proposed a long-term module with a sequence-to-sequence transformer to enforce temporal consistency in video sequence. 
To promote the development of this field, we rebuild a new dataset called \textbf{MoCA-Mask} with 87 high-quality video sequences, including 22,939 frames in total. It is the largest-scale pixel-level annotated dataset that allows object-level benchmark in video camouflaged object detection (VCOD). Compared with existing state-of-the-art baselines, our proposed network achieves fascinating results on two VCOD benchmarks.

\textbf{Broader Impact.}  
Camouflaged object can be used to detect and protect rare animal species, prevent wildlife trafficking, medical applications (\eg, detecting polyp or lung infection) and search-and-rescue work to name a few. Please note that our MoCA-Mask dataset does not contain any military or sensitive scenes. Aside from its important use-cases as mentioned above, our paper takes a solid step into understanding video contents when motion information is noisy.

%\clearpage
%%%%%%%%% REFERENCES
{\small
\bibliographystyle{ieee_fullname}
\bibliography{egbib}

\begin{thebibliography}{10}\itemsep=-1pt

\bibitem{beery2020context}
Sara Beery, Guanhang Wu, Vivek Rathod, Ronny Votel, and Jonathan Huang.
\newblock Context r-cnn: Long term temporal context for per-camera object
  detection.
\newblock In {\em CVPR}, 2020.

\bibitem{bideau2016s}
Pia Bideau and Erik Learned-Miller.
\newblock It’s moving! a probabilistic model for causal motion segmentation
  in moving camera videos.
\newblock In {\em ECCV}, 2016.

\bibitem{bideau2018moa}
Pia Bideau, Rakesh~R Menon, and Erik Learned-Miller.
\newblock Moa-net: self-supervised motion segmentation.
\newblock In {\em ECCV Workshops}, 2018.

\bibitem{Butler:ECCV:2012}
D.~J. Butler, J. Wulff, G.~B. Stanley, and M.~J. Black.
\newblock A naturalistic open source movie for optical flow evaluation.
\newblock In {\em ECCV}, 2012.

\bibitem{cheng2019noise}
Xuelian Cheng, Yiran Zhong, Yuchao Dai, Pan Ji, and Hongdong Li.
\newblock Noise-aware unsupervised deep lidar-stereo fusion.
\newblock In {\em CVPR}, 2019.

\bibitem{denton2018stochastic}
Emily Denton and Rob Fergus.
\newblock Stochastic video generation with a learned prior.
\newblock In {\em ICML}, 2018.

\bibitem{fan2017structure}
Deng-Ping Fan, Ming-Ming Cheng, Yun Liu, Tao Li, and Ali Borji.
\newblock {Structure-measure: A New Way to Evaluate Foreground Maps}.
\newblock In {\em ICCV}, 2017.

\bibitem{Fan2018Enhanced}
Deng-Ping Fan, Cheng Gong, Yang Cao, Bo Ren, Ming-Ming Cheng, and Ali Borji.
\newblock {Enhanced-alignment Measure for Binary Foreground Map Evaluation}.
\newblock In {\em IJCAI}, 2018.

\bibitem{fan2021concealed}
Deng-Ping Fan, Ge-Peng Ji, Ming-Ming Cheng, and Ling Shao.
\newblock Concealed object detection.
\newblock {\em IEEE TPAMI}, 2021.

\bibitem{21Fan_HybridLoss}
Deng-Ping Fan, Ge-Peng Ji, Xuebin Qin, and Ming-Ming Cheng.
\newblock Cognitive vision inspired object segmentation metric and loss
  function.
\newblock {\em SSI}, 2021.

\bibitem{fan2020Camouflage}
Deng-Ping Fan, Ge-Peng Ji, Guolei Sun, Ming-Ming Cheng, Jianbing Shen, and Ling
  Shao.
\newblock Camouflaged object detection.
\newblock In {\em CVPR}, 2020.

\bibitem{fan2020pra}
Deng-Ping Fan, Ge-Peng Ji, Tao Zhou, Geng Chen, Huazhu Fu, Jianbing Shen, and
  Ling Shao.
\newblock Pranet: Parallel reverse attention network for polyp segmentation.
\newblock In {\em MICCAI}, 2020.

\bibitem{Fan2019VideoSal}
Deng-Ping Fan, Wenguan Wang, Ming-Ming Cheng, and Jianbing Shen.
\newblock Shifting more attention to video salient object detection.
\newblock In {\em CVPR}, 2019.

\bibitem{jain2017fusionseg}
Suyog~Dutt Jain, Bo Xiong, and Kristen Grauman.
\newblock Fusionseg: Learning to combine motion and appearance for fully
  automatic segmentation of generic objects in videos.
\newblock In {\em CVPR}, 2017.

\bibitem{ji2021progressively}
Ge-Peng Ji, Yu-Cheng Chou, Deng-Ping Fan, Geng Chen, Huazhu Fu, Debesh Jha, and
  Ling Shao.
\newblock Progressively normalized self-attention network for video polyp
  segmentation.
\newblock In {\em MICCAI}, 2021.

\bibitem{ji2021FSNet}
Ge-Peng Ji, Keren Fu, Zhe Wu, Deng-Ping Fan, Jianbing Shen, and Ling Shao.
\newblock Full-duplex strategy for video object segmentation.
\newblock In {\em ICCV}, 2021.

\bibitem{7025056}
Pan Ji, Yiran Zhong, Hongdong Li, and Mathieu Salzmann.
\newblock Null space clustering with applications to motion segmentation and
  face clustering.
\newblock In {\em ICIP}, 2014.

\bibitem{kumar2021early}
Karthika~Suresh Kumar and Aamer~Abdul Rahman.
\newblock Early detection of locust swarms using deep learning.
\newblock In {\em MLCI}. 2021.

\bibitem{lamdouar2020betrayed}
Hala Lamdouar, Charig Yang, Weidi Xie, and Andrew Zisserman.
\newblock Betrayed by motion: Camouflaged object discovery via motion
  segmentation.
\newblock In {\em ACCV}, 2020.

\bibitem{ltnghia-CVIU2019}
Trung-Nghia Le, Tam~V. Nguyen, Zhongliang Nie, Minh-Triet Tran, and Akihiro
  Sugimoto.
\newblock Anabranch network for camouflaged object segmentation.
\newblock {\em CVIU}, 2019.

\bibitem{le2017deeply}
Trung-Nghia Le and Akihiro Sugimoto.
\newblock Deeply supervised 3d recurrent fcn for salient object detection in
  videos.
\newblock In {\em BMVC}, 2017.

\bibitem{le2018video}
Trung-Nghia Le and Akihiro Sugimoto.
\newblock Video salient object detection using spatiotemporal deep features.
\newblock {\em IEEE TIP}, 2018.

\bibitem{li2021arvo}
Dongxu Li, Chenchen Xu, Kaihao Zhang, Xin Yu, Yiran Zhong, Wenqi Ren, Hanna
  Suominen, and Hongdong Li.
\newblock Arvo: Learning all-range volumetric correspondence for video
  deblurring.
\newblock In {\em CVPR}, 2021.

\bibitem{li2018flow}
Guanbin Li, Yuan Xie, Tianhao Wei, Keze Wang, and Liang Lin.
\newblock Flow guided recurrent neural encoder for video salient object
  detection.
\newblock In {\em CVPR}, 2018.

\bibitem{liu2019concealed}
Ting Liu, Yao Zhao, Yunchao Wei, Yufeng Zhao, and Shikui Wei.
\newblock Concealed object detection for activate millimeter wave image.
\newblock {\em IEEE TIE}, 2019.

\bibitem{lv2021simultaneously}
Yunqiu Lv, Jing Zhang, Yuchao Dai, Aixuan Li, Bowen Liu, Nick Barnes, and
  Deng-Ping Fan.
\newblock Simultaneously localize, segment and rank the camouflaged objects.
\newblock In {\em CVPR}, 2021.

\bibitem{margolin2014evaluate}
Ran Margolin, Lihi Zelnik-Manor, and Ayellet Tal.
\newblock How to evaluate foreground maps?
\newblock In {\em CVPR}, 2014.

\bibitem{mei2021camouflaged}
Haiyang Mei, Ge-Peng Ji, Ziqi Wei, Xin Yang, Xiaopeng Wei, and Deng-Ping Fan.
\newblock Camouflaged object segmentation with distraction mining.
\newblock In {\em CVPR}, 2021.

\bibitem{michels2005high}
Jeff Michels, Ashutosh Saxena, and Andrew~Y Ng.
\newblock High speed obstacle avoidance using monocular vision and
  reinforcement learning.
\newblock In {\em ICML}, 2005.

\bibitem{papazoglou2013fast}
Anestis Papazoglou and Vittorio Ferrari.
\newblock Fast object segmentation in unconstrained video.
\newblock In {\em ICCV}, 2013.

\bibitem{Qin_2019_CVPR}
Xuebin Qin, Zichen Zhang, Chenyang Huang, Chao Gao, Masood Dehghan, and Martin
  Jagersand.
\newblock Basnet: Boundary-aware salient object detection.
\newblock In {\em CVPR}, 2019.

\bibitem{zhen2022cosformer}
Zhen Qin, Weixuan Sun, Hui Deng, Dongxu Li, Yunshen Wei, Baohong Lv, Junjie
  Yan, Lingpeng Kong, and Yiran Zhong.
\newblock cosformer: Rethinking softmax in attention.
\newblock In {\em ICLR}, 2022.

\bibitem{ranjan2019competitive}
Anurag Ranjan, Varun Jampani, Lukas Balles, Kihwan Kim, Deqing Sun, Jonas
  Wulff, and Michael~J Black.
\newblock Competitive collaboration: Joint unsupervised learning of depth,
  camera motion, optical flow and motion segmentation.
\newblock In {\em CVPR}, 2019.

\bibitem{song2018pyramid}
Hongmei Song, Wenguan Wang, Sanyuan Zhao, Jianbing Shen, and Kin-Man Lam.
\newblock Pyramid dilated deeper convlstm for video salient object detection.
\newblock In {\em ECCV}, 2018.

\bibitem{sundaram2010dense}
Narayanan Sundaram, Thomas Brox, and Kurt Keutzer.
\newblock Dense point trajectories by gpu-accelerated large displacement
  optical flow.
\newblock In {\em ECCV}, 2010.

\bibitem{teed2020raft}
Zachary Teed and Jia Deng.
\newblock Raft: Recurrent all-pairs field transforms for optical flow.
\newblock In {\em ECCV}, 2020.

\bibitem{tokmakov2017learning}
Pavel Tokmakov, Karteek Alahari, and Cordelia Schmid.
\newblock Learning motion patterns in videos.
\newblock In {\em CVPR}, 2017.

\bibitem{Wang_2021_CVPR}
Jianyuan Wang, Yiran Zhong, Yuchao Dai, Stan Birchfield, Kaihao Zhang, Nikolai
  Smolyanskiy, and Hongdong Li.
\newblock Deep two-view structure-from-motion revisited.
\newblock In {\em CVPR}, 2021.

\bibitem{NEURIPS2020_add5aebf}
Jianyuan Wang, Yiran Zhong, Yuchao Dai, Kaihao Zhang, Pan Ji, and Hongdong Li.
\newblock Displacement-invariant matching cost learning for accurate optical
  flow estimation.
\newblock In {\em NeurIPS}, 2020.

\bibitem{wang2017video}
Wenguan Wang, Jianbing Shen, and Ling Shao.
\newblock Video salient object detection via fully convolutional networks.
\newblock {\em IEEE TIP}, 2017.

\bibitem{wang2021pvtv2}
Wenhai Wang, Enze Xie, Xiang Li, Deng-Ping Fan, Kaitao Song, Ding Liang, Tong
  Lu, Ping Luo, and Ling Shao.
\newblock Pvtv2: Improved baselines with pyramid vision transformer.
\newblock {\em arXiv preprint arXiv:2106.13797}, 2021.

\bibitem{wang2021pyramid}
Wenhai Wang, Enze Xie, Xiang Li, Deng-Ping Fan, Kaitao Song, Ding Liang, Tong
  Lu, Ping Luo, and Ling Shao.
\newblock Pyramid vision transformer: A versatile backbone for dense prediction
  without convolutions.
\newblock In {\em ICCV}, 2021.

\bibitem{wei2020f3net}
Jun Wei, Shuhui Wang, and Qingming Huang.
\newblock F$^3$net: Fusion, feedback and focus for salient object detection.
\newblock In {\em AAAI}, 2020.

\bibitem{wu2021jcs}
Yu-Huan Wu, Shang-Hua Gao, Jie Mei, Jun Xu, Deng-Ping Fan, Rong-Guo Zhang, and
  Ming-Ming Cheng.
\newblock Jcs: An explainable covid-19 diagnosis system by joint classification
  and segmentation.
\newblock {\em IEEE TIP}, 2021.

\bibitem{Wu_2019_CVPR}
Zhe Wu, Li Su, and Qingming Huang.
\newblock Cascaded partial decoder for fast and accurate salient object
  detection.
\newblock In {\em CVPR}, 2019.

\bibitem{xingjian2015convolutional}
SHI Xingjian, Zhourong Chen, Hao Wang, Dit-Yan Yeung, Wai-Kin Wong, and
  Wang-chun Woo.
\newblock Convolutional lstm network: A machine learning approach for
  precipitation nowcasting.
\newblock In {\em NeurIPS}, 2015.

\bibitem{yan2019semi}
Pengxiang Yan, Guanbin Li, Yuan Xie, Zhen Li, Chuan Wang, Tianshui Chen, and
  Liang Lin.
\newblock Semi-supervised video salient object detection using pseudo-labels.
\newblock In {\em ICCV}, 2019.

\bibitem{yang2021selfsupervised}
Charig Yang, Hala Lamdouar, Erika Lu, Andrew Zisserman, and Weidi Xie.
\newblock Self-supervised video object segmentation by motion grouping.
\newblock In {\em ICCV}, 2021.

\bibitem{yang2019anchor}
Zhao Yang, Qiang Wang, Luca Bertinetto, Weiming Hu, Song Bai, and Philip~HS
  Torr.
\newblock Anchor diffusion for unsupervised video object segmentation.
\newblock In {\em CVPR}, 2019.

\bibitem{zhai2021mutual}
Qiang Zhai, Xin Li, Fan Yang, Chenglizhao Chen, Hong Cheng, and Deng-Ping Fan.
\newblock Mutual graph learning for camouflaged object detection.
\newblock In {\em CVPR}, 2021.

\bibitem{Zhang_2021_ICCV}
Jing Zhang, Deng-Ping Fan, Yuchao Dai, Xin Yu, Yiran Zhong, Nick Barnes, and
  Ling Shao.
\newblock Rgb-d saliency detection via cascaded mutual information
  minimization.
\newblock In {\em ICCV}, 2021.

\bibitem{zhao2019EGNet}
Jia-Xing Zhao, Jiang-Jiang Liu, Deng-Ping Fan, Yang Cao, Jufeng Yang, and
  Ming-Ming Cheng.
\newblock Egnet:edge guidance network for salient object detection.
\newblock In {\em ICCV}, 2019.

\bibitem{zhongicpr18}
Yiran Zhong, Yuchao Dai, and Hongdong Li.
\newblock 3d geometry-aware semantic labeling of outdoor street scenes.
\newblock In {\em ICPR}, 2018.

\bibitem{Zhong_2019_CVPR}
Yiran Zhong, Pan Ji, Jianyuan Wang, Yuchao Dai, and Hongdong Li.
\newblock Unsupervised deep epipolar flow for stationary or dynamic scenes.
\newblock In {\em CVPR}, 2019.

\bibitem{Zhong_2018_ECCV}
Yiran Zhong, Hongdong Li, and Yuchao Dai.
\newblock Open-world stereo video matching with deep rnn.
\newblock In {\em ECCV}, 2018.

\end{thebibliography}
}

\clearpage

% \title{Implicit Motion Handling for Video Camouflaged Object Detection \\-- Supplementary Material -- }

% \author{\textsuperscript{*}Xuelian Cheng$^{1}$,
% \textsuperscript{*}Huan Xiong$^{3}$,
% $^\dagger$Deng-ping Fan$^{4}$, Yiran Zhong$^{6,7}$, \\
% Mehrtash Harandi$^{1,8}$, Tom Drummond$^{1}$, Zongyuan Ge$^{1,2,5}$ \\
% $^{1}$Faculty of Engineering, Monash University, 
% $^{2}$eResearch Centre, Monash University \\
% $^{3}$Mohamed bin Zayed University of Artificial Intelligence, 
% $^{4}$CVL, ETH Zurich,\\
% $^{5}$Airdoc Research Australia, 
% $^{6}$SenseTime Research, $^{7}$Shanghai AI Laboratory,
% $^{8}$Data61, CSIRO \\
% %\tt\small{\{xuelian.cheng, mehrtash.harandi, zongyuan.ge\}@monash.edu} \\ 
% %\tt\small{\{huan.xiong.math, dengpfan, zhongyiran\}@gmail.com, tom.drummond@unimelb.edu.au}
% } 
% \maketitle
\section{Supplementary Material}

\paragraph{Short-term Correlation Pyramid Details}
To enable the network to learn detailed information, a correlation pyramid $\mathbf{C}^{i}, i \in \{2, 3, 4\}$ is construct by incorporating multi-scale features. Thus for a sequence of frame features   $\{\mathcal{F}_\theta(\mathbf{I}_{t}) , \mathcal{F}_\theta(\mathbf{I}_{t+1}) \} \in \mathbb{R}^{C \times H/2^{i+1} \times W/2^{i+1}}$, our short-term correlation pyramid can be denoted as $\mathbf{C}^i(\mathbf{I}_t,\,\mathbf{I}_{t+1}) \in \mathbb{R}^{ H/2^{i+1} \times W/2^{i+1} \times  H/2^{i+1} \times W/2^{i+1} }$. It outputs an aggregated feature map $f'^{(i)}_{t \leftarrow t+1 }(\mathbf{I}_{t\leftarrow t+1})$ at the pyramid scale $i, i \in \{2, 3, 4\}$, which has the same dimension as the reference frame feature $\mathcal{F}_{\theta}(\mathbf{I}_t)$. For downsampled neighboring frames, we set the $k=\{2,4,8\}$ with max-pooling kernels of growing size. We also repeat the correlative aggregation once on every other neighboring frame. In this way, we obtain aggregated feature maps $f'^{(i)}_{t \leftarrow t+1 }(\mathbf{I}_{t\leftarrow t+2})$. 

\paragraph{Semi-supervised Training Procedure}\label{sec:pseudo_label} 
\vspace{-5pt}
As the annotations are provided in the form of dense segmentation masks  for every five frames, %$5^\text{th}$ frame, 
we adopt a bi-directional consistency check strategy to generate pseudo masks for unlabelled frames.
Given five consecutive frames $\{\mathbf{I}_{t},\mathbf{I}_{t+1},\mathbf{I}_{t+2},\mathbf{I}_{t+3},\mathbf{I}_{t+4}\}$ and labelled ground-truth $\mathbf{gt}_{t}$, we first estimate forward and backward optical flow fields between frame $\mathbf{I}_{t}$ and $\mathbf{I}_{t+n}, n\in [1, 4]$. Then we can produce the warped ground-truth $\hat{\mathbf{gt}}_{t+n}$ with the inverse warping from ground-truth $\mathbf{gt}_{t}$.

\textbf{1.Flow Estimation.}
We take the ground-truth mask of the reference frame $\mathbf{I}_t$ as an example, to generate pseudo ground-truth of its immediate following frame $\mathbf{I}_{t+1}$. The optical flow estimation module\footnote{In practice, we make use of RAFT \cite{teed2020raft} to obtain the optical flow.} $\mathcal{O}$ takes $\mathbf{I}_t$ and $\mathbf{I}_{t+1}$ and predicts the optical flow field:
\begin{align}
\mathbf{u}_{t, t+1}^x,\,\mathbf{u}_{t, t+1}^y = \mathcal{O}(\mathbf{I}_t, \mathbf{I}_{t+1}),
\end{align}
where $\mathbf{u}_{t, t+1}^x$ and $\mathbf{u}_{t, t+1}^y$ denote the $x,\,y$ components of the estimated flow field, respectively. The flow field maps each pixel $(x,\,y)$ in ${\mathbf{I}}_{t+1}$ to its corresponding coordinates 
$(x^{\prime},\,y^{\prime})=(x+\mathbf{u}_{t,t+1}^x(x),\,y+\mathbf{u}_{t,t+1}^y(y))$ in ${\mathbf{I}}_{t}$.

\textbf{2.Forward/Backward Pseudo Labeling.}
Given the forward optical flow sequences $(\mathbf{flow}_{t}, \mathbf{flow}_{t+n}), n\in{1,2,3,4}$, we can obtain the aligned neighboring frame $\hat{\mathbf{gt}}_{t+n}$ by a warping interpolation on $\mathbf{gt}_{t}$ using the mapped coordinates. After repeating the explicit alignment step for the preceding frame, we acquire the sequence of warped input frames $\{\mathbf{gt}_{t},\hat{\mathbf{gt}}_{t+1},\hat{\mathbf{gt}}_{t+2},\hat{\mathbf{gt}}_{t+3},\hat{\mathbf{gt}}_{t+4}\}$. The backward pseudo ground-truth sequences are obtained by performing warping ground-truth masks with backward optical flows in the reverse order.

\begin{figure}[t!]
\begin{center}
%\vspace{-5mm}
\begin{overpic}[width=.98\columnwidth]{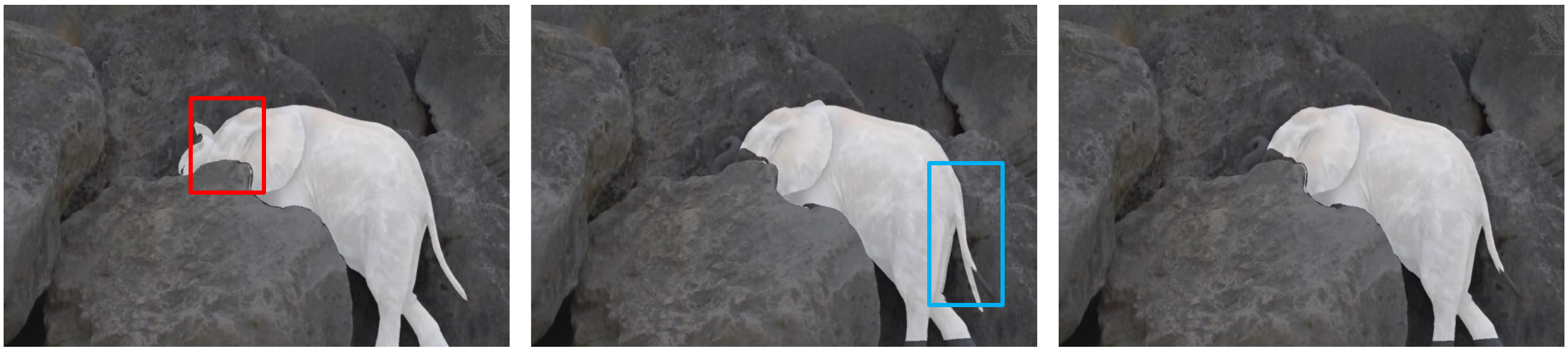}
\put(16, -8) {\small (a) Forward}
\put(90, -8) {\small (b) Backward}
\put(165, -8) {\small (c) Bi-directional}
\end{overpic}
\vspace{-2mm}
\end{center}
\caption{Illustration of forward-backward consistency check. After bi-directional check, undesirable ghosting artifacts, \ie~the nose (red box) of the elephant in forward direction and the tail (blue box) in backward direction, and occlusions can be effectively removed. }
\label{fig:bi_directional}
\end{figure}

\textbf{3.Bidirectional Consistency Check.} To identify valid masks, we adopt forward-backward consistency check to eliminate inconsistent regions. Under the forward-backward consistency assumption \cite{sundaram2010dense}, traversing flow vector forward and then backward should arrive at the same position. We mark pixels as invalid whenever this constraint is violated. As shown in \figref{fig:bi_directional}, the invalid regions emphasized by the orange boxes are marked as background.

\paragraph{Training Details}
We implement both long-term and short-term architecture in PyTorch. The input images are resized to $352 \times 352$.  We train the short-term architecture with a batch size of 8 on an NVIDIA V100 GPU and use Adam optimizer with initial learning rate of 1e-4, decreasing every 50k iterations. %The exact architecture description and training schedule can be found in Appendix.
For the long-term optimization, our model takes 10 frames as the input at one time with the frame sampling rate 1.
For our pseudo ground-truth generation, we exploit RAFT \cite{teed2020raft} as the optical flow estimation module and pre-trained weights on Sintel dataset~\cite{Butler:ECCV:2012}.

\begin{figure*}[ht]
\begin{center}
\includegraphics[width=0.8\linewidth]{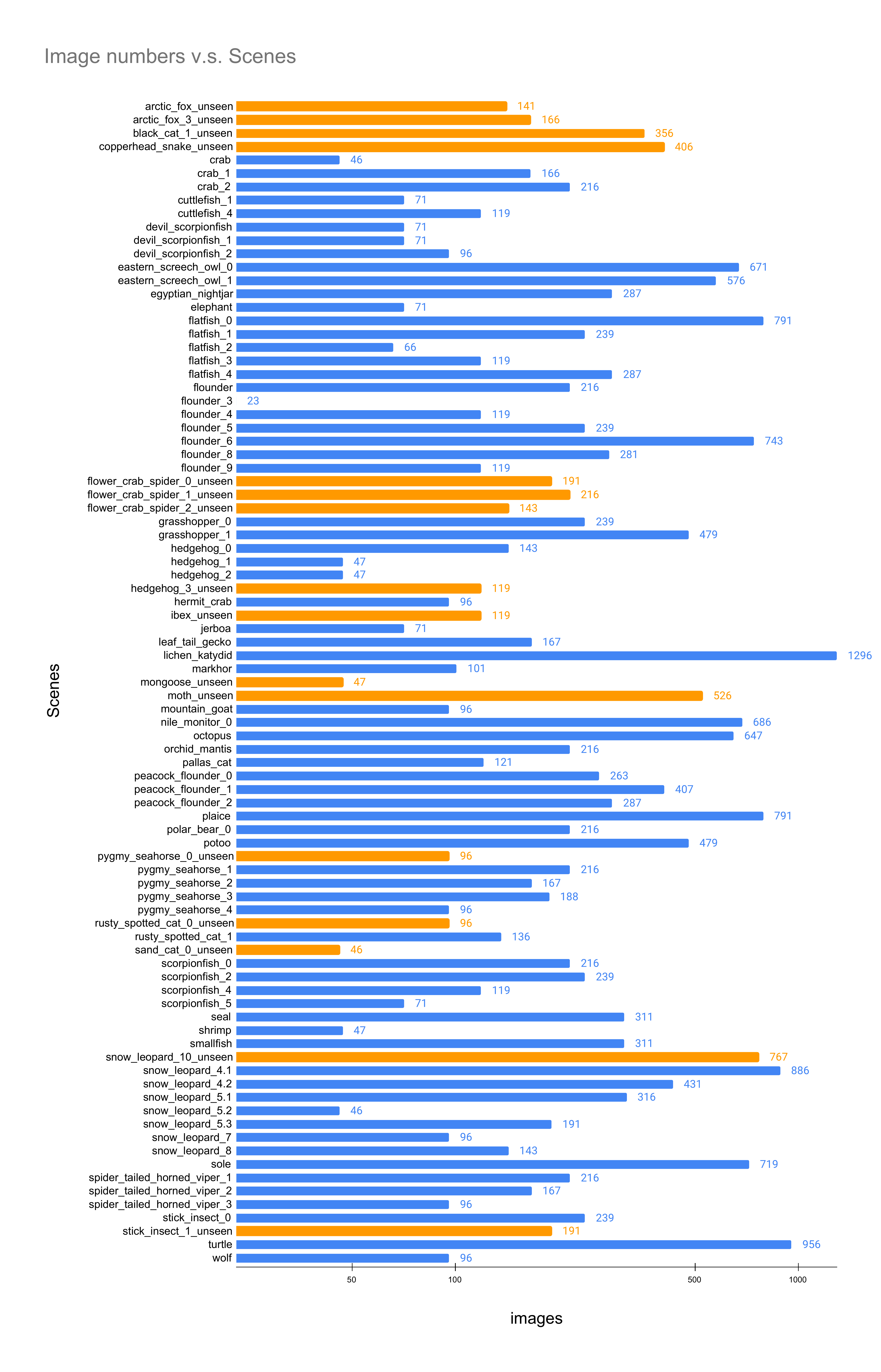}
\end{center}
\vspace{-8mm}
\caption{Summary for training and test set distribution. Our MoCA-Mask dataset includes 87 video sequences in total, in which 16 sequences were tagged as ``unknow'' (colored in orange). This split is used to validate the sensitivity of different models on novel samples. Zoom-in for details.}
\label{fig:data_distribution}
\end{figure*}

\paragraph{Data Curation}
\begin{compactitem}
    % \vspace{-3pt}
    \item \textbf{Remove Invalid Scenes.} We first select and exclude scenarios in that animals are obvious and easy to identify from the background at our first glance. After cleaning the dataset, our new subset includes 87 video sequences, 22,939 frames in total.
    % \vspace{-3pt}
    \item \textbf{Segmentation Masks.} For annotations, we further provide accurate human-labeled segmentation masks for every five frames. Thus our  GT consists of two formats, that is 4,691 bounding box annotations as well as 4,691 pixel-level masks.
    % \vspace{-3pt}
    \item \textbf{Pseudo Masks.} We use a bidirectional optical flow-based strategy to generate the pseudo GT masks, refer to the \textit{SM}. Note that these pseudo masks still contain motion estimation errors, requiring algorithms to have the capability to handle noise labels when using them. %Section~\ref{sec:pseudo_label}. 
    %Given a sequence of GT for every five frames, we first estimate the forward and backward directions of optical flows and warp the GT with the corresponding optical flows. Then invalid pixels are eliminated if their bidirectional check (consistency in forward/backward warping) fails.
    %\MH{what do you mean by processing the bidirectional check? is it, ``Invalid pixels are removed if their bidirectional check (consistency in forward/backward warping) fails.}
    % \vspace{-3pt}
    \item \textbf{Dataset Split.} The whole dataset is split into 71 sequences, 19,313 frames for training, and 16 sequences, 3,626 frames selected for testing. The summary of each sub-sequence distribution could be found in Fig. \ref{fig:data_distribution}. 
\end{compactitem}

\begin{figure*}[!ht]
\begin{center}
\vspace{-5mm}
\includegraphics[width=0.85\linewidth]{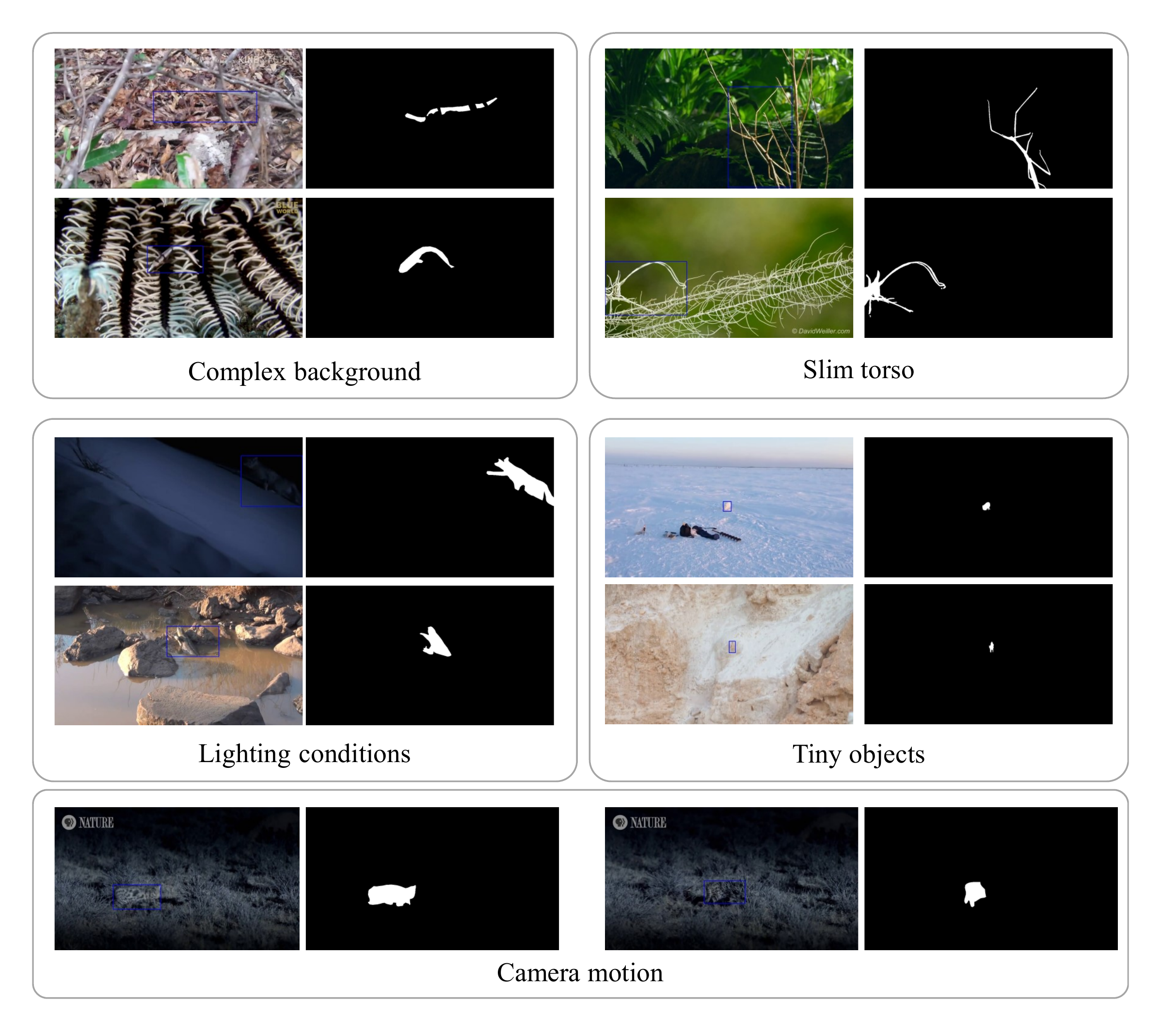}
\end{center}
\vspace{-8mm}
\caption{Representative samples from MoCA-Mask. The dataset is quite challenging including diverse scenes, suash as various lighting conditions, \ie~ dark and sunny, complex background, camera motions, small ratio of animals and tiny body structures, such as slim torso /limbs.}
\label{fig:samples}
\end{figure*}

\begin{figure*}[t!]
\small
    \centering
    \tabcolsep=0.02cm
    \renewcommand{\arraystretch}{0.8}
    \begin{tabular}{c c c c c}
    (a) &
    \includegraphics[width=0.21\linewidth]{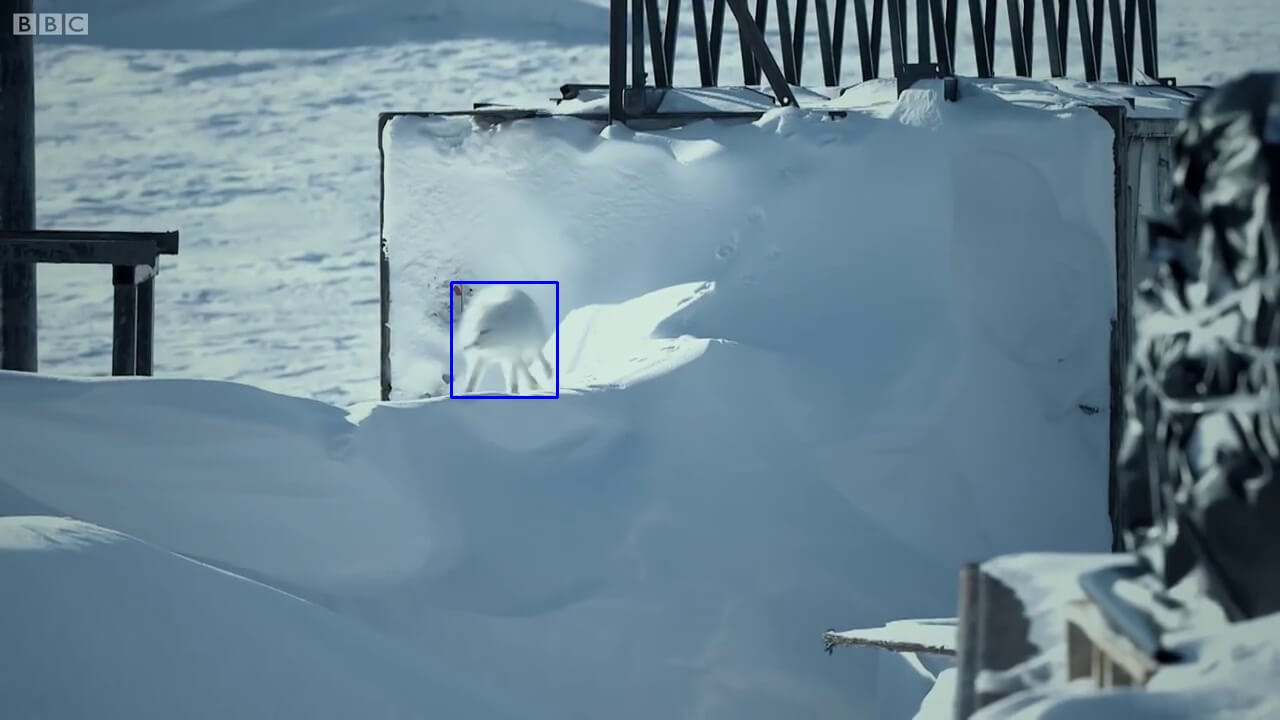} & 
    \includegraphics[width=0.21\linewidth]{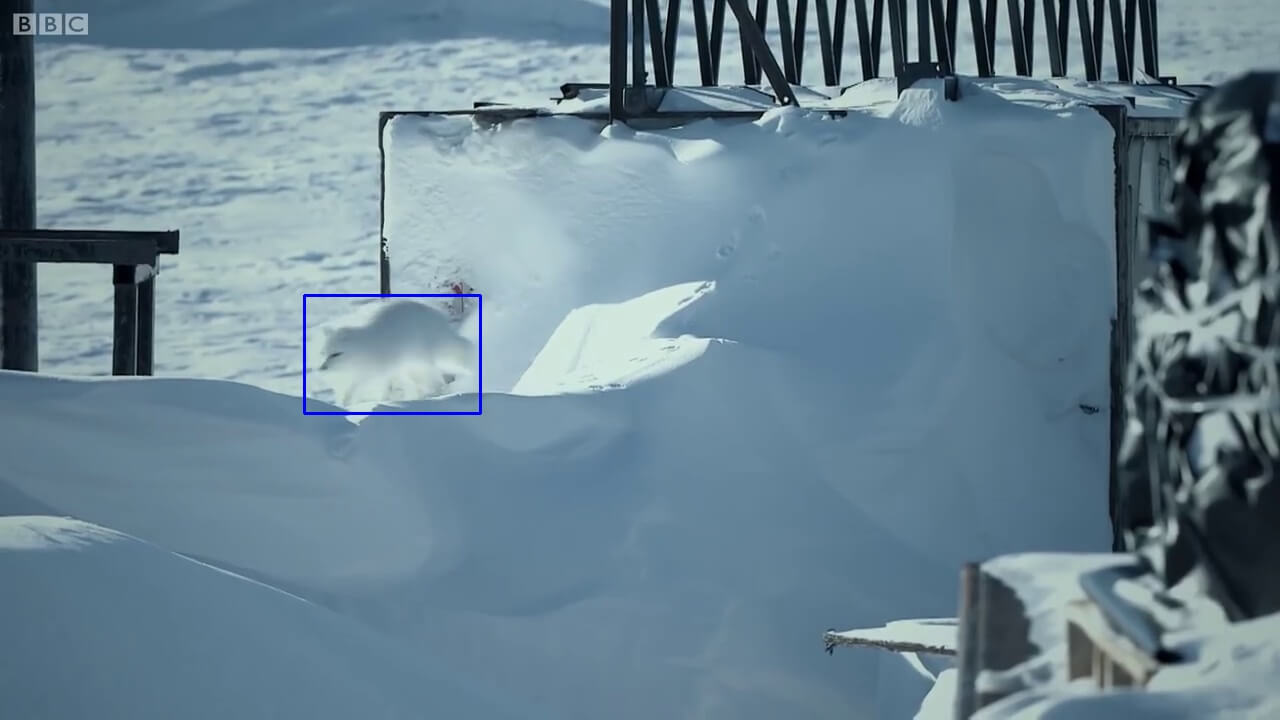} & 
    \includegraphics[width=0.21\linewidth]{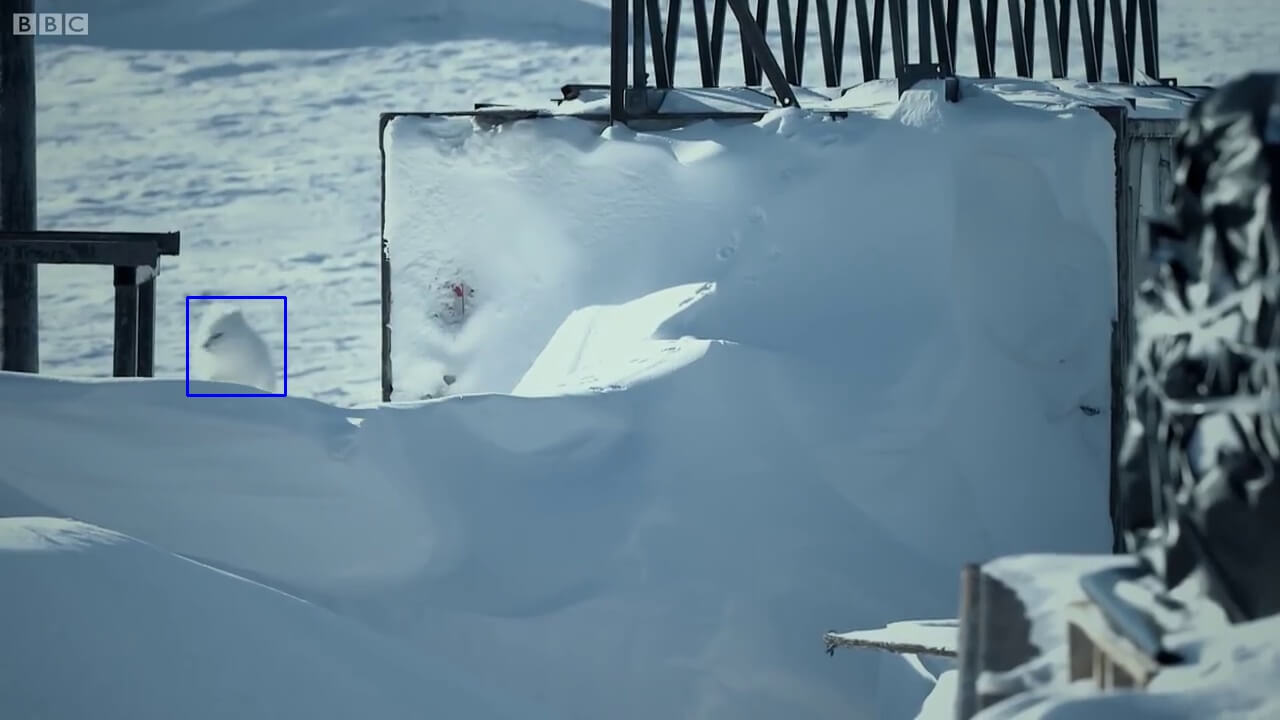} & 
    \includegraphics[width=0.21\linewidth]{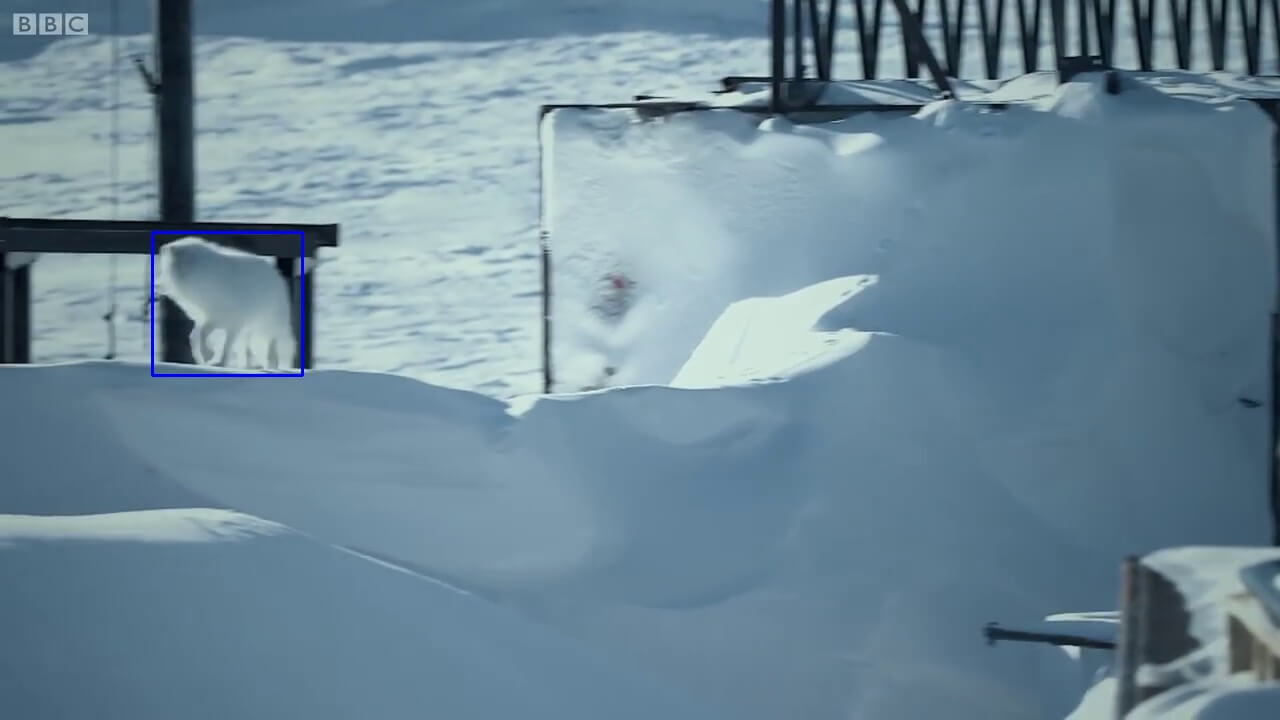} 
    \\
    (b) &
    \includegraphics[width=0.21\linewidth]{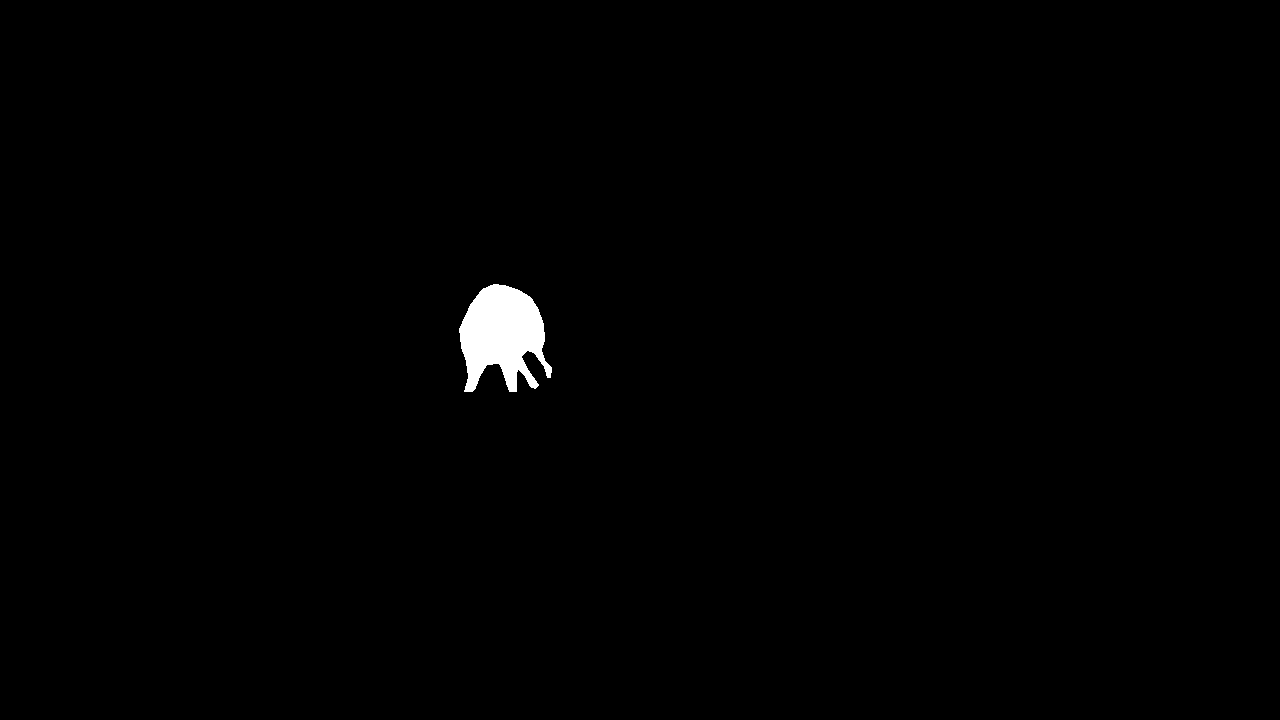} & 
    \includegraphics[width=0.21\linewidth]{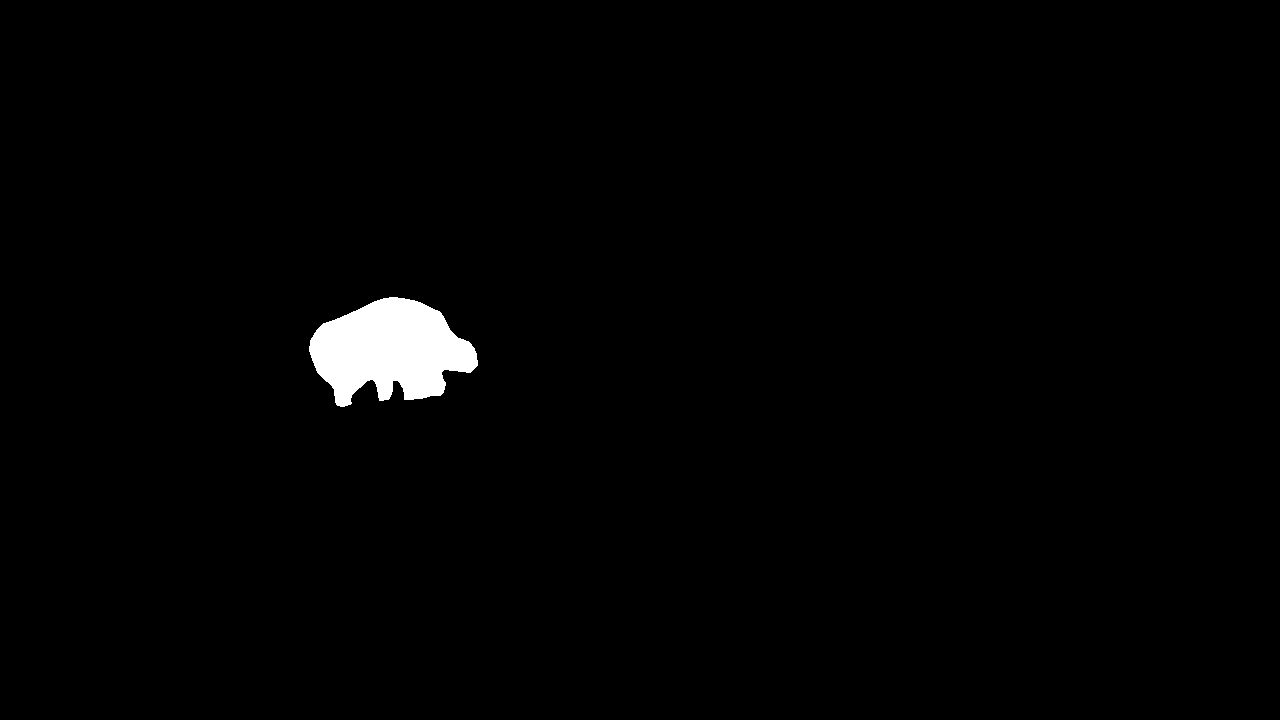} & 
    \includegraphics[width=0.21\linewidth]{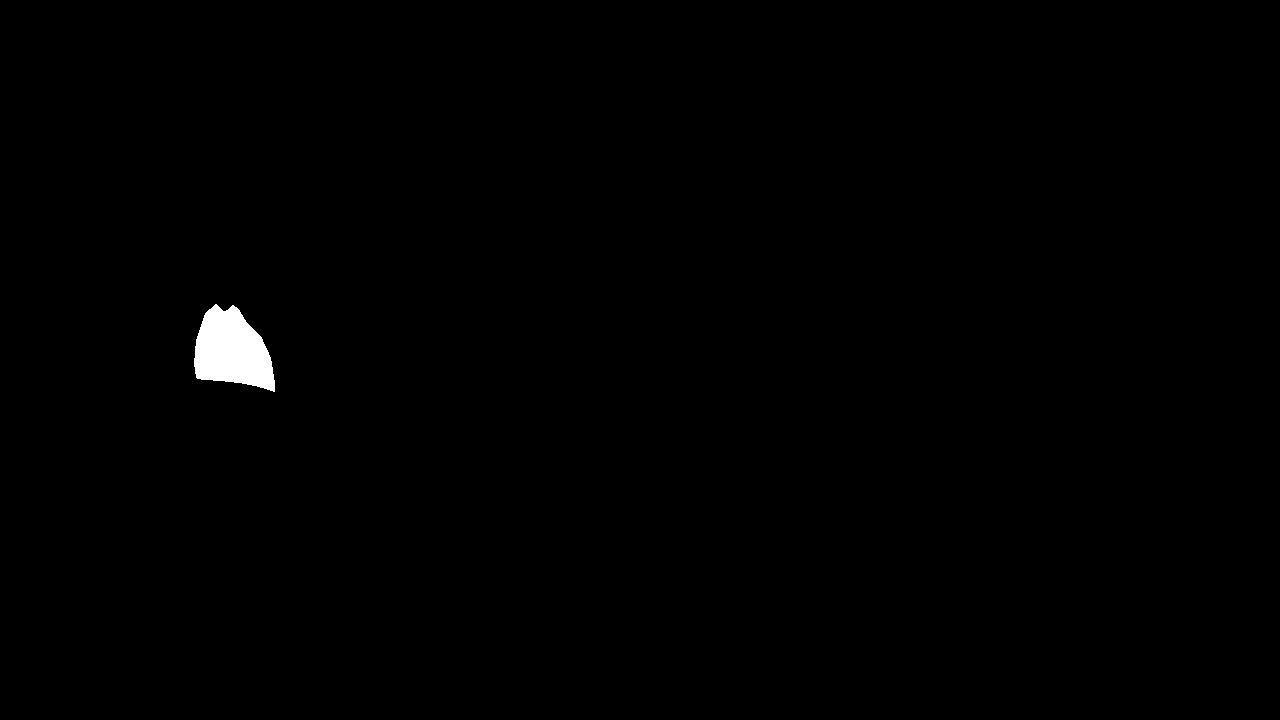} & 
    \includegraphics[width=0.21\linewidth]{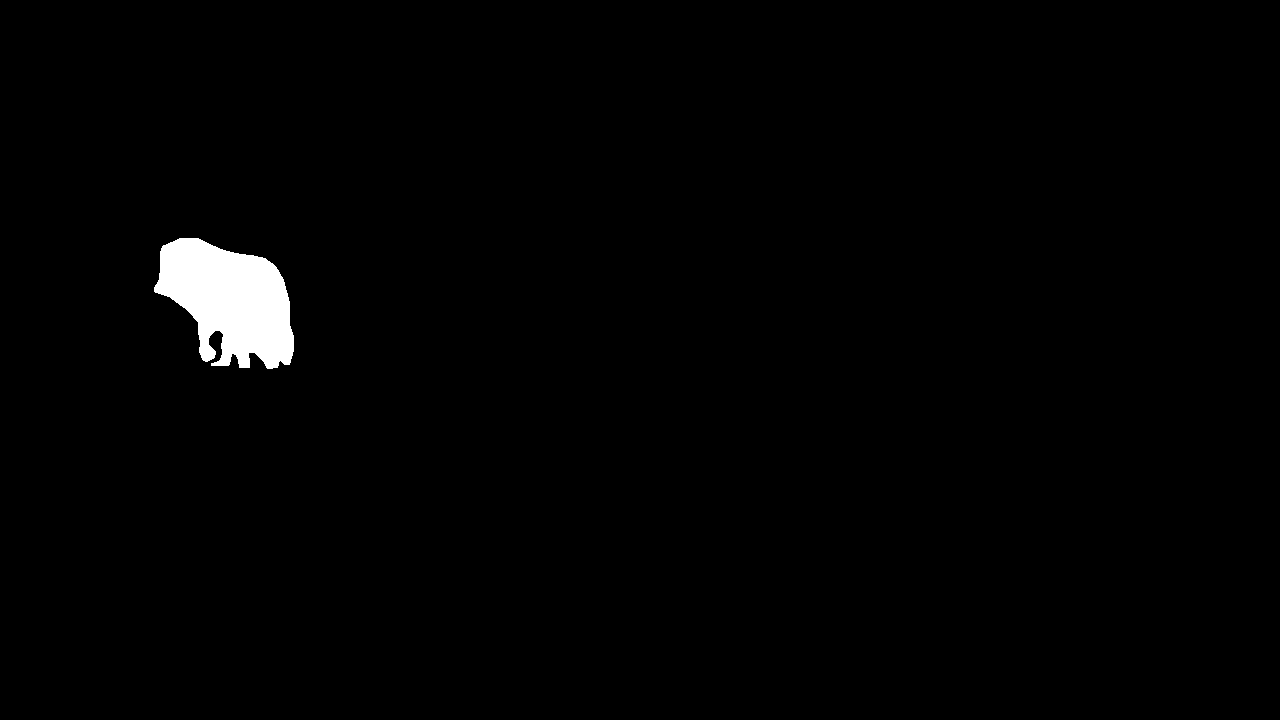} 
    \\
    (c) &
    \includegraphics[width=0.21\linewidth]{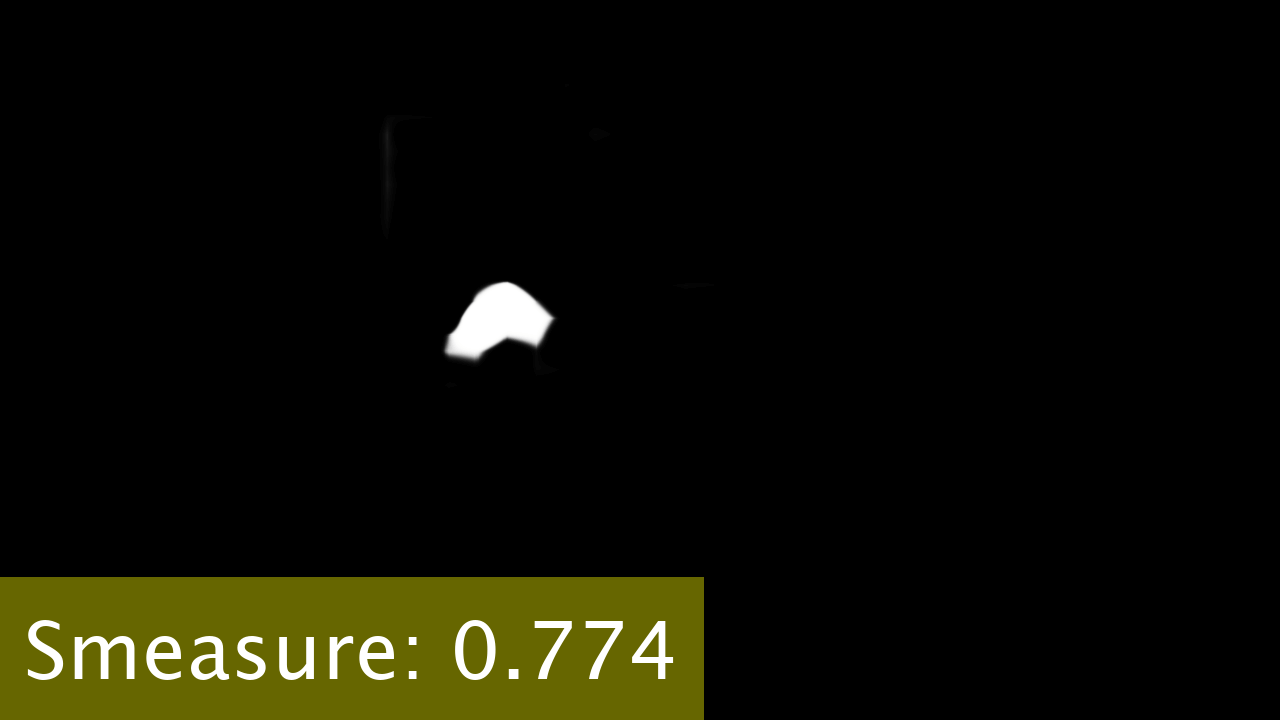} & 
    \includegraphics[width=0.21\linewidth]{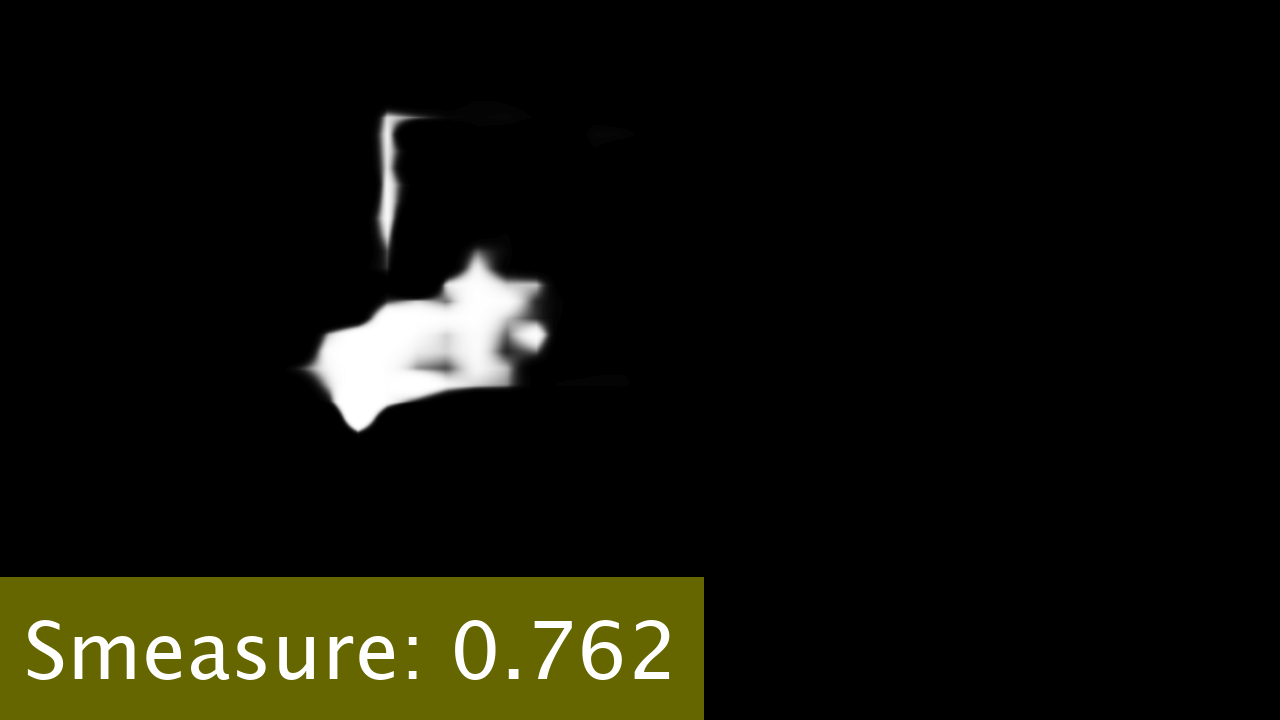} & 
    \includegraphics[width=0.21\linewidth]{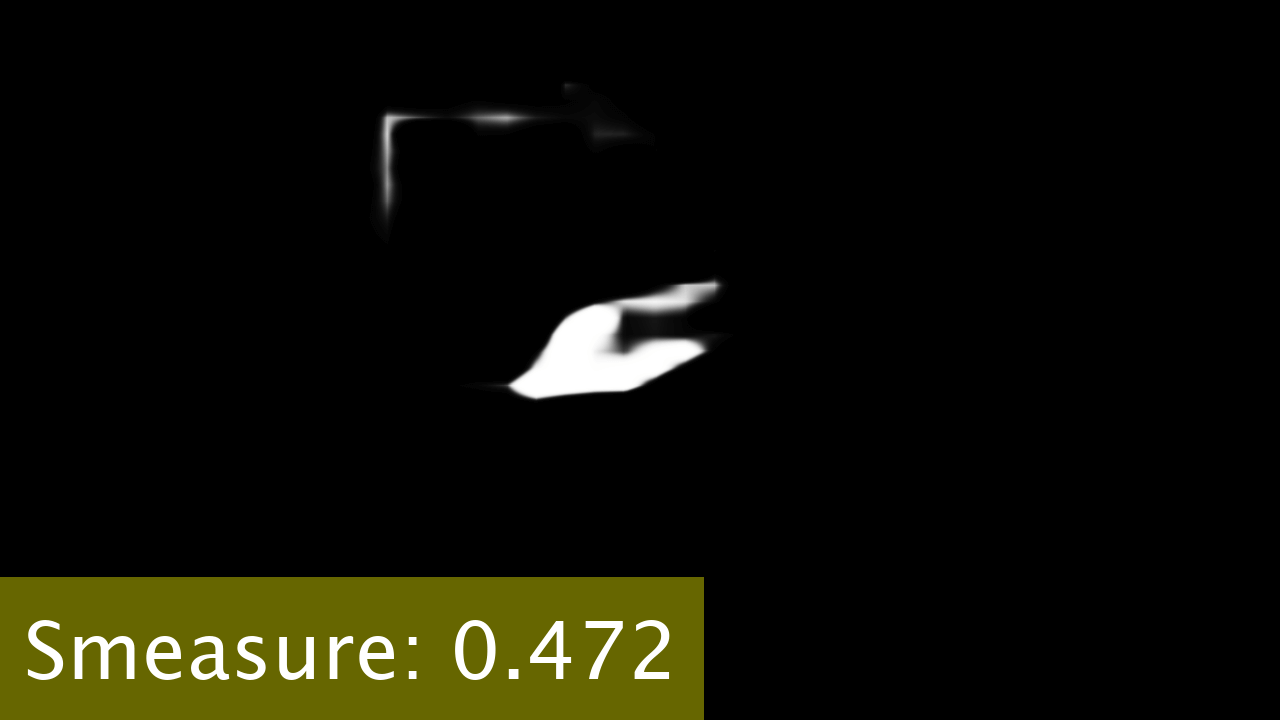} & 
    \includegraphics[width=0.21\linewidth]{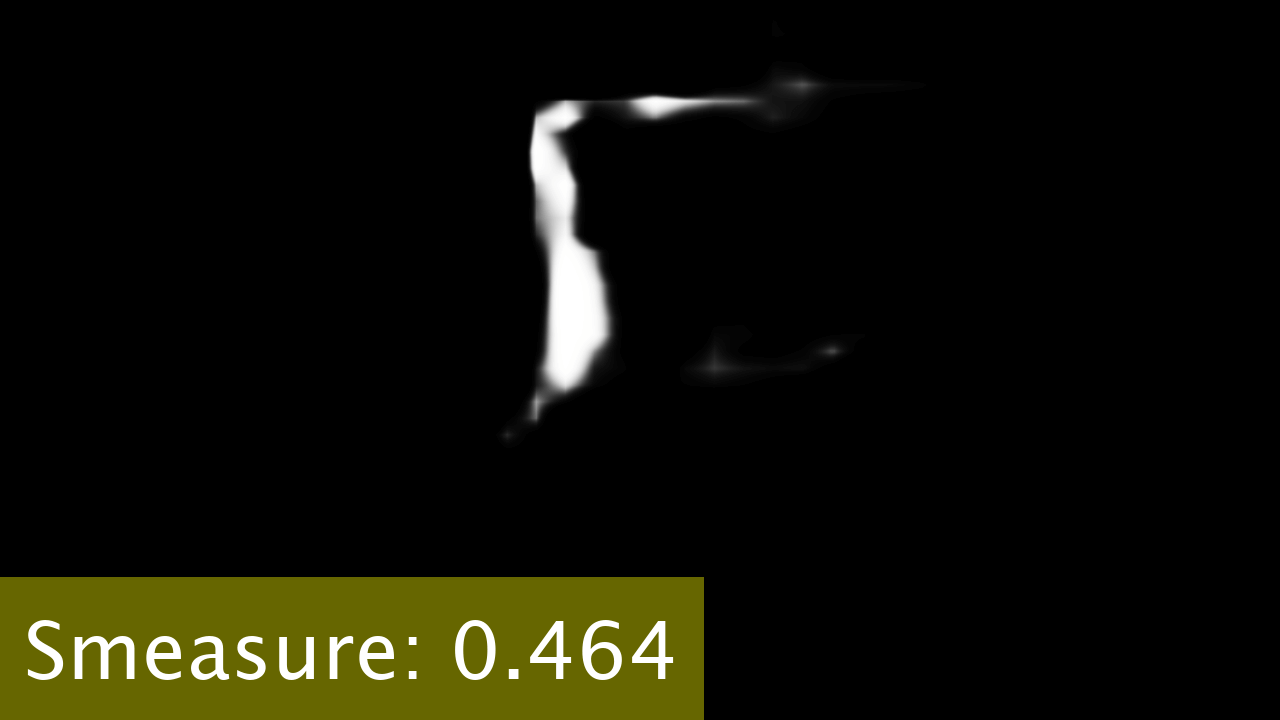} 
    \\
    (d) &
    \includegraphics[width=0.21\linewidth]{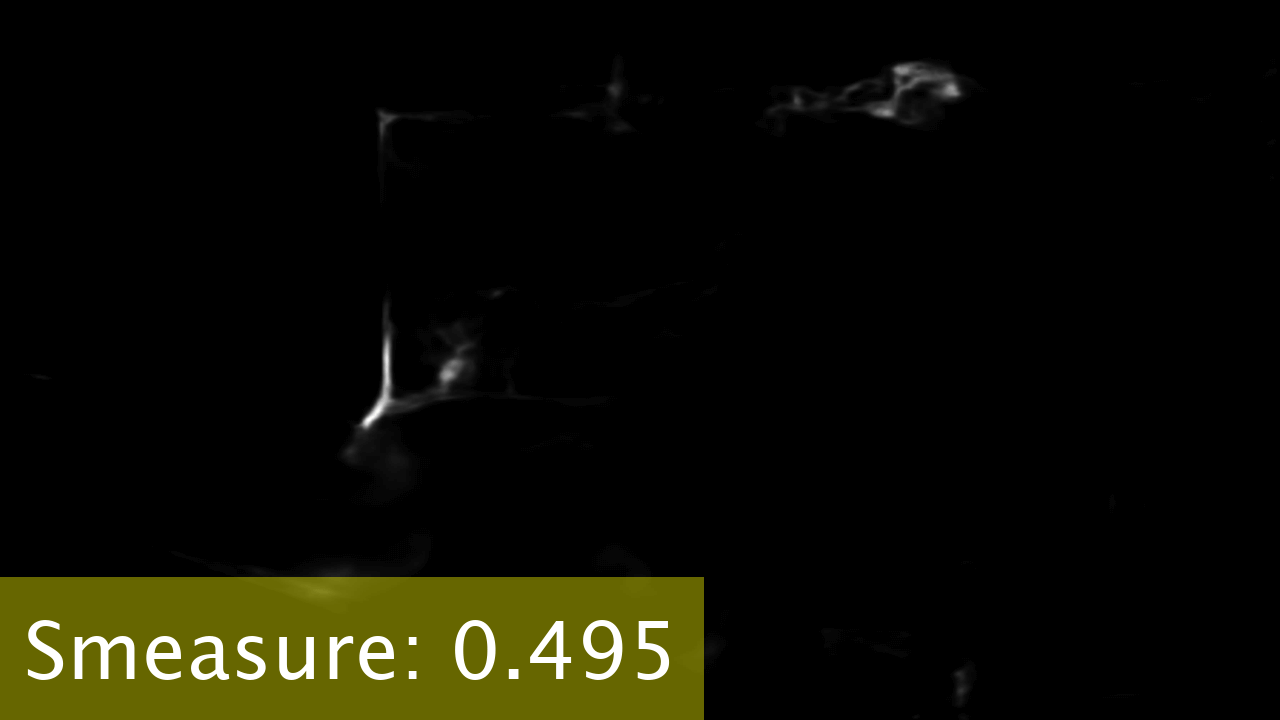} & 
    \includegraphics[width=0.21\linewidth]{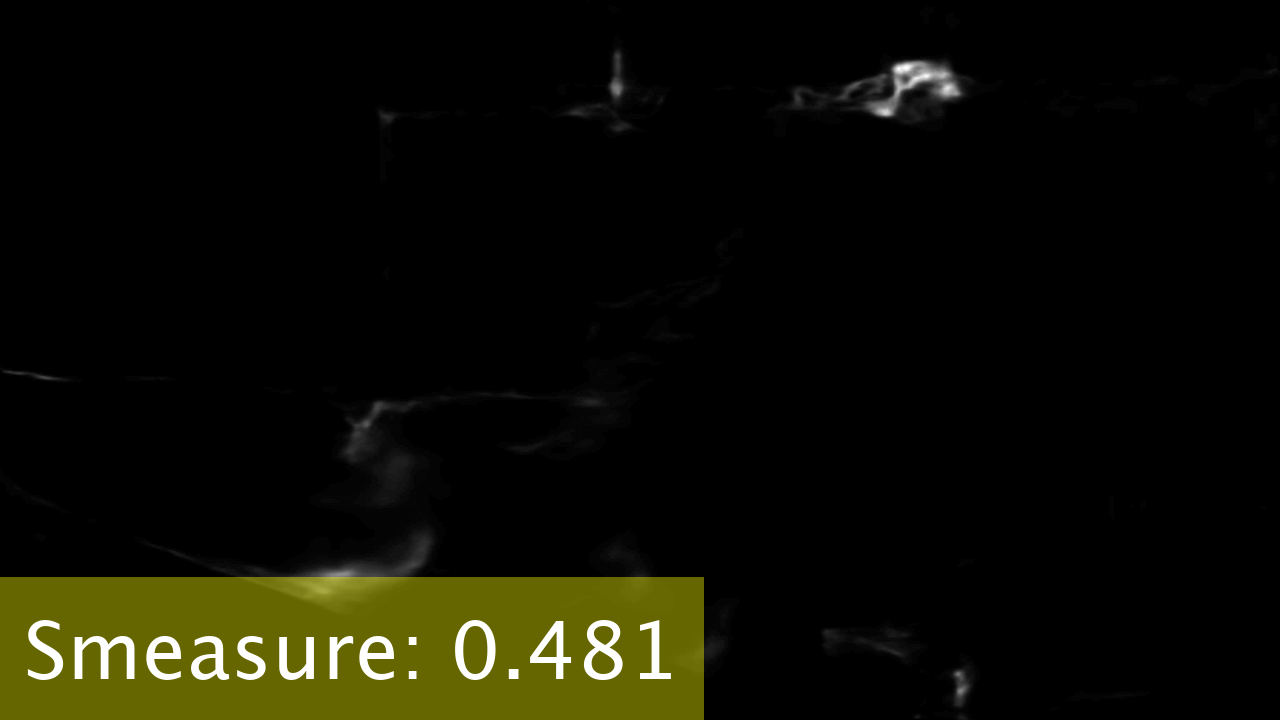} & 
    \includegraphics[width=0.21\linewidth]{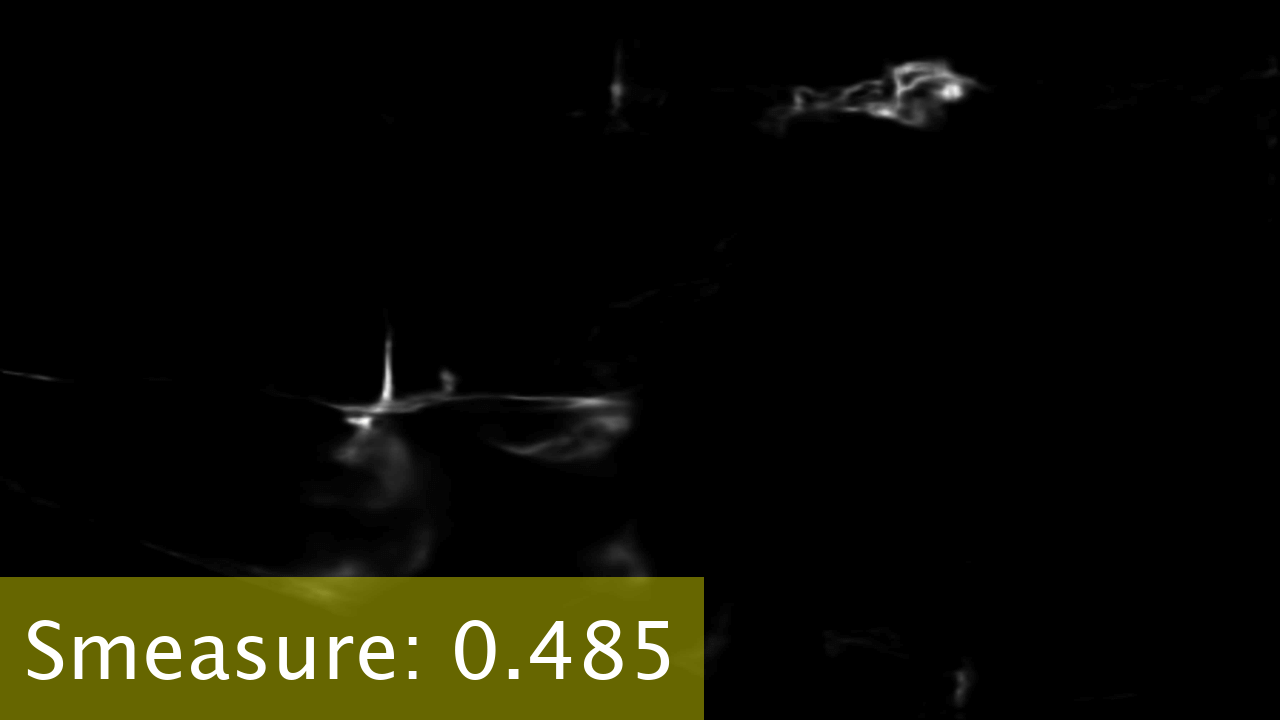} & 
    \includegraphics[width=0.21\linewidth]{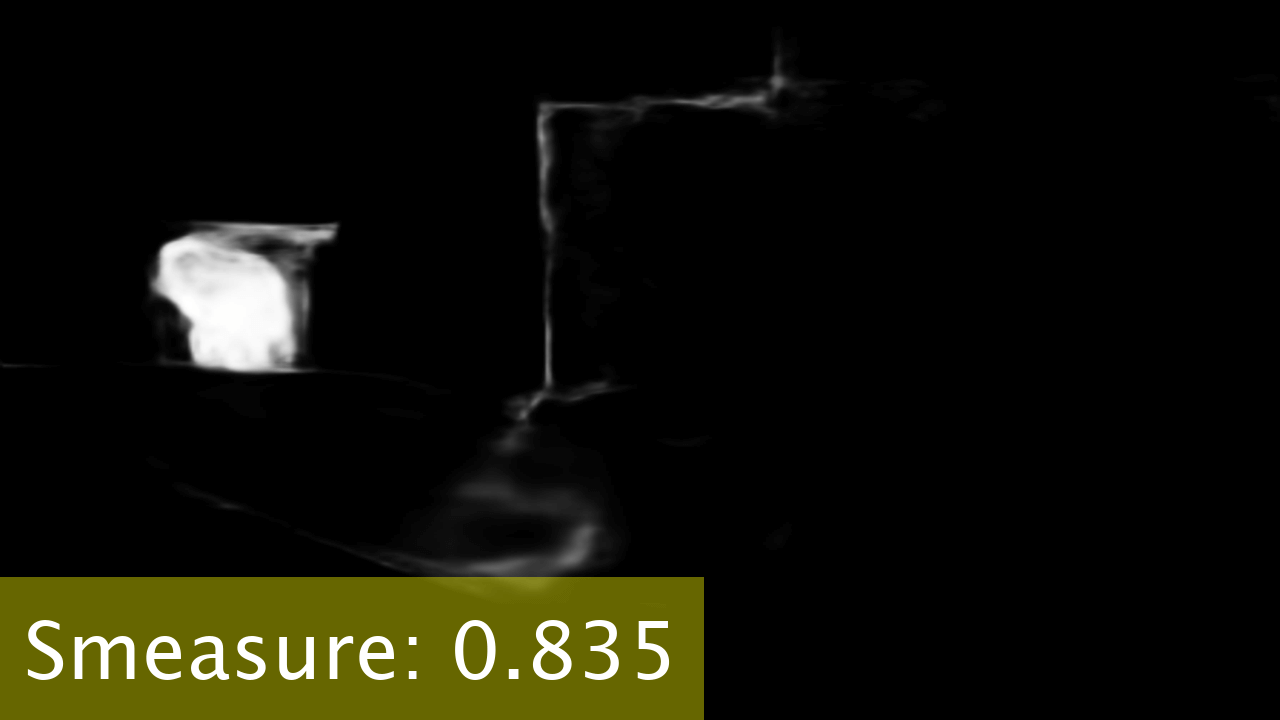}
    \\
    (e) &
    \includegraphics[width=0.21\linewidth]{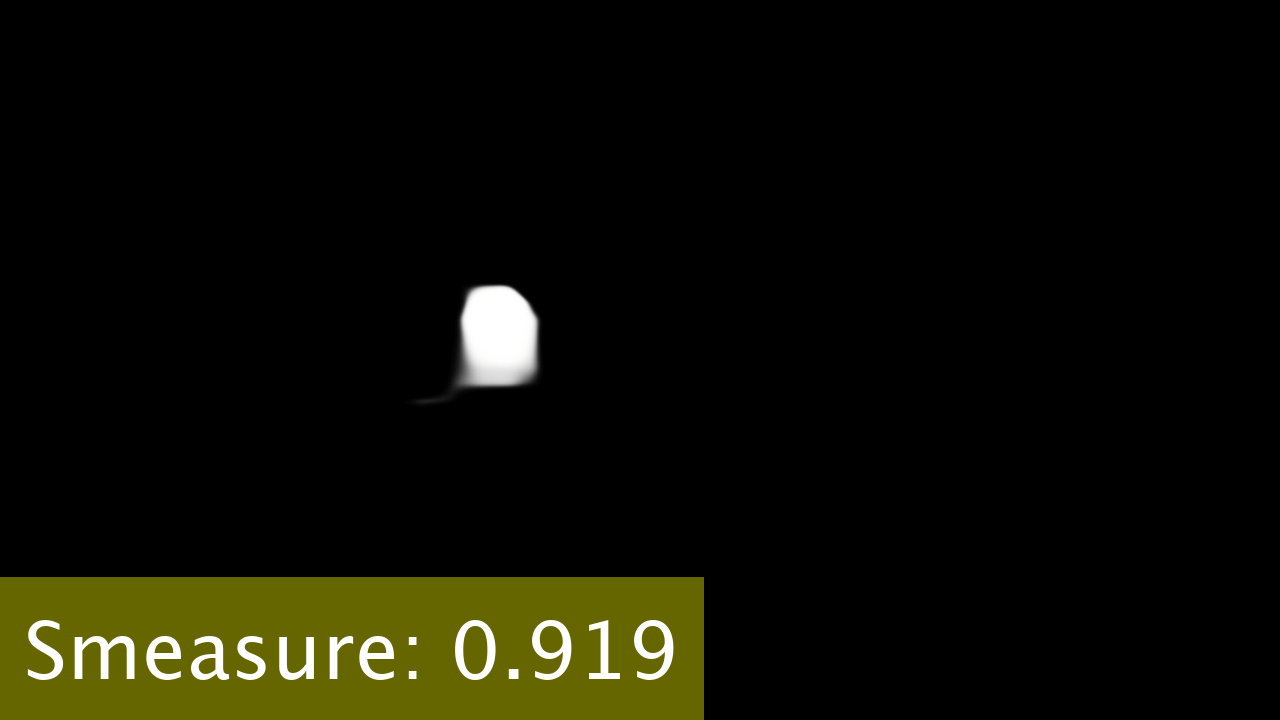} & 
    \includegraphics[width=0.21\linewidth]{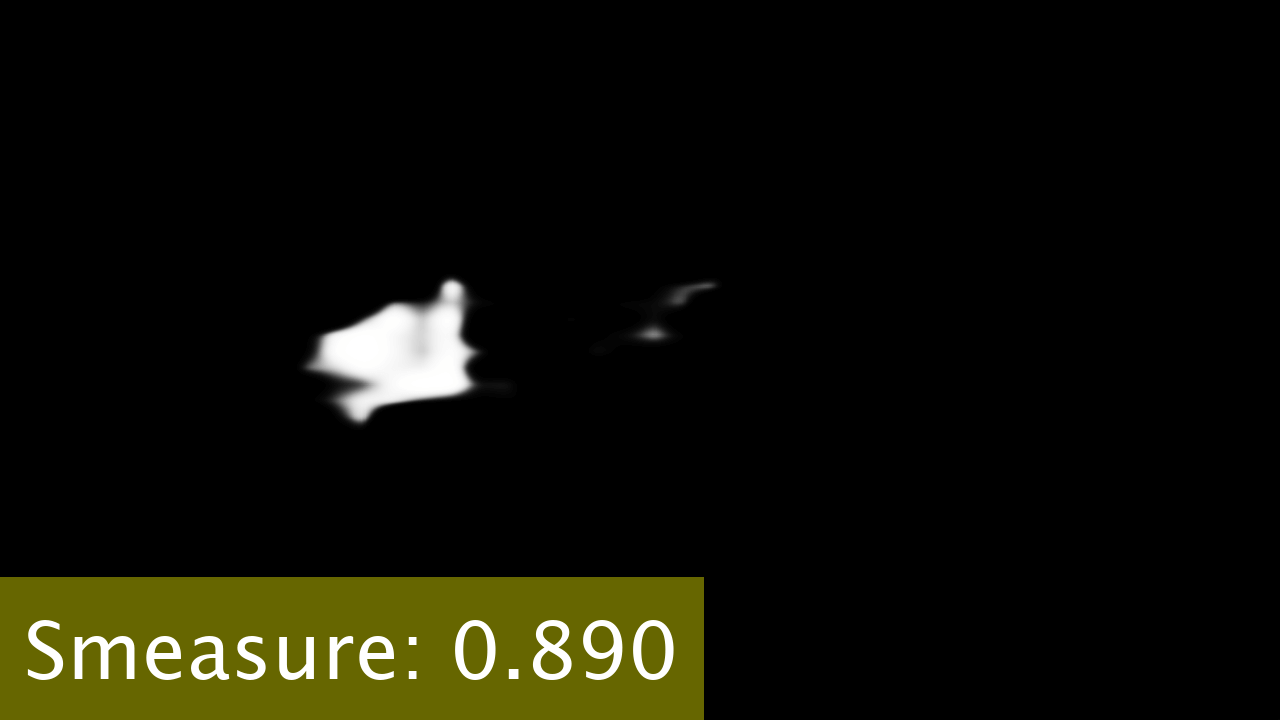} & 
    \includegraphics[width=0.21\linewidth]{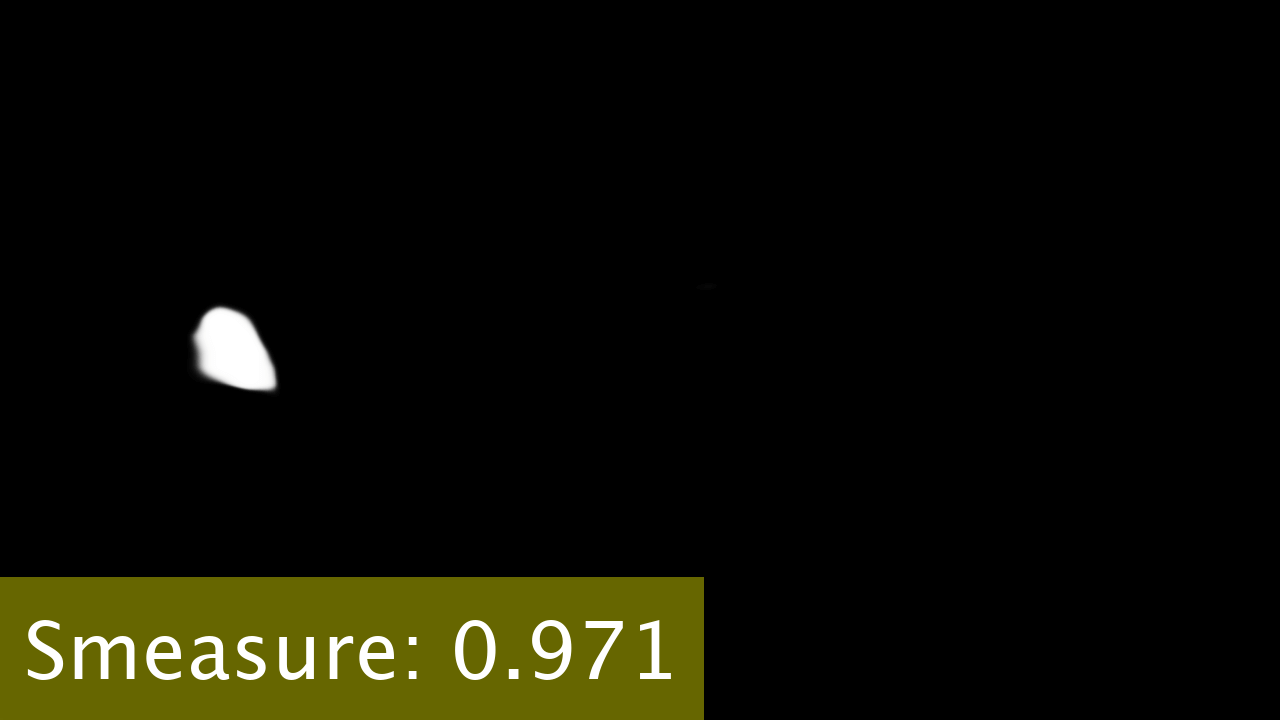} & 
    \includegraphics[width=0.21\linewidth]{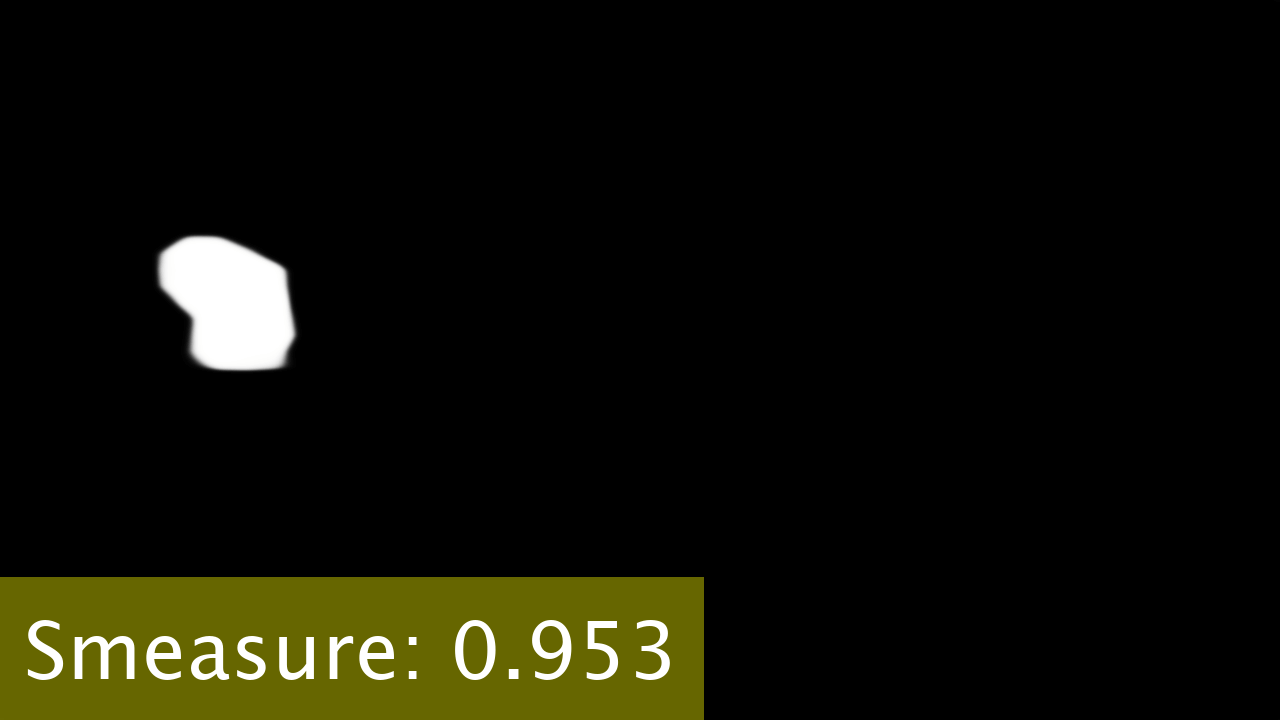} 
    \\
    (f) &
    \includegraphics[width=0.21\linewidth]{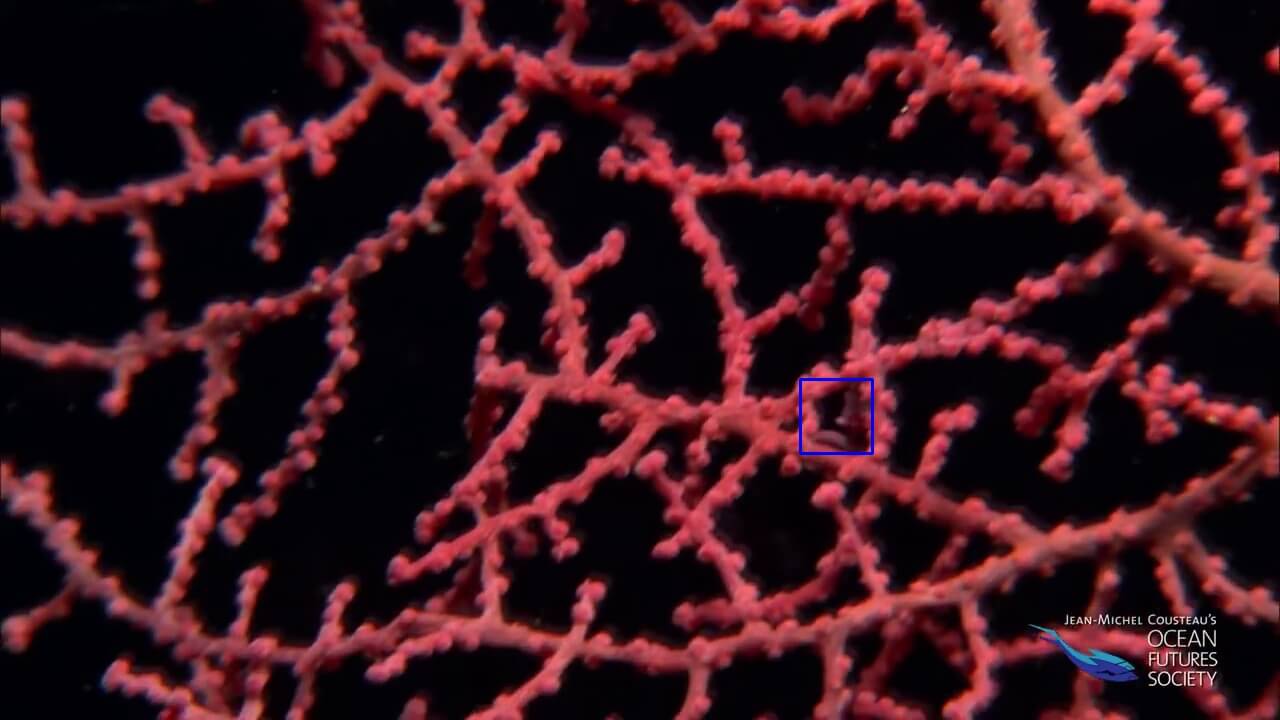} & 
    \includegraphics[width=0.21\linewidth]{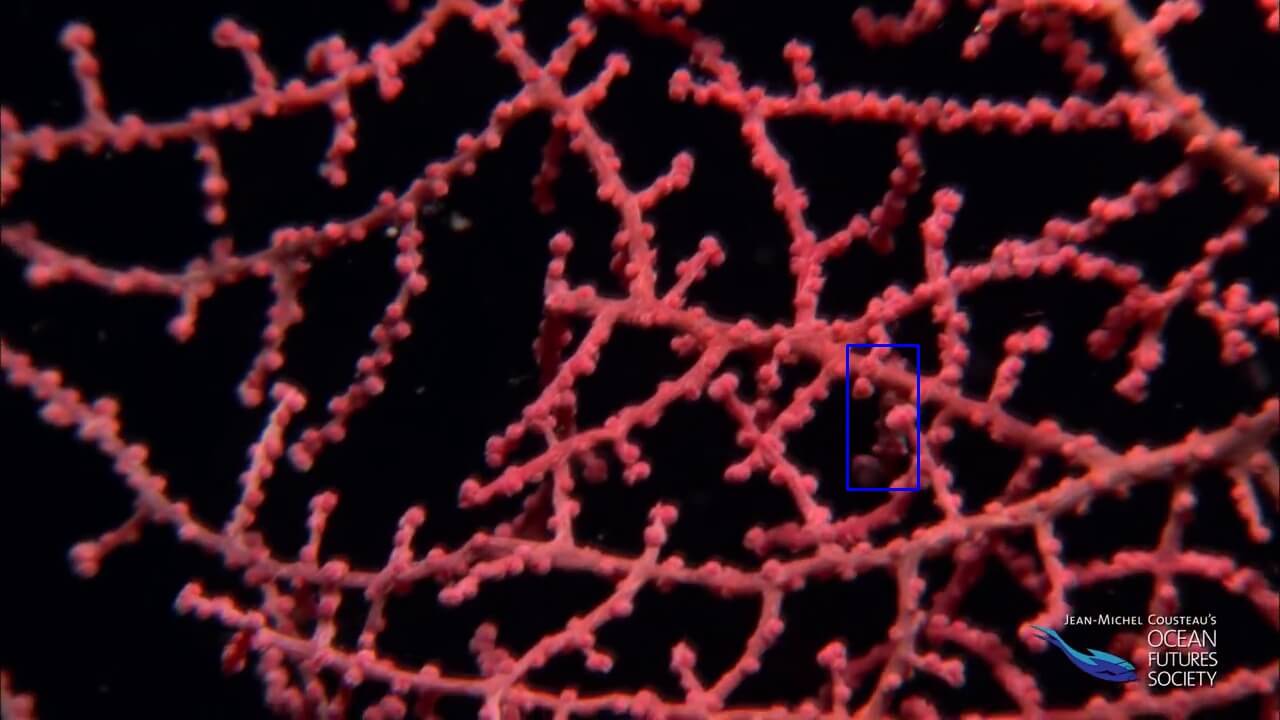} & 
    \includegraphics[width=0.21\linewidth]{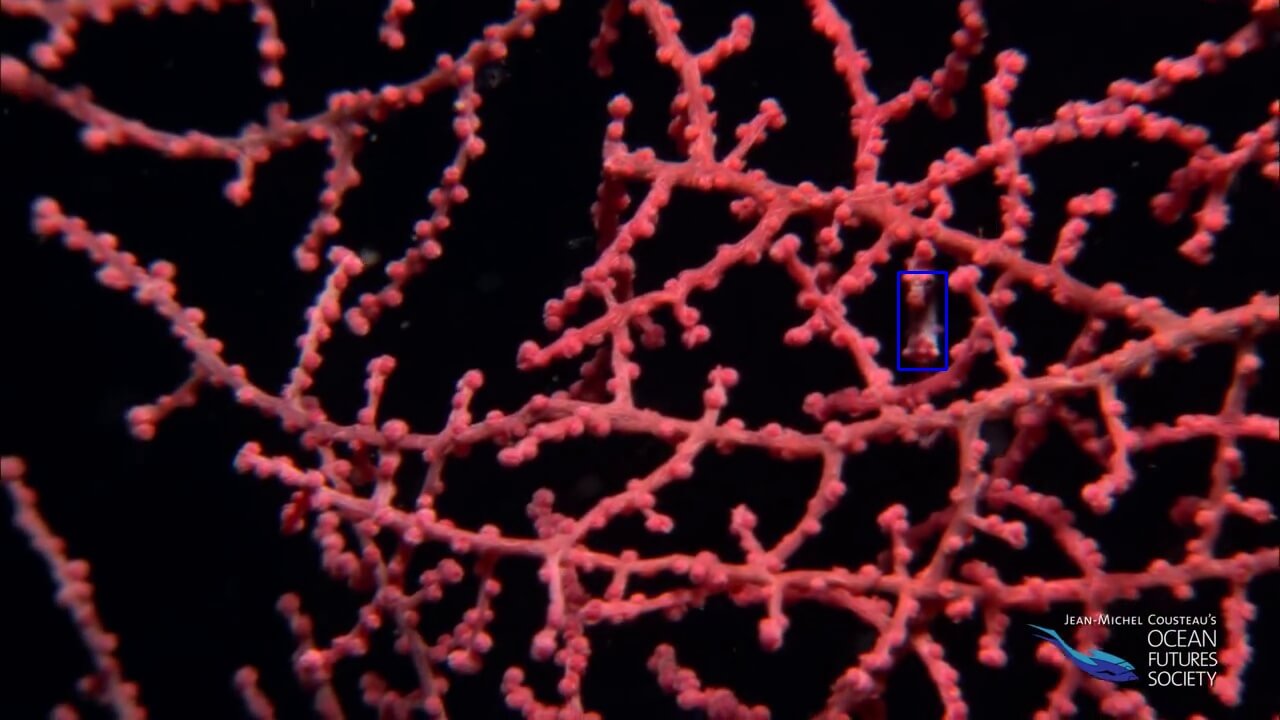} & 
    \includegraphics[width=0.21\linewidth]{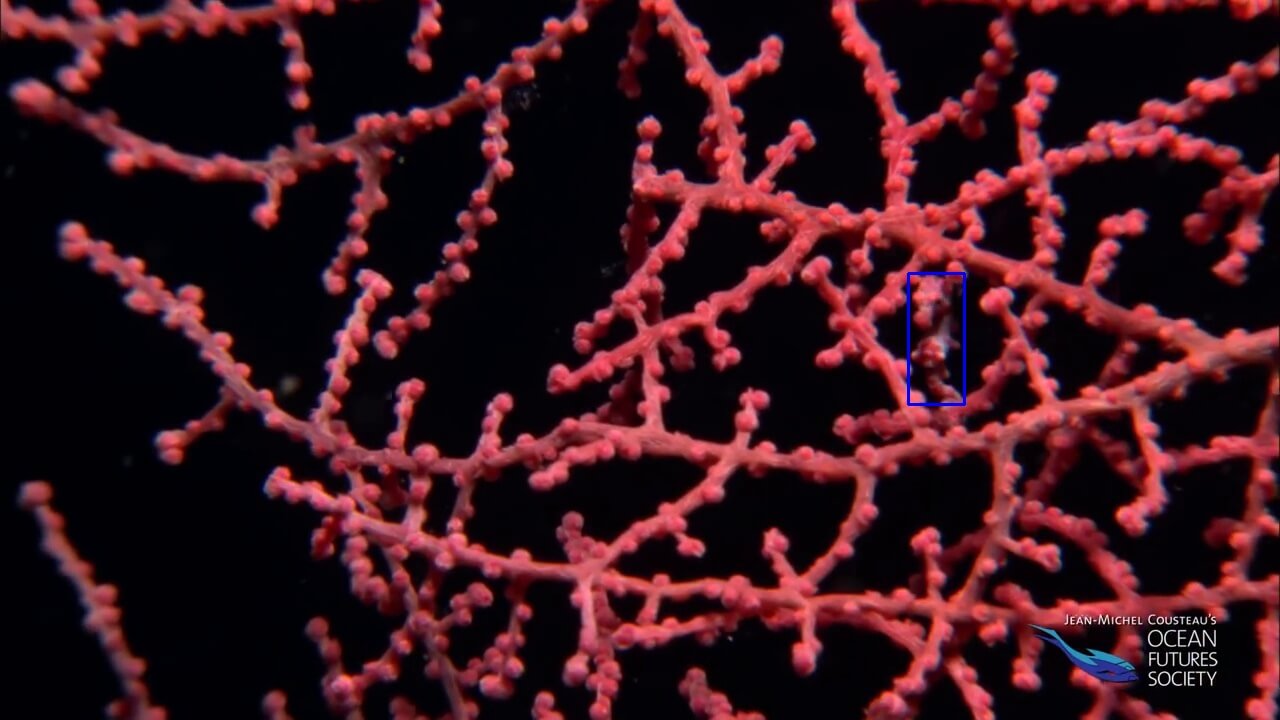} 
    \\
    (g) &
    \includegraphics[width=0.21\linewidth]{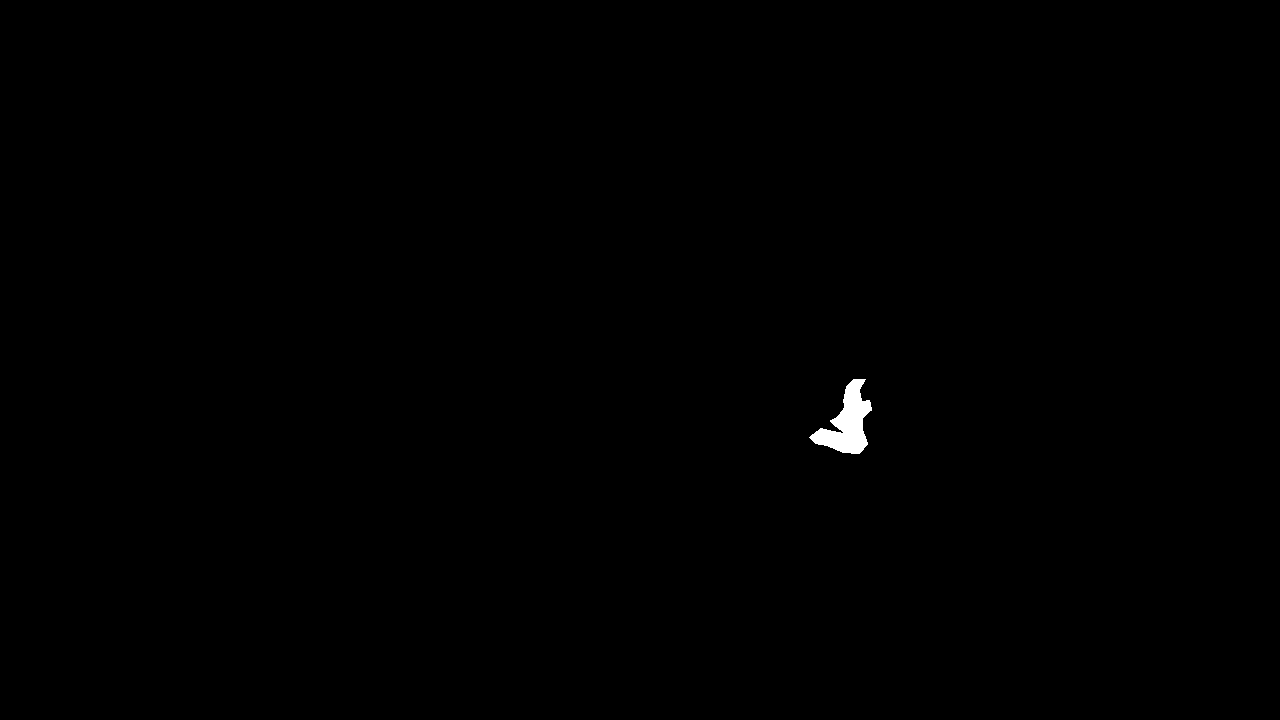}& 
    \includegraphics[width=0.21\linewidth]{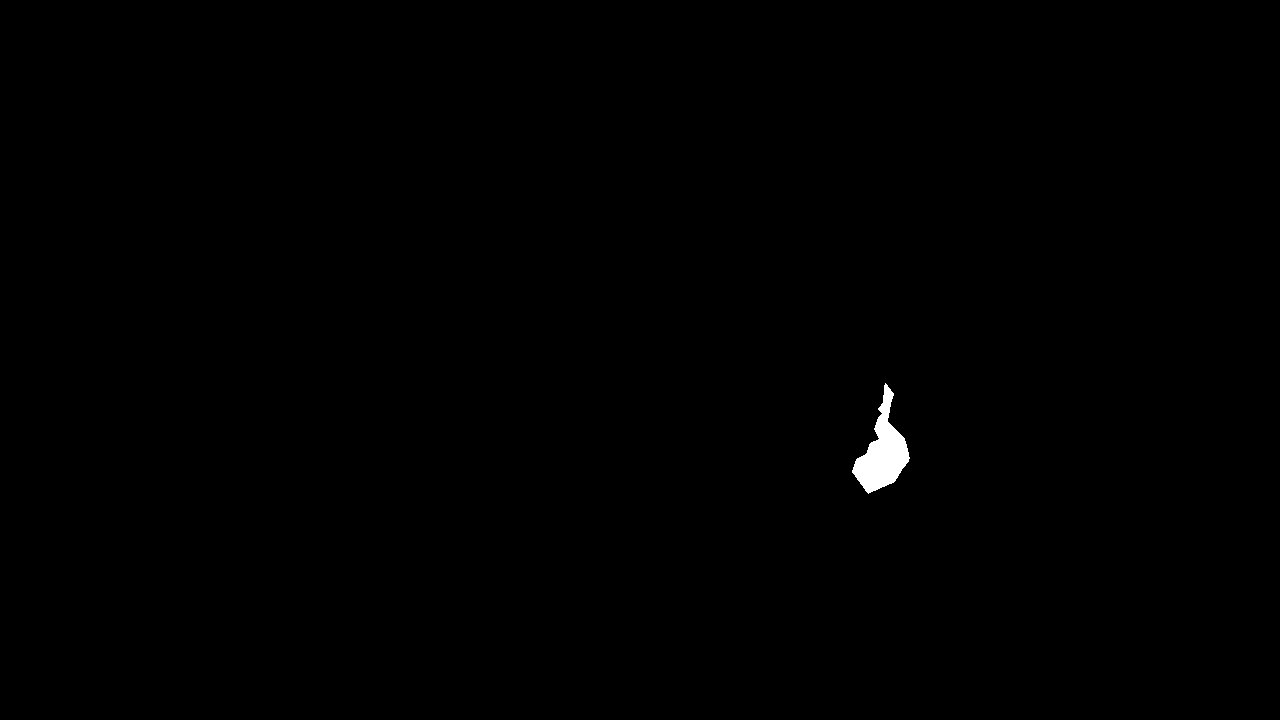} & 
    \includegraphics[width=0.21\linewidth]{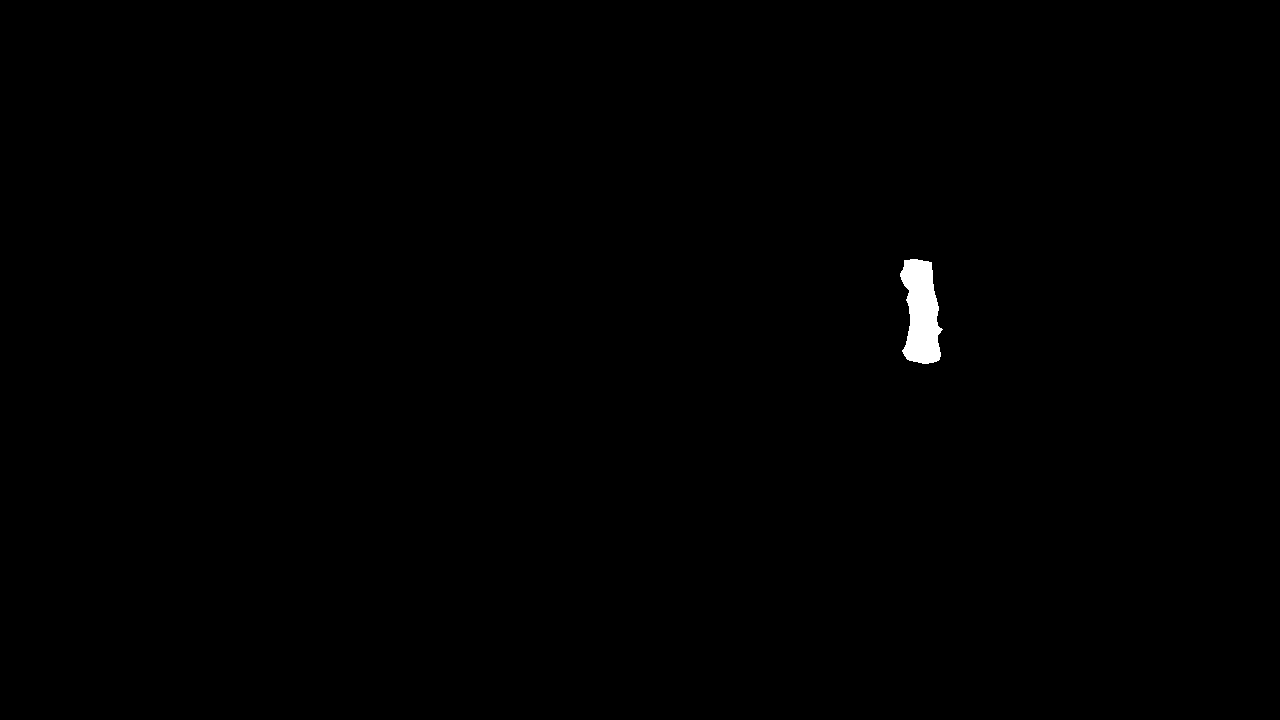} & 
    \includegraphics[width=0.21\linewidth]{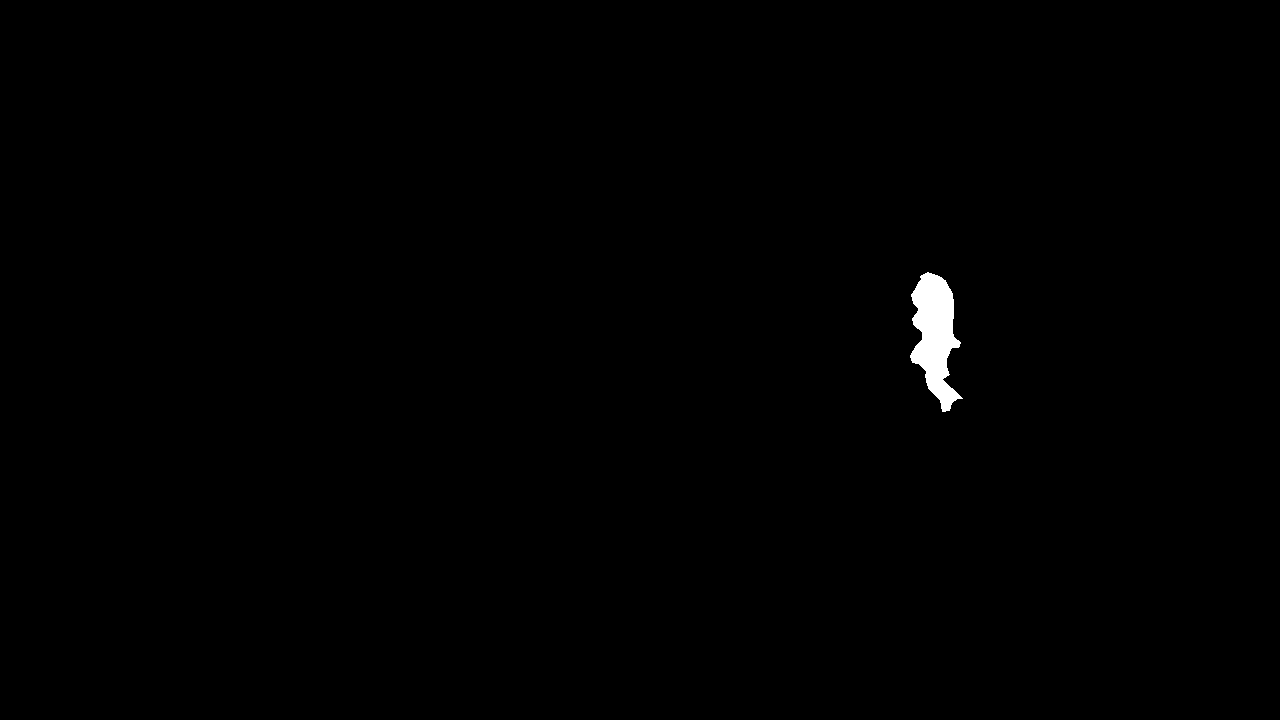} 
    \\
    (h) &
    \includegraphics[width=0.21\linewidth]{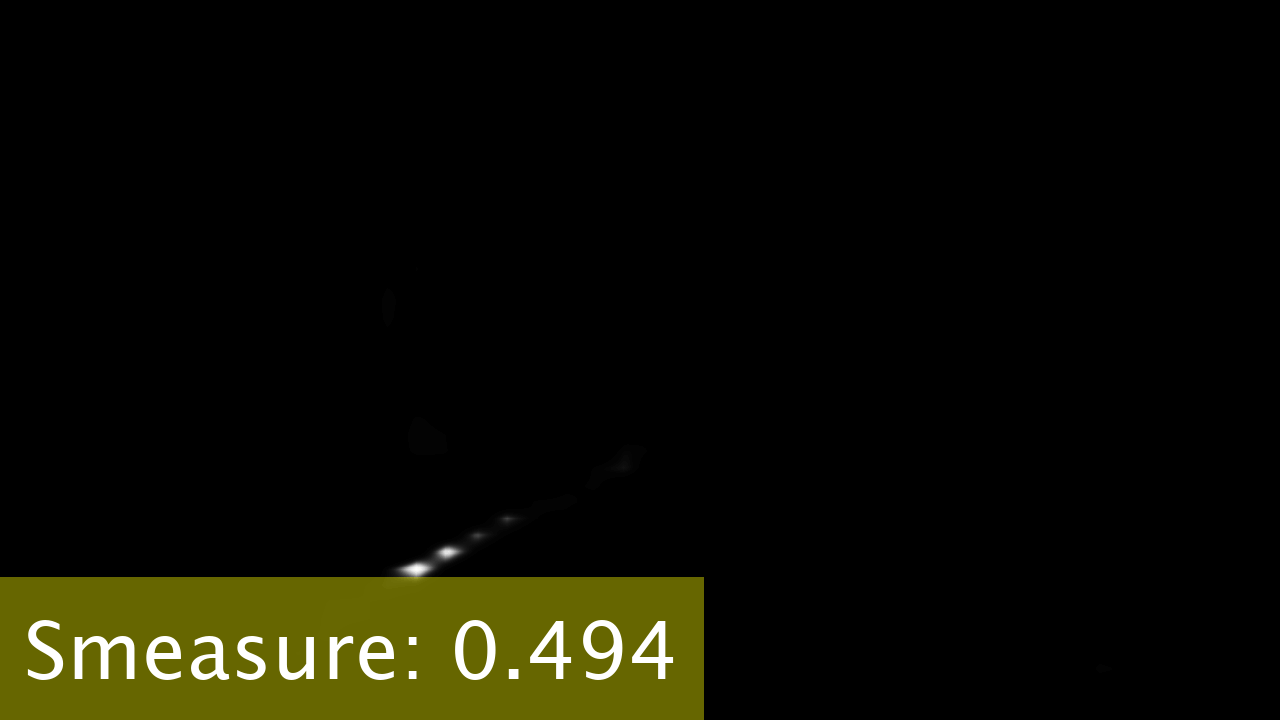}& 
    \includegraphics[width=0.21\linewidth]{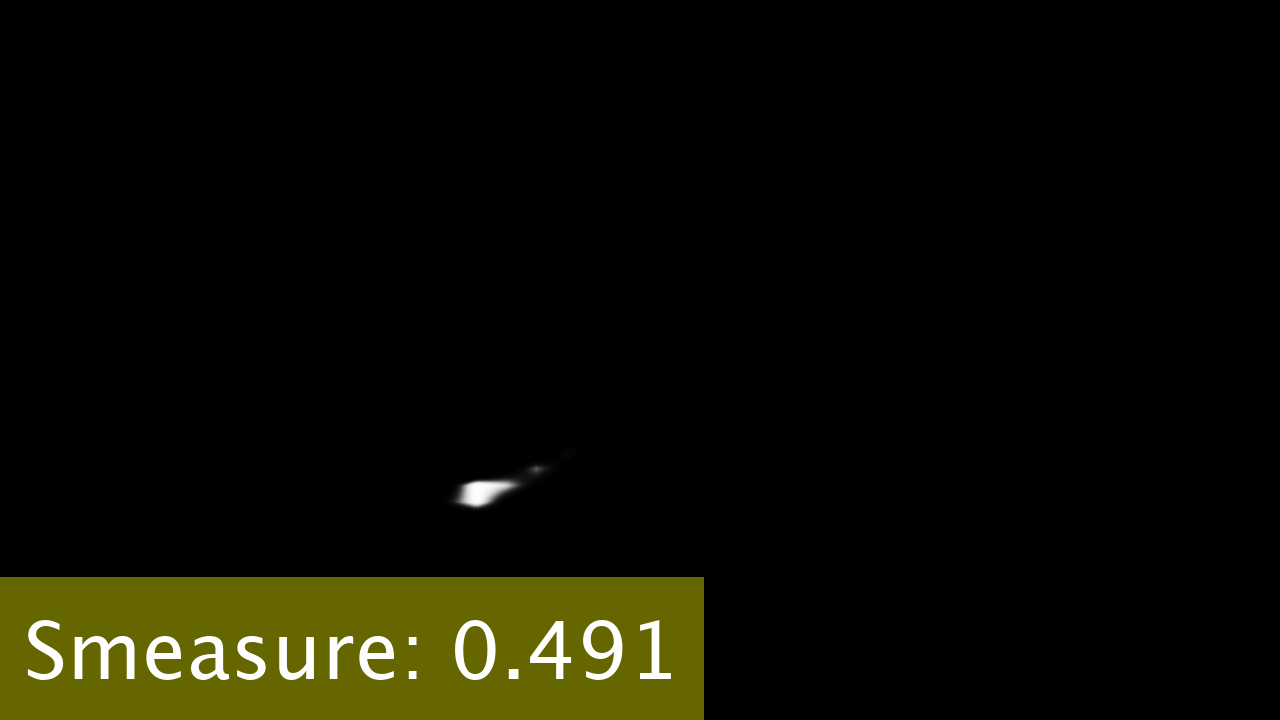} & 
    \includegraphics[width=0.21\linewidth]{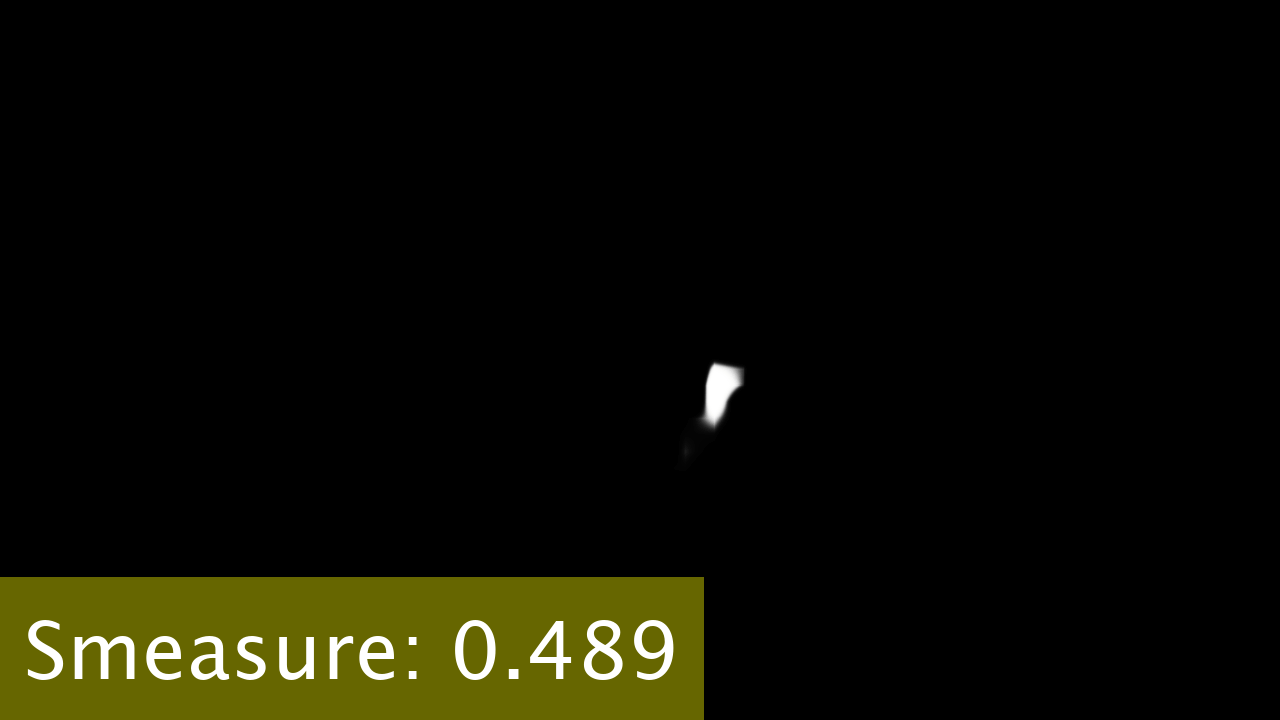} & 
    \includegraphics[width=0.21\linewidth]{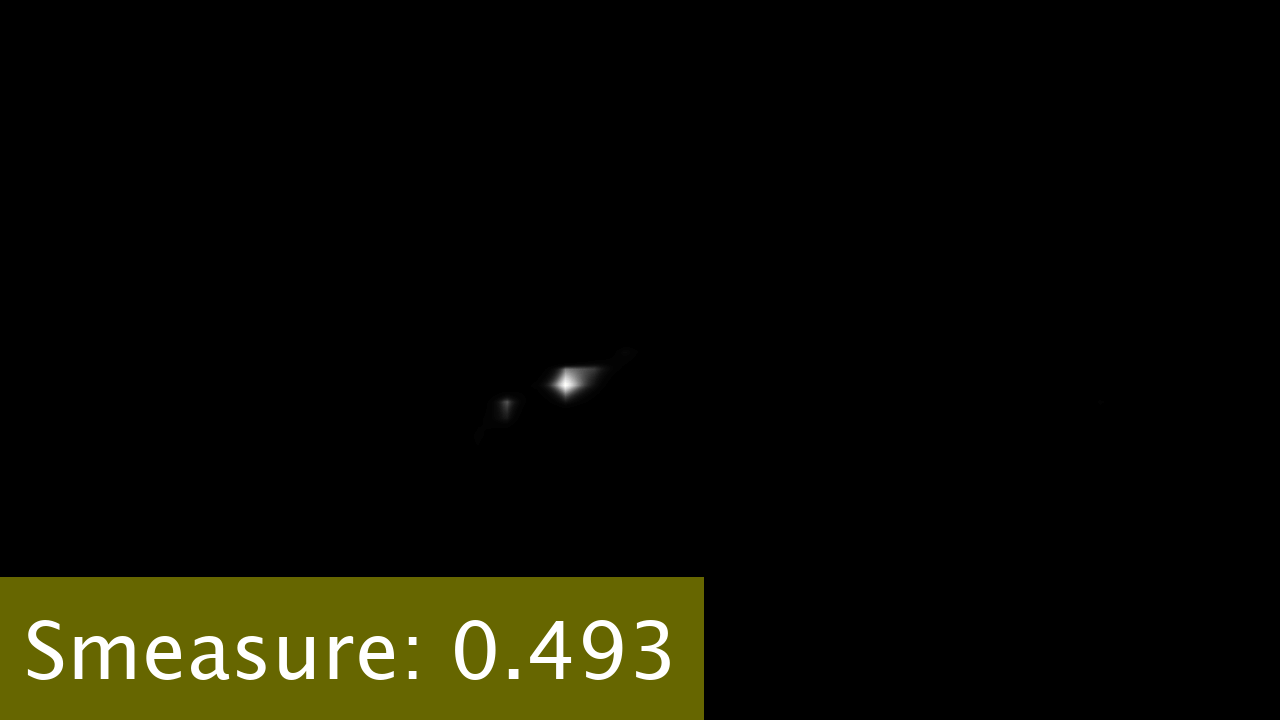} 
    \\
    (i) &
    \includegraphics[width=0.21\linewidth]{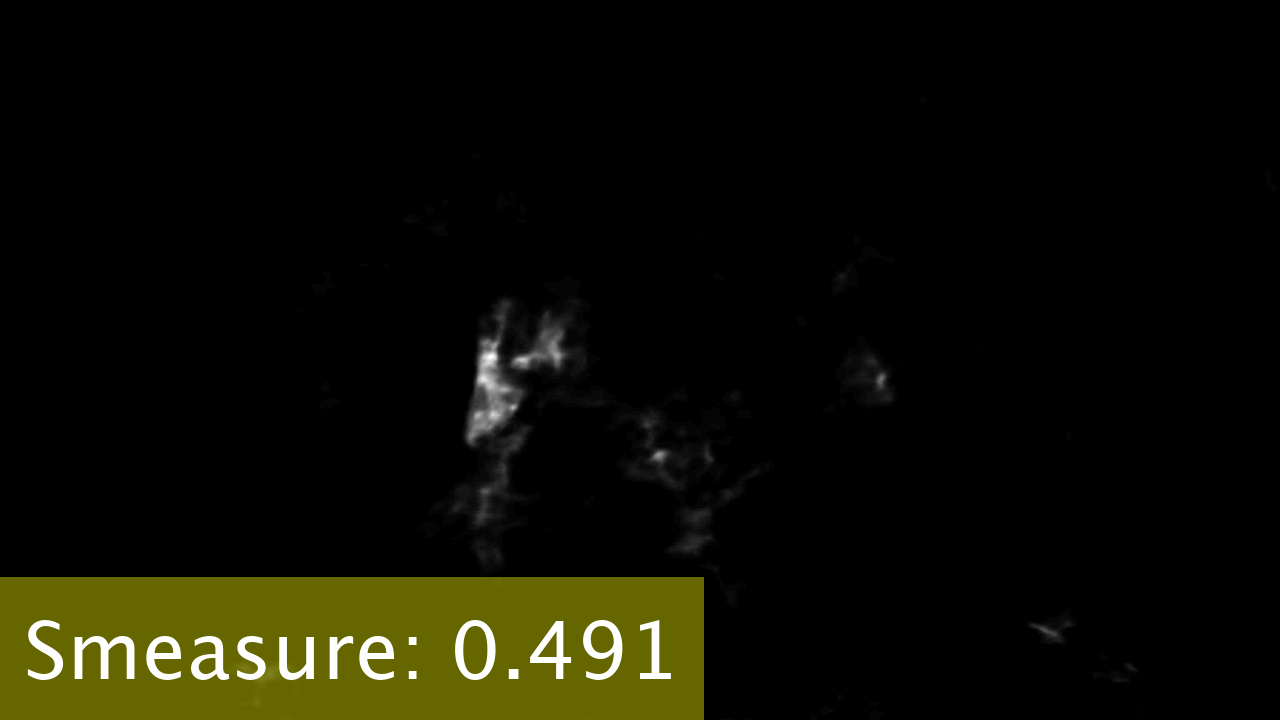}& 
    \includegraphics[width=0.21\linewidth]{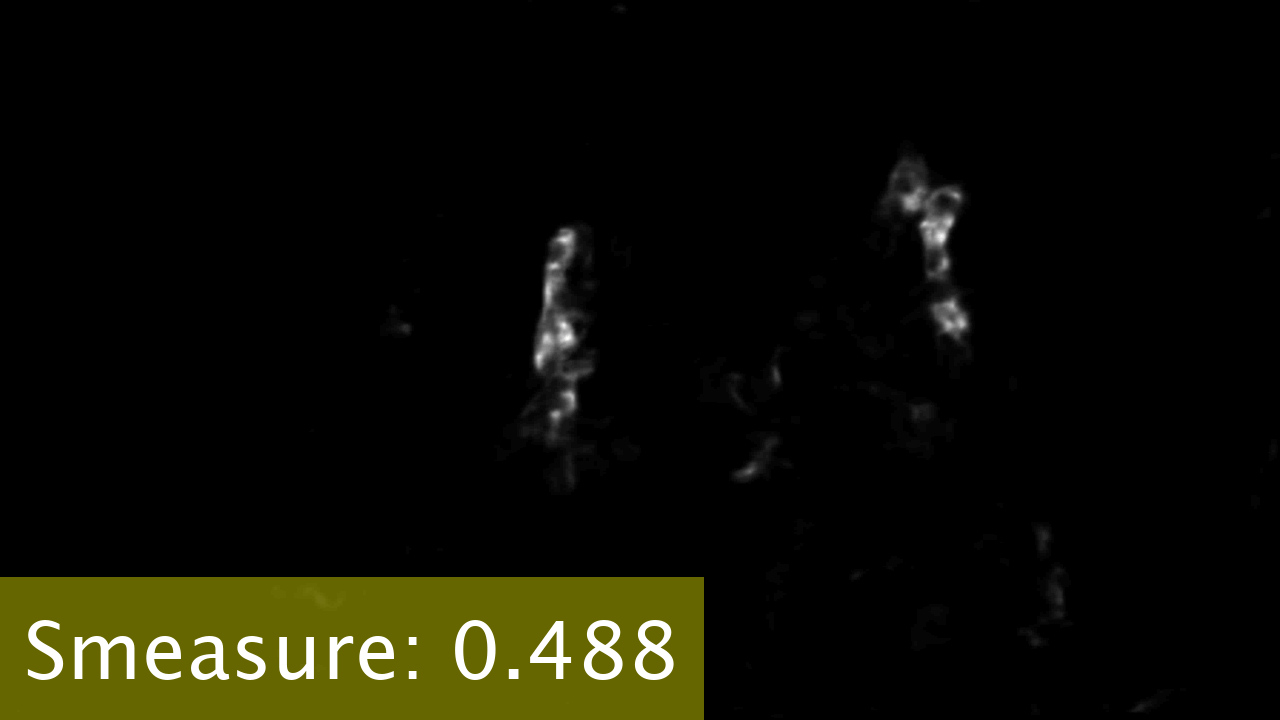} & 
    \includegraphics[width=0.21\linewidth]{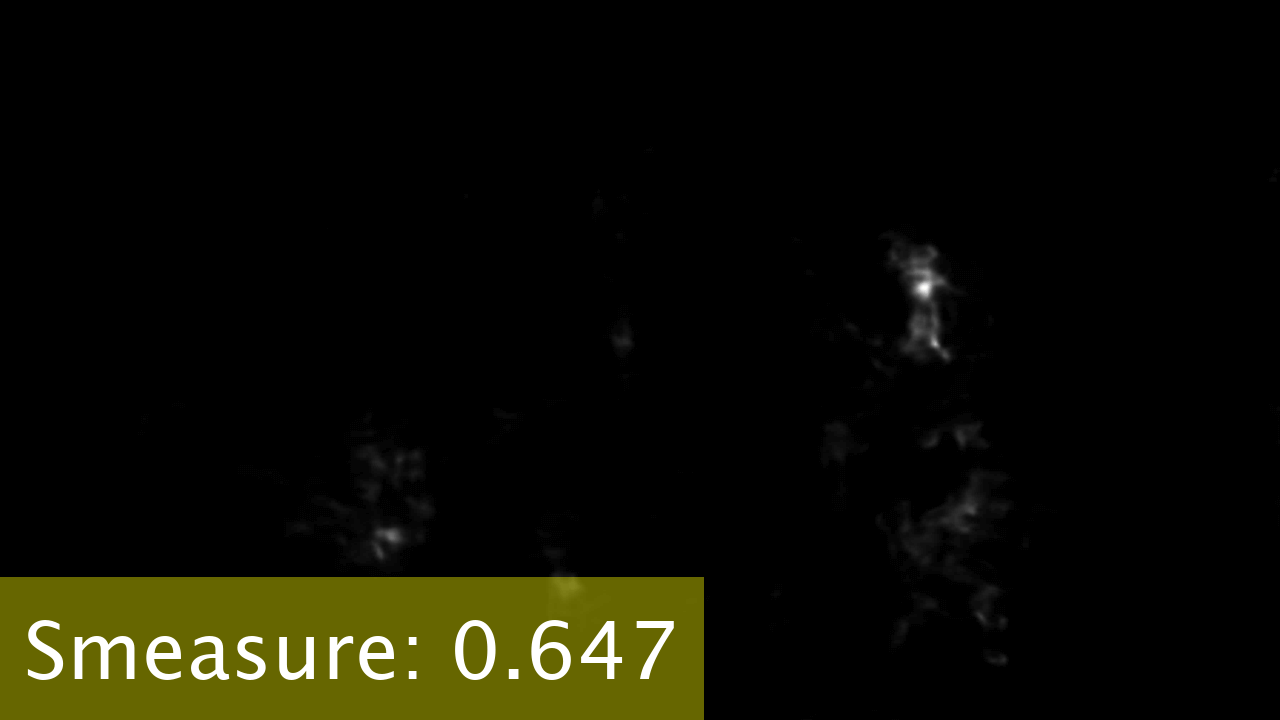} & 
    \includegraphics[width=0.21\linewidth]{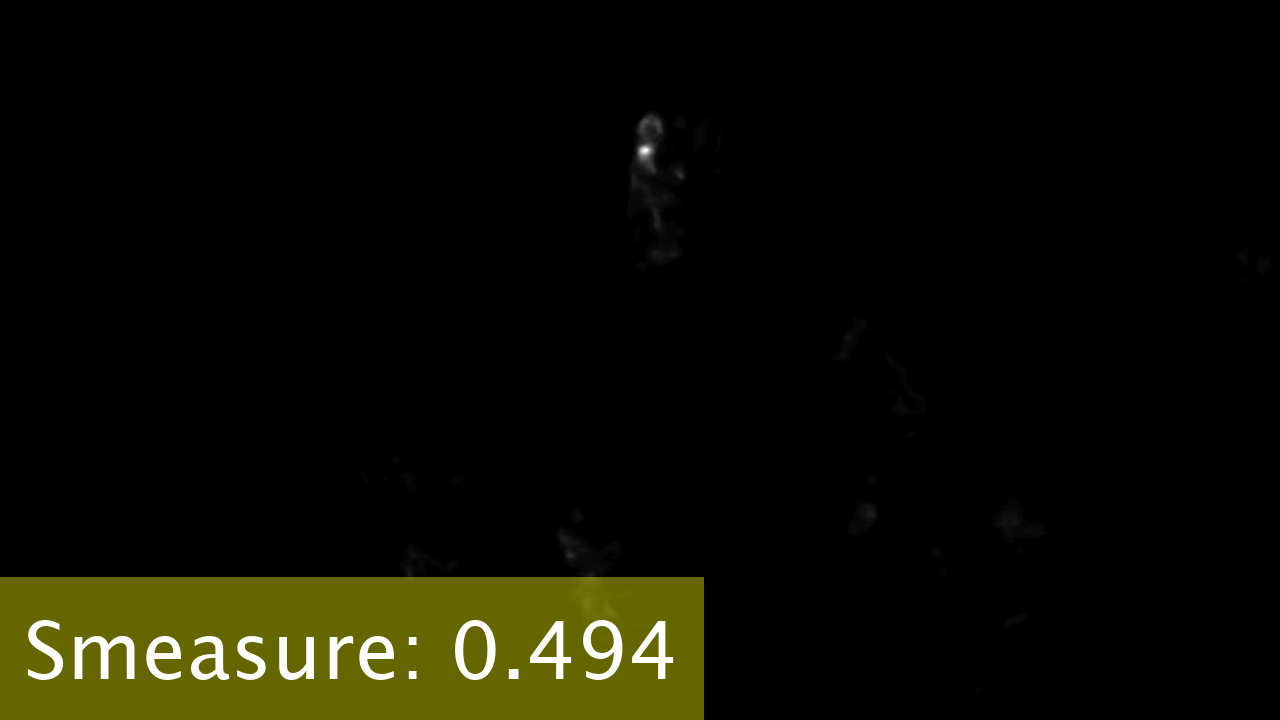} 
    \\
    (j) &
    \includegraphics[width=0.21\linewidth]{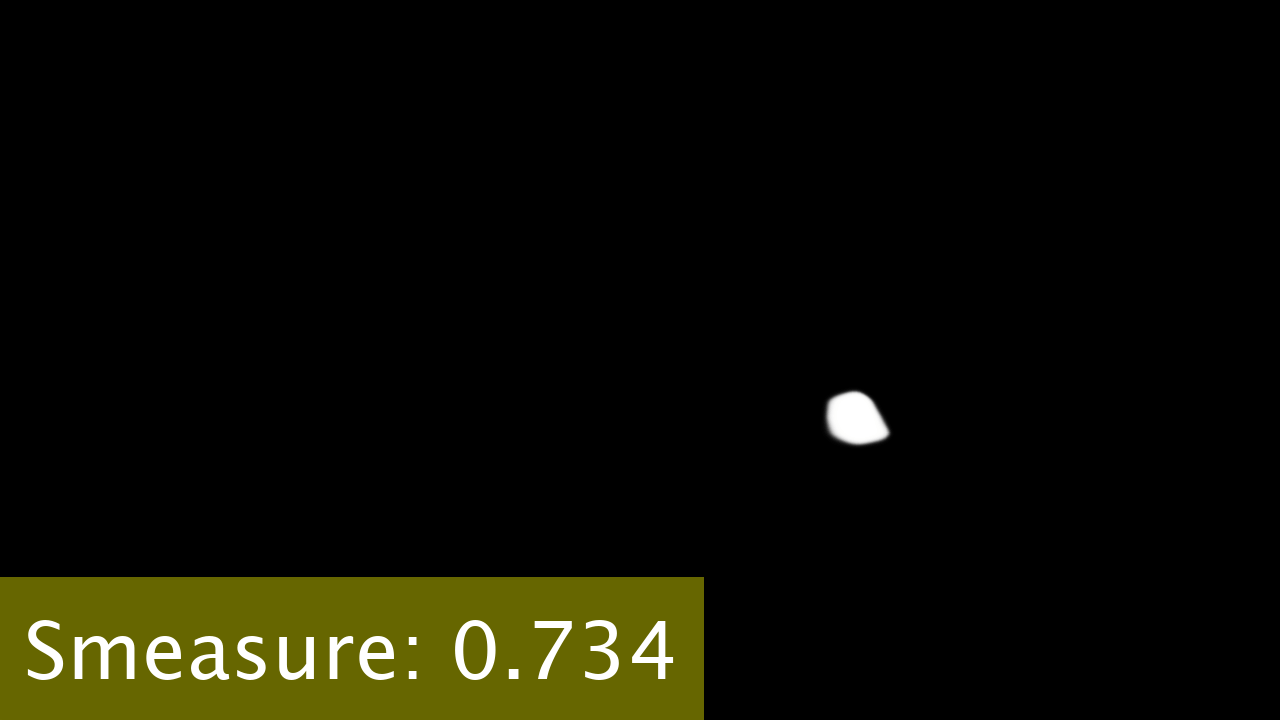}& 
    \includegraphics[width=0.21\linewidth]{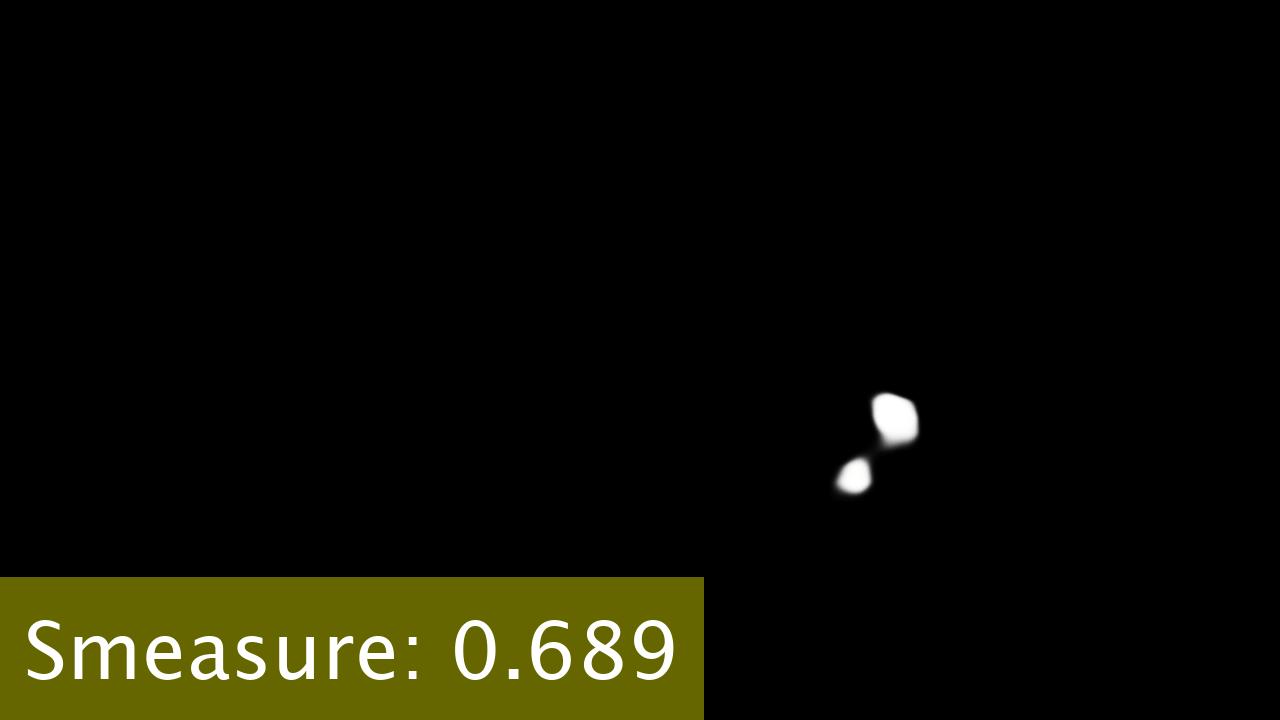} & 
    \includegraphics[width=0.21\linewidth]{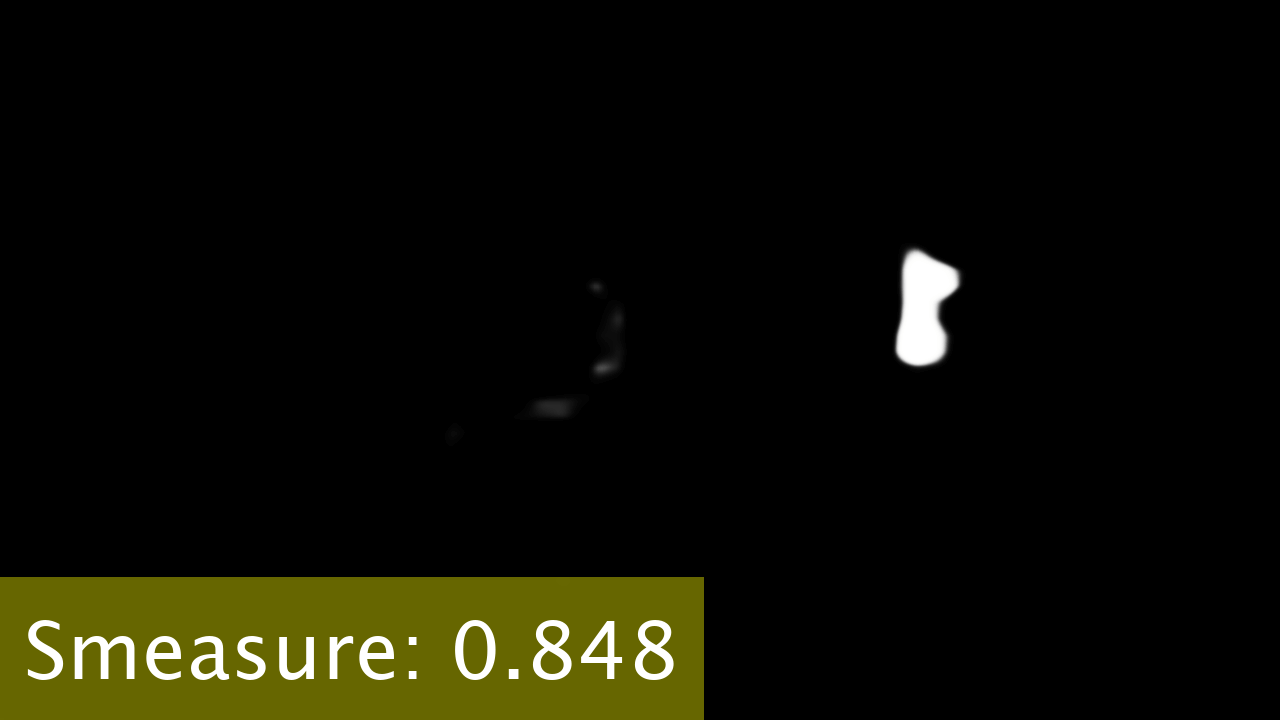} & 
    \includegraphics[width=0.21\linewidth]{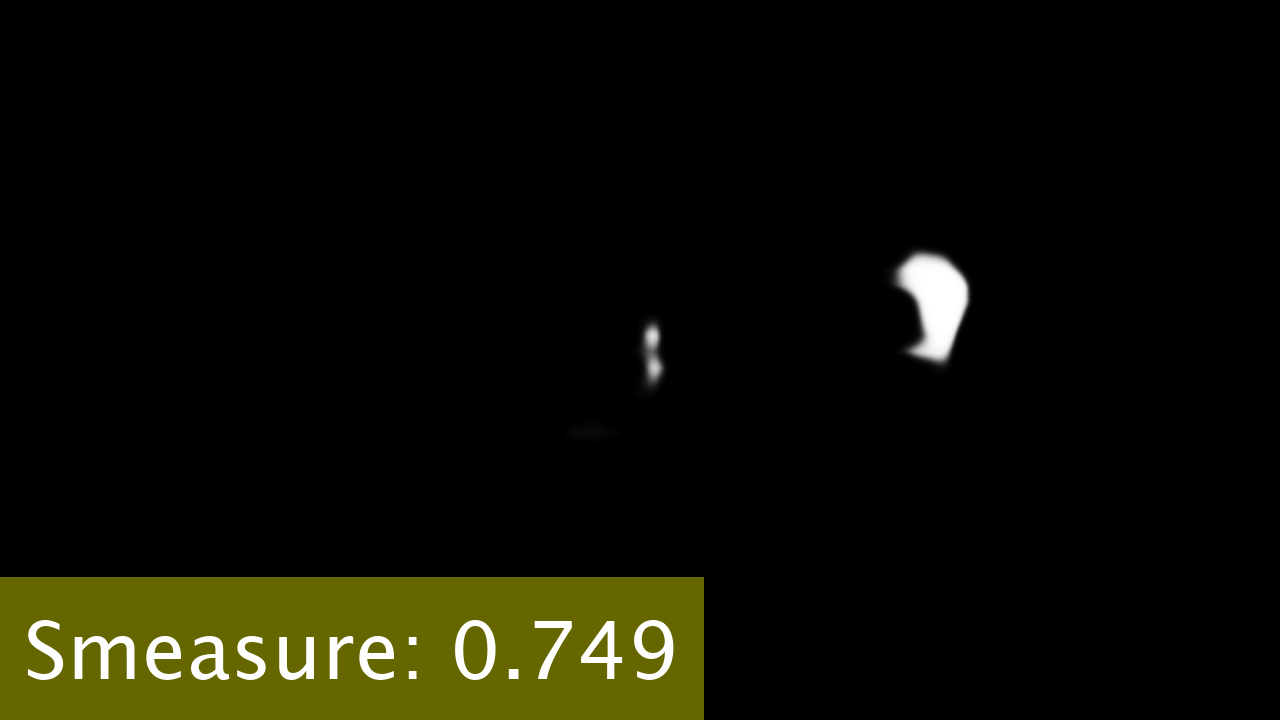} 
    \\
    \end{tabular}
    \vspace{-10pt}
    \caption{Comparison of our proposed network with two top-performing baselines on MoCA-Mask test dataset. Example squences of each row means: (a) (f) Frames, (b) (g) GT, (c) (h) SINet \cite{fan2020Camouflage}, (d) (i) RCRNet \cite{yan2019semi}, (e) (j) SLT-Net (Ours). }
    \label{fig:MoCA_details}
    \vspace{-10pt}
\end{figure*}

\end{document}